\documentclass[10pt,twocolumn,letterpaper]{article}

\usepackage{cvpr}
\usepackage{times}
\usepackage{epsfig}
\usepackage{graphicx}
\usepackage{amsmath}
\usepackage{amssymb}

\usepackage{url}
\usepackage{amsthm}
\usepackage{subcaption}
\usepackage{booktabs}
\usepackage{multirow}
\usepackage{pbox}
\usepackage[utf8]{inputenc}
\usepackage{mathtools}
\usepackage[export]{adjustbox} 
\usepackage{longtable}
\usepackage{rotating}
\usepackage{adjustbox}

\usepackage[moderate,mathdisplays=normal,indent=normal]{savetrees}

\usepackage{import}
\usepackage{tikz}
\usepackage{extarrows}
\usepackage{xparse} 
\usepackage{calc} 
\usepackage{colortbl}
\usepackage{booktabs}
\usepackage{xfrac}
\usepackage{tablefootnote}
\usepackage{tabularx}
\usepackage{standalone}
\definecolor{rowblue}{RGB}{220,230,240}
\definecolor{Orange}{RGB}{255,204,153}
\definecolor{Blue}{RGB}{153,204,255}

\makeatletter
\define@key{Gin}{expandoptions}{%
  \begingroup
  \edef\x{\endgroup
    \noexpand\setkeys{Gin}{#1}%
  }\x
}
\makeatother

\colorlet{innercolor}{black!60}%
\colorlet{outercolor}{gray!05}%

\usetikzlibrary{backgrounds}%

\pgfdeclarelayer{shadow}%
\pgfsetlayers{shadow,main}%

\ExplSyntaxOn
\NewDocumentCommand{\aspectratio}{smo}
 {
  \hbox_set:Nn \l_tmpa_box {\includegraphics{#2}}
  \IfNoValueTF{#3}
   {
    \__student_aspectratio:nn { \box_wd:N \l_tmpa_box } { \box_ht:N \l_tmpa_box }
   }
   {
    \IfBooleanTF{#1}{ \tl_gset:Nx } { \tl_set:Nx } #3
     {
      \__student_aspectratio:nn { \box_wd:N \l_tmpa_box } { \box_ht:N \l_tmpa_box }
     }
   }
 }

\cs_new:Nn \__student_aspectratio:nn
 {
  \fp_eval:n {round( #1 / #2 , 5)}
 }
\ExplSyntaxOff

\newcommand{\tableDir}{src/tables/}%
\newcommand{\figDir}{src/figs/}%
\def\x{\mathbf{x}}%
\def\m{\mathbf{m}}%
\def\z{\mathbf{z}}%
\newlength{\imwidth}%

\usepackage[pagebackref=true,breaklinks=true,letterpaper=true,colorlinks,bookmarks=false]{hyperref}

\cvprfinalcopy 

\ifcvprfinal\fi
\begin{document}

\title{
Generating Object Stamps 
}
%






\author{Youssef A.~Mejjati\\
University of Bath\\
\and
Zejiang Shen\\
Brown University\\
\and
Michael Snower \\
Brown University\\
\and
Aaron Gokaslan \\
Brown University\\
\and
Oliver Wang \\
Adobe Research\\
\and
James Tompkin \\
Brown University\\
\and
Kwang In Kim \\
UNIST\\
}

\twocolumn[{%
\renewcommand\twocolumn[1][]{#1}%
\maketitle
\centering
\newcommand{\mysize}{.17\linewidth}
\begin{tabular}{*{5}{c@{\hspace{2px}}}}
 \includegraphics[width=\mysize,clip,trim=0 0 0 20]{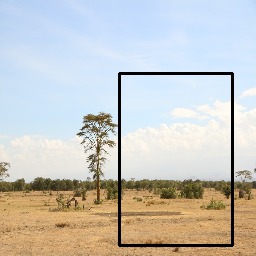} &
  \includegraphics[width=\mysize,clip,trim=0 0 0 20]{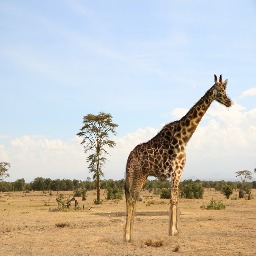} &
 \includegraphics[width=\mysize,clip,trim=0 0 0 20]{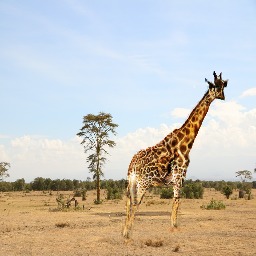} &
  \includegraphics[width=\mysize,clip,trim=0 0 0 20]{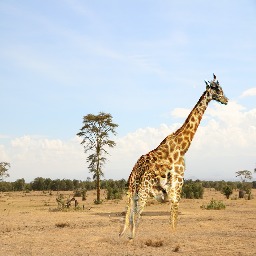}&
  \includegraphics[width=\mysize,clip,trim=0 0 0 20]{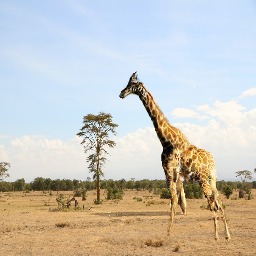} \\
 
  \includegraphics[width=\mysize,clip,trim=0 0 0 20]{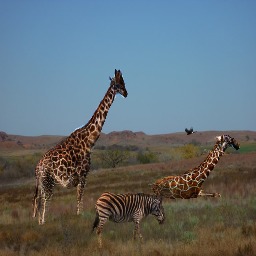} &
 \includegraphics[width=\mysize,clip,trim=0 0 0 20]{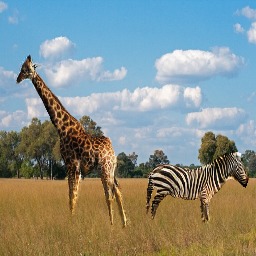} &
 \includegraphics[width=\mysize,clip,trim=0 0 0 20]{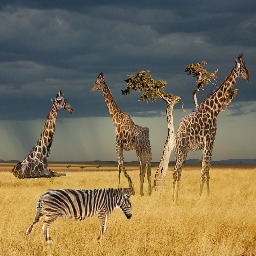} &
 \includegraphics[width=\mysize,clip,trim=0 0 0 20]{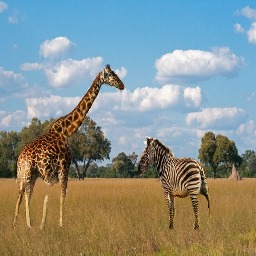} &
 \includegraphics[width=\mysize,clip,trim=0 0 0 20]{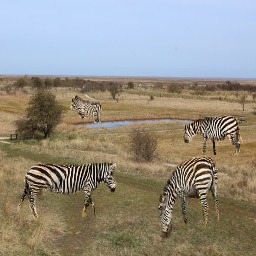} 
\end{tabular}
 \label{fig:teaser}
 \captionof{figure}{\emph{Top row:} Given a user-provided background image, object class (giraffe), and bounding box (\emph{far left}), our method generates objects with diverse shapes and textures (\emph{right}). \emph{Bottom row:} We combine multiple object classes across scenes and match illumination.}
\vspace{5mm}    
}]


\begin{abstract}
We present an algorithm to generate diverse foreground objects and composite them into background images using a GAN architecture.
Given an object class, a user-provided bounding box, and a background image, we first use a mask generator to create an object shape, and then use a texture generator to fill the mask such that the texture integrates with the background.
By separating the problem of object insertion into these two stages, we show that our model allows us to improve the realism of diverse object generation that also agrees with the provided background image.
Our results on the challenging COCO dataset show improved overall quality and diversity compared to state-of-the-art object insertion approaches.
\end{abstract}



\section{Introduction}

Compositing objects into images is a common editing task. 
Given a database of images of a target object class and a specific background image, the task typically proceeds in three steps: 
1) Find an instance of the object in a suitable pose and under similar lighting to the background image; 
2) define the location and size of the object in the background image; and 
3) composite the object onto the background in a visually-consistent way.
Often this process can be cumbersome, and requires time and skill.

Our goal is to reduce the burdens of steps 1 and 3 to create a simple user interface for object compositing.
We begin with a database of images with object masks. 
Importantly, one requirement of our method is that it must work on images taken `in the wild', where objects have varied shapes, backgrounds, are not centered and scaled within the image, and can exist in multiple instances per image sometimes under significant occlusion.
We wish to learn how to represent the object class' appearance variation, and how to match the appearance with background scenes. 
This would let the user generate an object by simply dragging a bounding box over novel background images, and then by sampling multiple shapes and textures from the learned space of the object class.

\paragraph{Motivation:} 
This problem is challenging as it requires understanding both the object class' diversity in appearance and how that appearance changes with the rest of the scene, \eg, how object appearance changes under scene lighting. 
Existing whole-image synthesis methods, such as conditional generative adversarial networks (cGANs), are at odds with our application scenario of placing objects against a \emph{specific} background. 
Further, whole-image methods can `cheat' at foreground generation by altering the background to satisfy the discriminator (\eg, changing green grass to brown savanna when mapping from horse to zebra), which reduces translation quality.
Another option is to simultaneously generate object shape and texture, and paste the result into a specific background. 
However, doing so has several limitations.
For one, it prohibits the disentanglement of shape and texture, which is a useful interaction tool.
Second, the discriminator's ability to verify realism is overwhelmingly driven by texture~\cite{Geirhos2018}, which allows the generator to explore \emph{implausible shapes} to satisfy the discriminator.


\begin{figure*}[t]
    \centering
    \includegraphics[clip,width=\linewidth]{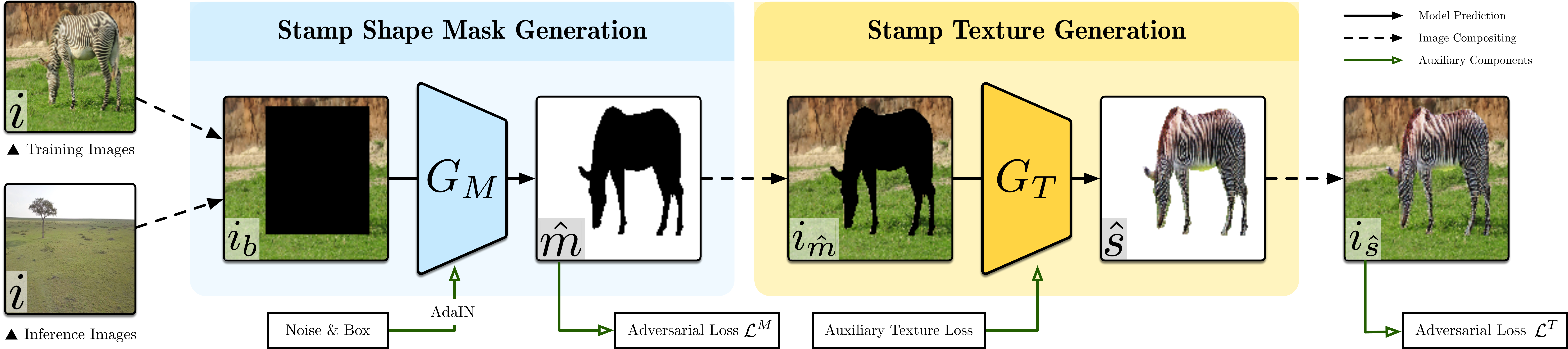}
    \caption{\emph{Overview of our image generation pipeline.} 
    Given an input image $i$, a bounding box is cut out to create $i_b$, and this is then fed into the mask generator $G_M$. $G_M$ creates a plausible object mask from the desired category (in this case, zebra). 
    Our texture generator then fills in the mask to produce a final composited image, while a texture discriminator ensure image realism.
    }
    \label{fig:architecture}
    \vspace{-0.25cm}
\end{figure*}

\paragraph{Our approach:} Inspired by curriculum learning approaches, we decompose the generation process into two steps. 
First, we synthesize foreground masks (representing the shape of the object) conditioned on both a specific background image and user specified bounding box.
Next, we generate realistic texture within the mask via \emph{conditioning on the shape mask and background image}. 
This allows us to fix the shape while varying the texture (or vice versa), and allows us to explicitly enforce the realism of the shape and texture with separate discriminators.
%
At each stage, we inject randomness expressed by a latent random variable, which allows the generation of multiple shape masks and textures given the same bounding box coordinates.

We demonstrate our method trained on images and segmentation masks ``in-the-wild'' from COCO~\cite{MSCOCO2014}.
Recent work on the related task of image-to-image translation has shown that methods fail when trained on COCO images due to the large variation in the pose and appearance of objects and backgrounds~\cite{wu2019transgaga}.
However, by splitting the problem into shape and texture generation, and leveraging masks to isolate foreground objects, we show that our method can generate plausible results despite the dataset variance.

Disentangling shape from texture allows our method to be used for three editing applications: generating an object with both new shape and texture and compositing it with a background image (`object stamps'), changing the texture of an existing object within an image (`object retexturing'), and adding objects with an existing shape into a new background image (`object insertion').

We compare our approach with solutions which synthesize a whole image at once, and well as state-of-the-art approaches that separate foreground and background generation~\cite{HierarchicalSemanticImageManpulation,finegan}. On the challenging COCO dataset with high variance, we found that existing approaches can mode collapse during training. However, our method is more robust: our curriculum learning strategy is easier to train than on both tasks jointly given a fixed parameter and data budget, and produces higher-quality mask shapes and textures than compared methods.





\section{Related Work}

\paragraph{Image compositing:} Compositing has a long history of use in graphics and vision research~\cite{perez2003poisson,agarwala2004interactive}, with recent works applying deep neural networks to increase quality and flexibility.
Deep Image Harmonization~\cite{tsaiCVPR2017} trains a neural network to realistically change the lighting and fine texture of a given foreground region by considering context from segmentation maps. 
STN-GAN~\cite{lin2018st} takes a background image and foreground object, and performs a low-dimensional warp on the foreground object to make its appearance more natural.
Similarly, SEI-GAN~\cite{ostyakov2018seigan} learns to insert an existing object into another image. 
While our goal is related, we aim to learn how to generate \emph{new} instances of the object class rather than amend an existing image. 
We wish to generate realistic foreground shapes and then fill them in with corresponding texture such that the foreground is blended with the background.

\paragraph{Generative Adversarial Networks (GANs):} 
GANs have shown tremendous progress in learning-based image generation~\cite{goodfellow2014generative}. 
Networks like StyleGAN~\cite{karras2018style} can produce realistic results; however, they generate entire images, and have been shown to perform best on datasets with restricted variability (e.g., aligned faces or street scenes). 
In particular, background variability can causes GAN generators to produce blurry outputs.
This is a problem in the case of image classes with large background variability, like the animal classes in COCO, and often leads to a significant decrease in quality. 
Gau-GAN~\cite{park2019SPADE} uses semantic segmentation to provide hints about the shape and location of each object in the scene. 
However, since their technique generates images only from a segmentation map, it cannot be easily applied to existing background images. 
Zhan et al.~\cite{zhan2019adaptive} demonstrate image composition by learning geometric transforms on existing instances, but this technique does not generate new object shapes and textures. 
Unpaired image translation networks~\cite{kim2017learning,multimodal2018,zhu2017unpaired, mejjati2018unsupervised} perform fully automated domain translation, but struggle to cope with large deformations~\cite{Gokaslan2018} making them unsuitable for object stamping. 
Instead, our goal is to generate a realistic foreground object which integrates into an existing background image, which cannot be easily accomplished with these methods.
\paragraph{Curriculum image generation:}
Singh \etal's FineGAN~\cite{finegan} can generate convincing images from a database of a given class by separating the operation into sequential background and foreground generation. 
Composite-GAN~\cite{kwak2016generating} and LRGAN~\cite{yang2017lr} similarly recursively generate the background and foreground, but do not disentangle shape and appearance in the same way as FineGAN.
Our method is inspired by such an approach, as it demonstrates the power of decomposed generation. 
However, our work differs in two key ways:
First, we generate a foreground with respect to an \emph{existing} background.
Second, our method allows for control over the location and scale of the generated foreground object conditioned on the background content.

Most similar to our work, Hong \etal~\cite{HierarchicalSemanticImageManpulation} `complete' a user bounding box by generating an object instance.
Our model has three key differences: we can generate multiple object shapes and textures, we use the entire background for conditioning to improve harmonization, and we generate content for the mask region only to improve result quality. 
We compare our result directly with Hong \etal's approach, and show that we can generate results with overall higher quality and instance diversity.


\section{Method}
\label{sec:Method}

Our model takes as input a background image $i\in \mathbb{R}^{W\times H\times 3}$ of width $W$ and height $H$ drawn from a domain $\mathcal{I}$, a bounding box $\mathbf{b}\in\mathbb{R}^4$ containing rectangle vertices, and an object class $c$.
From $\mathbf{b}$, we construct a binary bounding box \emph{mask image} $b\in \mathbb{R}^{W\times H}$ with the region inside the box set to 1, and 0 otherwise.

The goal is to generate an object stamp inside the bounding box and to composite it with the background image $i$ (Figure~\ref{fig:architecture}).
We achieve this by first generating a stamp mask $\hat{m}\in\mathbb{R}^{W\times H}$, and then generating a textured stamp $\hat{s}\in\mathbb{R}^{H\times W\times3}$ such that when composited into the final image $i_{\hat{s}}$ is indistinguishable from images in $\mathcal{I}$, where $i_{\hat{s}}=i\odot(1- \hat{m}) + \hat{s}\odot\hat{m}$, and where $\odot$ is the element-wise product.

\begin{figure}[t]
    \centering
    \includegraphics[clip,width=0.8\linewidth]{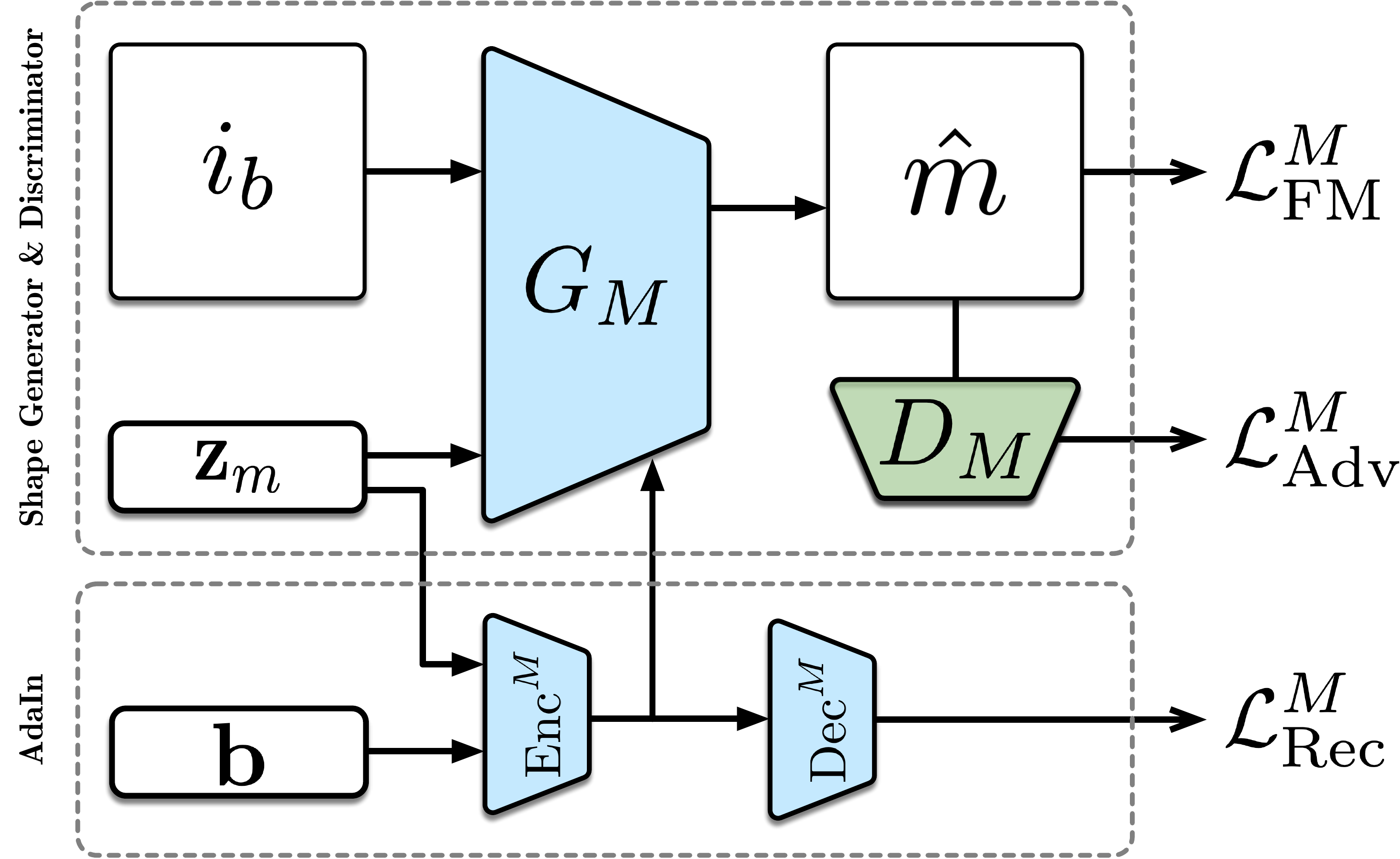}
    \caption{\emph{Our stamp shape generation model.} This generates a shape mask $\hat{m}$ given a random vector $\mathbf{z}_m$ and input image $i_b$ with bounding box region zeroed-out. We use a discriminator $D_M$ and adversarial loss $\mathcal{L}_{\text{Adv}}^M$ to train the generator. To improve output quality, we apply a feature matching loss $\mathcal{L}_{\text{FM}}^M$ on $\hat{m}$, and use AdaIN injection of $\mathbf{z}_m$ and $\mathbf{b}$. To maintain the diversity of generated shapes, we add a random vector reconstruction loss $\mathcal{L}_{\text{Rec}}^M$.}
    \label{fig:architecture-shape}
\end{figure}

\subsection{Stamp shape generation model}
\label{sec:Method-shape}

We train a generator $G_M$ conditioned on a bounding box, background image, and a random vector $\mathbf{z}_m$ drawn from a Gaussian with mean 0 and variance 1: $\mathbf{z}_m \sim \mathcal{N}(0,I)$.
Importantly, we train the generator on images that contain object instances and their respective masks.
Therefore, to allow us to condition on background content at test time that does \emph{not} contain the object in question, we first zero-out the bounding box region $i_b=i\odot(1- b)$.
This prevents the network from trivially learning to segment existing instances in the training data.

The generator $G_M$ produces a binary mask $\hat{m}$ for the shape of the stamp object inside the bounding box region, \eg, $\hat{m} = G_M(i_b, \mathbf{z}_m)$. 
We train $G_M$ adversarially: $G_M$ attempts to generate realistic shapes to fool a discriminator $D_M$, while $D_M$ attempts to classify generated masks separately from real training data masks. 
$D_M$ is a CNN which takes as input the shape mask and corresponding bounding box: $D_M(m,b)$. 
We use a hinge-GAN loss $\mathcal{L}^M_{\text{Adv}}$ to train both $G_M$ and $D_M$ for better stability~\cite{lim2017geometric,tran2017deep,miyato2018spectral}:
%
%
%
\begin{align}
    \mathcal{L}^M_{\text{Adv}}(G_M,D_M) = & \mathbb{E}_{m}{[\min(0, D_M(m,b)-1)]} + \nonumber \\ 
    &\mathbb{E}_{\hat{m}}{[ \min(0,-D_M(\hat{m},b))-1]},
    \label{eq:Loss_M_Adv}
\end{align}
where $m$ is a ground-truth shape mask and $\hat{m}$ is generated. 
%
%
%

Next, we describe how we use adaptive instance normalization (AdaIN)~\cite{huang2017adain} in $G_M$ to condition the network on the input noise $\mathbf{z}_m$.
In our case, we wish to inject the bounding box $b$ and the latent vector $\mathbf{z}_m$ during shape generation. 
For this, similar to prior work~\cite{huang2017adain}, we use a small fully-connected feed-forward network (MLP) encoder $\text{Enc}^M$ to take input $b$ and $\mathbf{z}_m$ and predict affine transformation parameters $\text{Enc}^M(b,\mathbf{z}_m)$ for the instance normalization layers.

One issue we found is that AdaIN can learn to ignore $\mathbf{z}_m$ by using only $b$, which reduces diversity in generation. 
To overcome this, we propose a reconstruction loss $\mathcal{L}^M_{\text{Rec}}$ over $\mathbf{z}_m$ via an MLP decoder complement $\text{Dec}^M$ to $\text{Enc}^M$:
\begin{align}
    \mathcal{L}^M_{\text{Rec}}(\text{Enc}^M, \text{Dec}^M) = \frac{1}{|\mathbf{z}_m|}{\|\mathbf{z}_m - \text{Dec}^M(\text{Enc}^M(b, \mathbf{z}_m))\|_1^2}.
    \label{eq:Loss_M_Rec}
\end{align}
Unlike prior work~\cite{zhu2017bicycle}, $\text{Dec}^M$ decodes the latent vector from $\text{Enc}^M(b,\mathbf{z}_m)$ rather than the output mask $\hat{m}$, which we found to perform better in our experiments.
This loss directly enforces that the AdaIN parameters depend on the random vector $\mathbf{z}_m$, and so helps to maintain diversity.

This yields diverse results, but we found that the masks lacked fine detail.
Therefore, we propose a variant on the commonly used deep feature matching loss~\cite{salimans2016improved,zhang2018unreasonable}, which has been shown to enhance image sharpness by enforcing that real and generated images elicit similar feature responses in each layer $l$ of the discriminator $D_M^{(l)}$, via a squared $L_2$ norm.
In our case, since our generated mask $\hat{m}$ does not match a single real mask, but rather a \emph{distribution} (\eg, it varies by $\mathbf{z}_m$), we cannot directly compare the feature responses from ground truth masks $m$ with $\hat{m}$.
Instead, we compute a moving average of the feature response from each batch of training examples $\bar{D}_M^{(l)}$ to obtain a \emph{mean distribution} of the training shapes for comparison:
\begin{align}
    \mathcal{L}^M_{\text{FM}}(D_M) = \mathbb{E}_{\hat{m}, m}\sum_{l=1}^{L}||{D_M^{(l)}(\hat{m},b) - \bar{D}_M^{(l)}(m,b)}||_2^2.
    \label{eq:Loss_M_FM}
\end{align}
Rather than averaging over generated masks, which would be blurry, our proposed minibatch feature matching loss approach generates sharp binary image masks. We show in our experimental ablation study that this quantitatively improves results.

%


Our combined learning objective for the shape mask generator $\mathcal{L}^M$ is a weighted combination of the aforementioned losses:
\begin{align}
    \mathcal{L}^M = \mathcal{L}^M_{\text{Adv}}+ \lambda^M_{\text{FM}}\mathcal{L}^M_{\text{FM}} + \lambda^M_{\text{Rec}}\mathcal{L}^M_{\text{Rec}},
    \label{eq:Loss_M}
\end{align}
where $\lambda^M_{\text{FM}}$, $\lambda^M_{\text{Rec}}\geq 0$ are hyperparameters.
\begin{figure}[t]
    \centering
    \includegraphics[clip,width=0.8\linewidth]{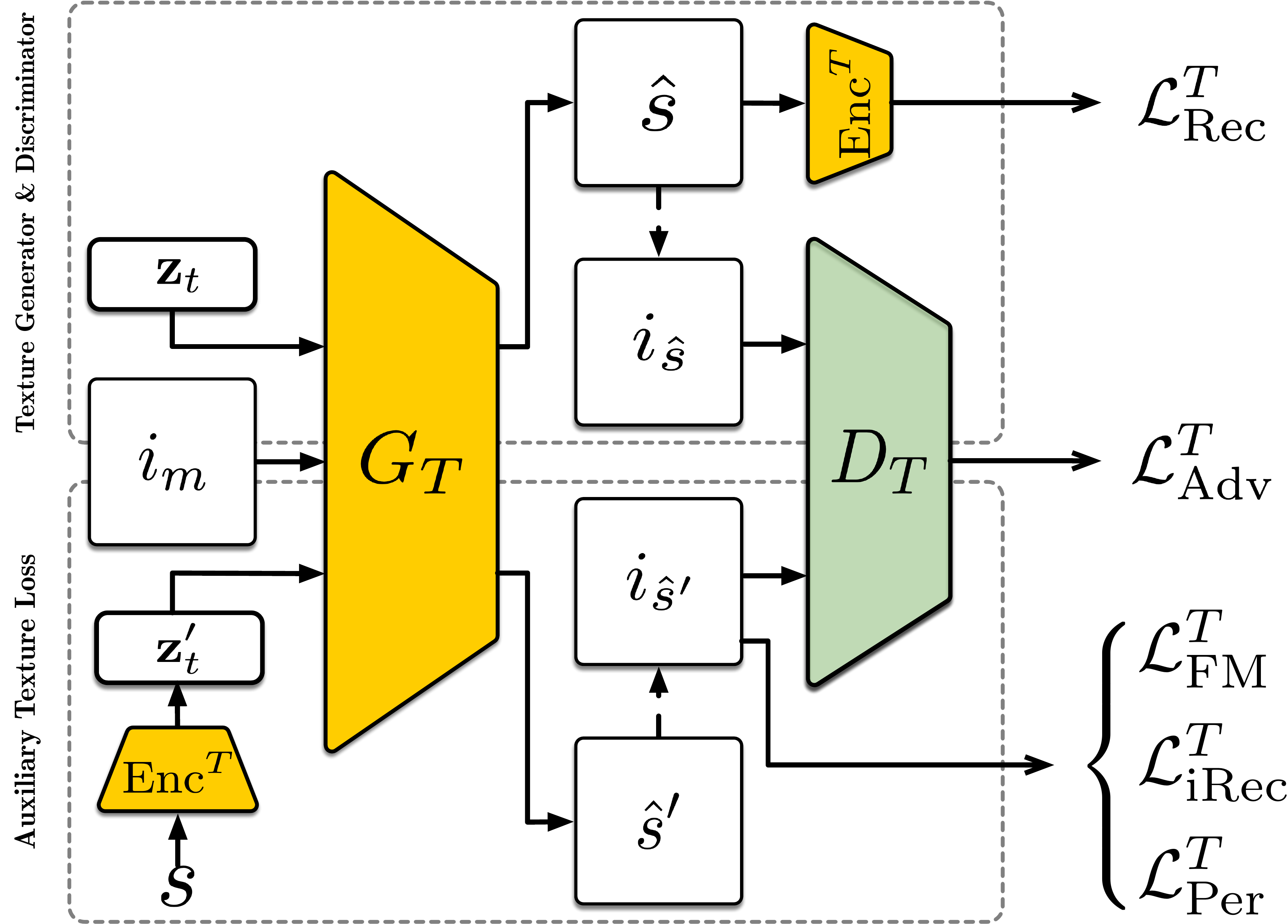}
    \caption{\emph{Our stamp texture generation model.} We train a generator $G_T$ via a discriminator $D_T$. Feature matching $\mathcal{L}^T_\text{{FM}}$, perceptual $\mathcal{L}^T_\text{{Per}}$, and auxiliary losses $\mathcal{L}^T_{Rec}$ and $\mathcal{L}^T_\text{{iRec}}$ help to simultaneously maintain fine texture detail and diversity, while conditioning on random vector $\mathbf{z}_t$ provides user control.}
    \label{fig:architecture-texture}
\end{figure}

\subsection{Stamp texture generation model}
\label{sec:Method-texture}

The goal of the texture generator $G_T$ is to create realistic textures that match both the pose in the shape mask (e.g., frontal \vs profile) and the lighting in the background. 
Given a generated shape mask $\hat{m}$, we first zero-out the shape from the input image to mark the area that needs texturing: $i_{\hat{m}}=i\odot(1-\hat{m})$. 
Then, the stamp texture generation $G_T$ synthesizes the texture inside the empty region while still having access to the texture of the surrounding background: $\hat{s}=G_T(i_{\hat{m}}, \mathbf{z}_t)$. 
$\hat{s}\in\mathbb{R}^{H\times W\times3}$ is the generated object stamp image, and $\mathbf{z}_t \sim \mathcal{N}(0,I)$ is a random vector that adds stochasticity for diverse generation. 
For $G_T$, we use a BicycleGAN-like \cite{zhu2017toward} architecture to preserve both texture quality and diversity.
Similar to $G_M$, we train $G_T$ adversarially via a texture discriminator $D_T$.



The input to the discriminator $D_{T}$ is the channel-wise concatenation of the real image and mask tuple $(i_s,m)$ or the fake equivalent $(i_{\hat{s}},\hat{m})$. 
Passing the mask to $D_T$ tells the discriminator to expect an instance of the object at that location.
This is important as, at training time, the background image $i_{\hat{s}}$ may contain \emph{multiple} instances of the target class, and so $D_T$ would be satisfied by the generator filling in the mask with background instead of object.

Unlike $D_M$, we do not use a moving average feature matching loss as we found it to blur the feature responses and cause the generator to miss object details.
Instead, we use a CNN to encode the ground-truth texture of the foreground $s=i \odot m$ as a latent vector $\mathbf{z}_t'=\text{Enc}^T(s)$, and then use $\mathbf{z}_t'$ and $m$ to generate another stamp texture $\hat{s}'=G_T(i_m,\mathbf{z}_t')$ and corresponding stamped image $i_{\hat{s}'}=i\odot(1-m)+\hat{s}'\odot m$. 
As $\mathbf{z}_t'$ depends on $s$, this indirectly conditions $i_{\hat{s}'}$ on $s$, and allows us to apply a feature matching loss without losing detail:
\begin{align}
    \mathcal{L}^T_{\text{FM}}(D_T) = \mathbb{E}_{i_{\hat{s}'}, i_s}\sum_{l}||{D_T^{(l)}(i_{\hat{s}'},m) - D_T^{(l)}(i_s,m)}||_1^2.
    \label{eq:Loss_T_FM}
\end{align}
Here, $D_T^{(l)}$ is the output of the $l$-th layer in $D_T$. 
We also apply an additional $L_1$ image reconstruction loss $\mathcal{L}^T_{\text{iRec}}$ to aid in the description of the texture corresponding to $\mathbf{z}_t'$: $\mathcal{L}^T_{\text{iRec}} =\frac{1}{|i_s|}\sum_{l=1}||{i_{\hat{s}'}- i_s}||_1^2$.


%

As an extra constraint on $\mathbf{z}_t'$, we wish for its distribution $\mathcal{Z}_t'$ to be similar to the distribution of $\mathbf{z}_t$, such that the generator $G_T$ cannot use any distribution difference to fool $D_T$. 
Thus, we use the re-parametrization trick~\cite{kingma2013auto} on $\text{Enc}^T$ and add a KL-divergence loss $\text{KL}(\mathcal{Z}_t'||\mathcal{N}(0,I))$ to promote distribution consistency. 


\paragraph{Texture architecture:}
Compared to our mask architecture, we make five additional changes for our texture model. First, unlike for masks, for texture there is no need for an AdaIN component to inject auxiliary information into the generator: Since $D_T$ is trained on the full image $i_s$ and $i_{\hat{s}}$, the generator must use the shape information even without AdaIN. 

Second, to make sure that the latent vector $\mathbf{z}_t$ is not ignored, we penalize a latent texture reconstruction loss on $\mathbf{z}_t$ via an encoder on $\hat{s}$:
\begin{align}
    \mathcal{L}^T_{\text{Rec}}(\text{Enc}^T) = \frac{1}{|\mathbf{z}_t|}{||\mathbf{z}_t - \text{Enc}^T(\hat{s})) ||_1^2}.
    \label{eq:Loss_T_Rec}
\end{align}

Similar to Zhu et al.~\cite{zhu2017toward}, we do not update $\text{Enc}^T$ when propagating the gradients from Eq.~\ref{eq:Loss_T_Rec}. This avoids that $\text{Enc}^T$ hides information in the data, making it easy to reconstruct~\cite{chu2017cyclesteg}.

Third, to aid realism in both $i_{\hat{s}}$ and $i_{\hat{s}'}$, we use both to train $G_T$ and $G_M$. As such, the adversarial loss $\mathcal{L}_{\text{Adv}}^T$ becomes:
\begin{align}
    \mathcal{L}^T_{\text{Adv}}(G_T,D_T) = \
    &\mathbb{E}_{i_s}{[\min(0, D_T(i_s,m)-1)]} + \nonumber \\
    &\mathbb{E}_{i_{\hat{s}}}{[\min(0, -D_T(i_{\hat{s}},m)-1)]} + \nonumber \\ 
    &\mathbb{E}_{i_{\hat{s}'}}{[\min(0, -D_T(i_{\hat{s}'},m)-1)]}.
    \label{eq:Loss_T_Adv}
\end{align}

Fourth, we add a perceptual loss~\cite{johnson2016perceptual} to help recreate fine textural details, \eg, the tail of a giraffe.
This loss uses a pre-trained VGG16 network to extract features for two image instances, then enforces that their feature activations are as similar as possible:
\begin{align}
    \mathcal{L}^T_{\text{Per}}(G_T) = \frac{1}{N} \left\|\phi(i) - \phi(i_{\hat{s}'})\right\|_1^2,
    \label{eq:Loss_T_Per}
\end{align}
where $\phi(i)$ is the third layer output the VGG16 network, and extracts $N$ features from $i$.

Fifth, and finally, we add Gaussian noise $\mathcal{N}(0,I)$ to the texture decoder \emph{a la} StyleGAN~\cite{karras2018style}. This helps the diversity of the results and improves image generation quality (Table~\ref{tab:KIDtexture}).

The overall texture generation training loss $\mathcal{L}^T$ is:
\begin{align}
    \mathcal{L}^T =  
    \mathcal{L}^T_{\text{Adv}} 
  &+\lambda^T_{\text{Rec}}\mathcal{L}^T_{\text{Rec}} +
     \lambda^T_{\text{KL}}\text{KL}(\mathcal{Z}_t'||\mathcal{N}(0,I))  \nonumber \\ 
  &+\lambda^T_{\text{FM}}\mathcal{L}^T_{\text{FM}} +
     \lambda^T_{\text{Per}}\mathcal{L}^T_{\text{Per}} + 
     \lambda^T_{\text{iRec}}\mathcal{L}^T_{\text{iRec}},
    \label{eq:Loss_T}
\end{align}
where $\lambda^T_{\text{Rec}}$, $\lambda^T_{\text{KL}}$, $\lambda^T_{\text{FM}}$,$\lambda^T_{\text{Per}}$,and $\lambda^T_{\text{iRec}}\geq 0$ are hyperparameters. 
$\lambda^T_{\text{Rec}}$ and the $\text{KL}$ divergence increase diversity, while the $\lambda^T_{\text{FM}}$, $\lambda^T_{\text{Per}}$, and $\lambda^T_{\text{iRec}}$  improve the quality of the generated texture.
We present a detailed architecture description in the supplemental material.

\section{Experiments}
\label{sec:experiments}

\begin{figure*}[tb]
\centering
\newcommand{\tableTitle}[1]{\scriptsize{#1}}%
\setlength{\imwidth}{0.122\linewidth}
\setlength{\tabcolsep}{0.3pt}%

\setbox1=\hbox{\includegraphics[width=\imwidth]{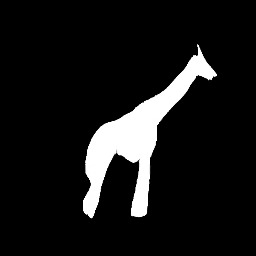}}%
\setbox2=\hbox{\includegraphics[width=\imwidth]{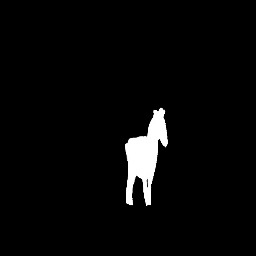}}%

\begin{tabular}{cc@{\hspace{1mm}}cc@{\hspace{2mm}}cc@{\hspace{1mm}}cc}
\includegraphics[width=\imwidth]{figures_cvpr/figgiraffe/m_24_2.jpg}\llap{\makebox[\wd1][l]{\raisebox{0.5\imwidth}{\includegraphics[cfbox=white,width=0.5\imwidth]{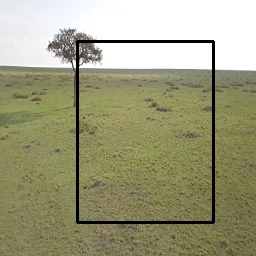}}}}&
\includegraphics[width=\imwidth]{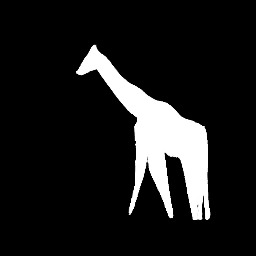}& 
\includegraphics[width=\imwidth]{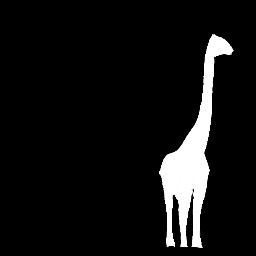}\llap{\makebox[\wd1][l]{\raisebox{0.5\imwidth}{\includegraphics[cfbox=white,width=0.5\imwidth]{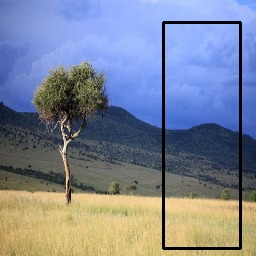}}}}& 
\includegraphics[width=\imwidth]{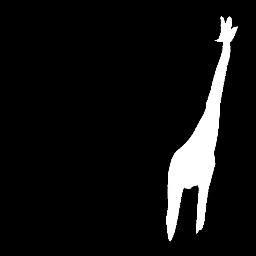}&
\includegraphics[width=\imwidth]{figures_cvpr/figzebra/m_4_0.jpg}\llap{\makebox[\wd1][l]{\raisebox{0.5\imwidth}{\includegraphics[cfbox=white,width=0.5\imwidth]{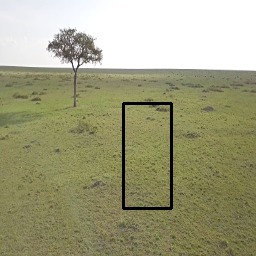}}}}&
\includegraphics[width=\imwidth]{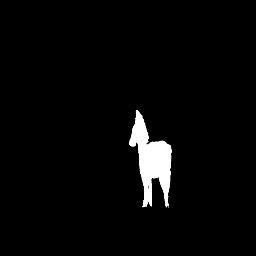}& 
\includegraphics[width=\imwidth]{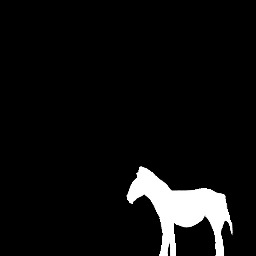}\llap{\makebox[\wd1][l]{\raisebox{0.5\imwidth}{\includegraphics[cfbox=white,width=0.5\imwidth]{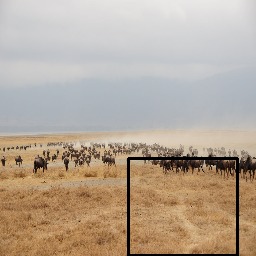}}}}& 
\includegraphics[width=\imwidth]{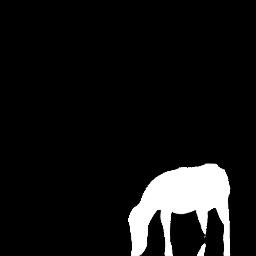}\\

\includegraphics[width=\imwidth]{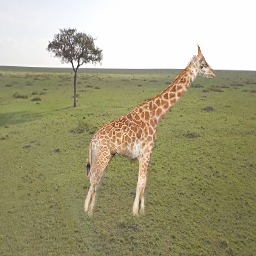}&
\includegraphics[width=\imwidth]{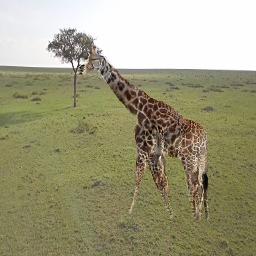}& 
\includegraphics[width=\imwidth]{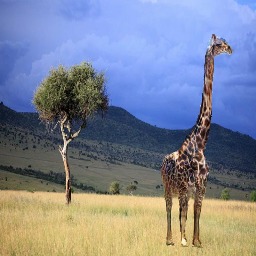}& 
\includegraphics[width=\imwidth]{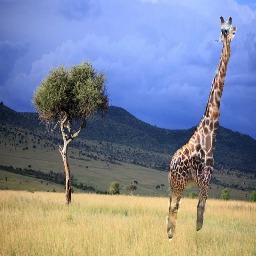}&
\includegraphics[width=\imwidth]{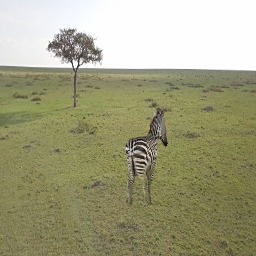}&
\includegraphics[width=\imwidth]{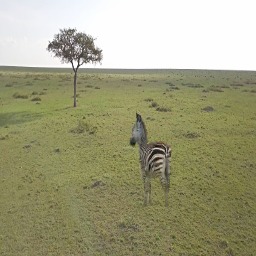}& 
\includegraphics[width=\imwidth]{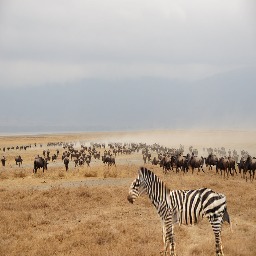}& 
\includegraphics[width=\imwidth]{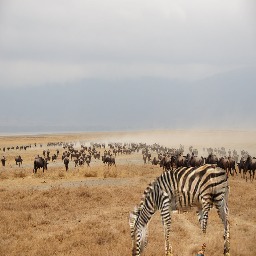}\\

\includegraphics[width=\imwidth]{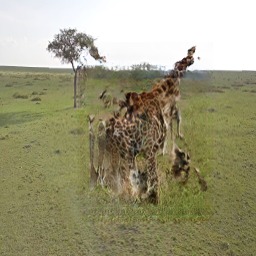}&
\includegraphics[width=\imwidth]{figures_cvpr/figgiraffe/im1_template_25.jpg}&
\includegraphics[width=\imwidth]{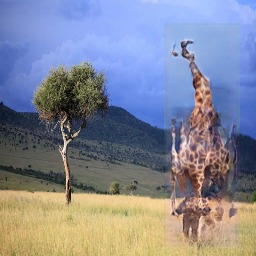}&
\includegraphics[width=\imwidth]{figures_cvpr/figgiraffe/im3_template_50.jpg}&
\includegraphics[width=\imwidth]{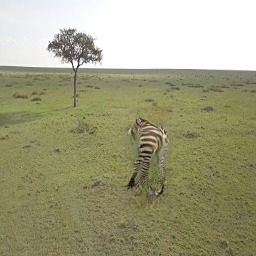}&
\includegraphics[width=\imwidth]{figures_cvpr/figzebra/im1_template_5.jpg}&
\includegraphics[width=\imwidth]{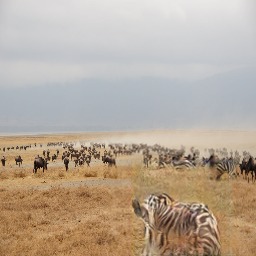}&
\includegraphics[width=\imwidth]{figures_cvpr/figzebra/im2_template_23.jpg}\\

\includegraphics[width=\imwidth]{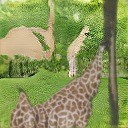}&
\includegraphics[width=\imwidth]{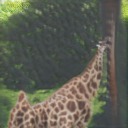}&
\includegraphics[width=\imwidth]{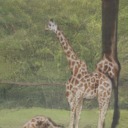}&
\includegraphics[width=\imwidth]{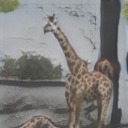}&
\includegraphics[width=\imwidth]{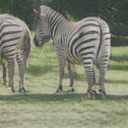}&
\includegraphics[width=\imwidth]{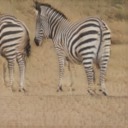}&
\includegraphics[width=\imwidth]{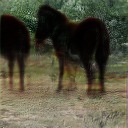}&
\includegraphics[width=\imwidth]{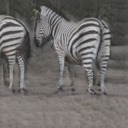}\\

\includegraphics[width=\imwidth]{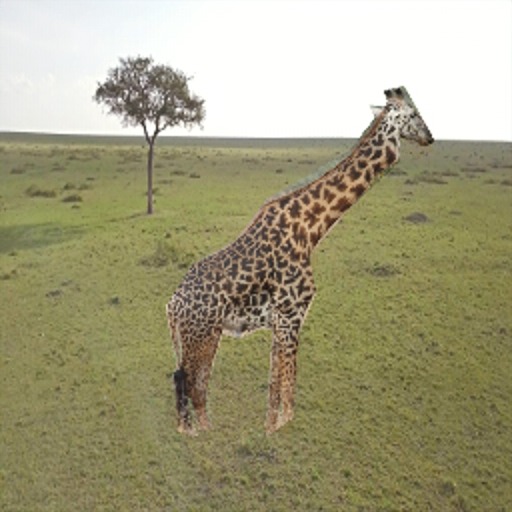}&
\includegraphics[width=\imwidth]{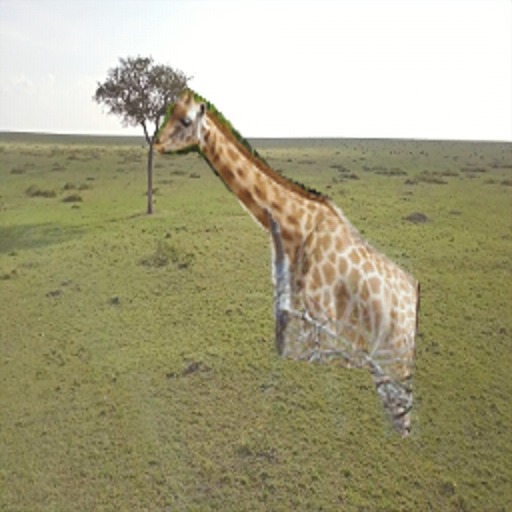}& 
\includegraphics[width=\imwidth]{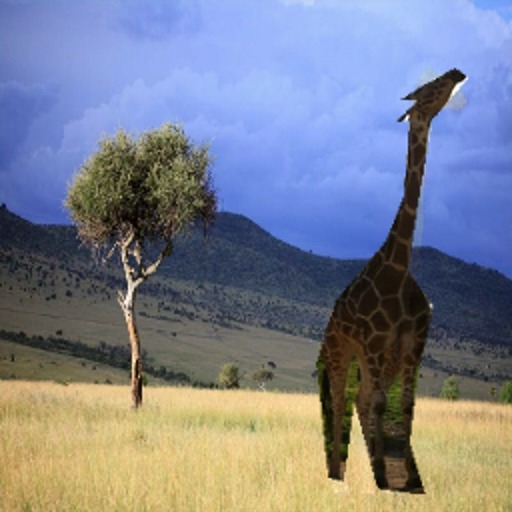}& 
\includegraphics[width=\imwidth]{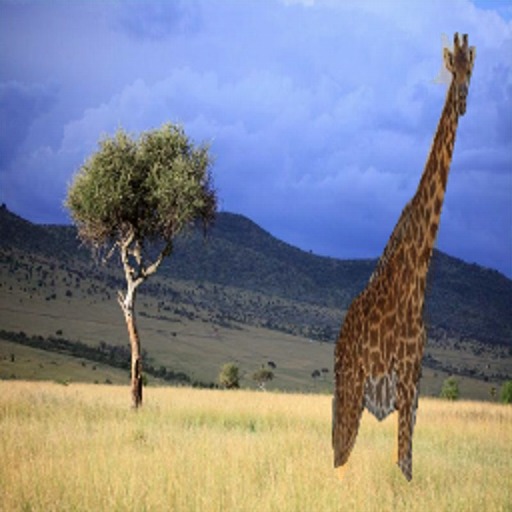}&
\includegraphics[width=\imwidth]{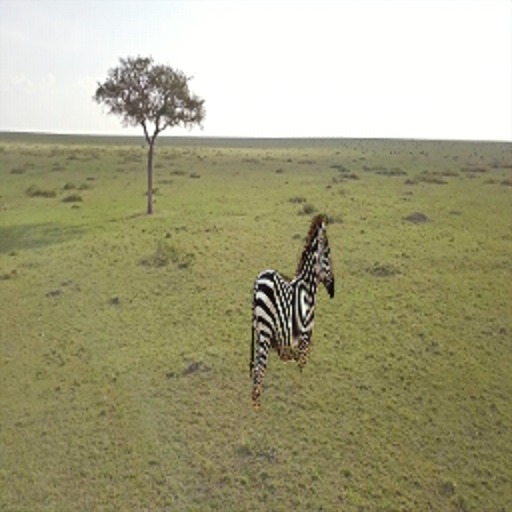}&
\includegraphics[width=\imwidth]{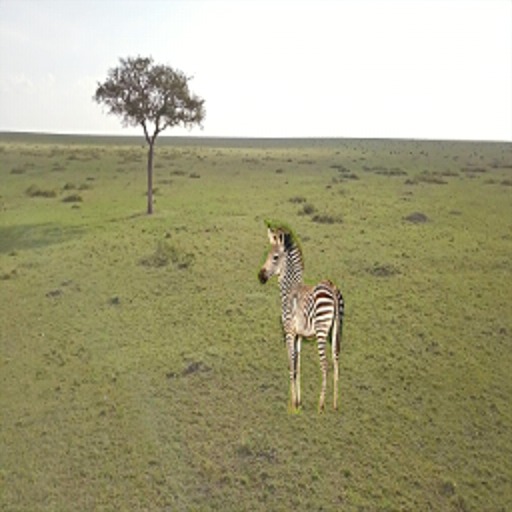}&
\includegraphics[width=\imwidth]{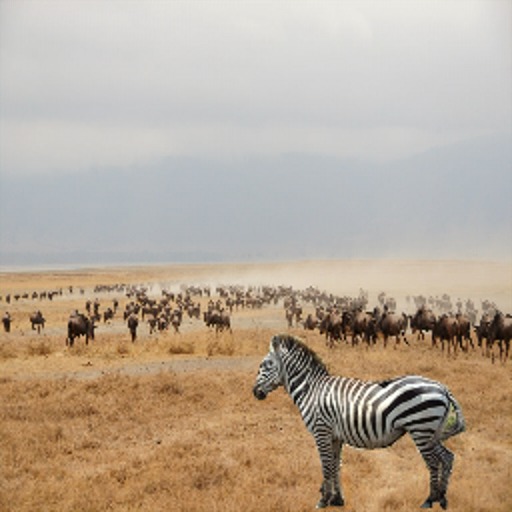}&
\includegraphics[width=\imwidth]{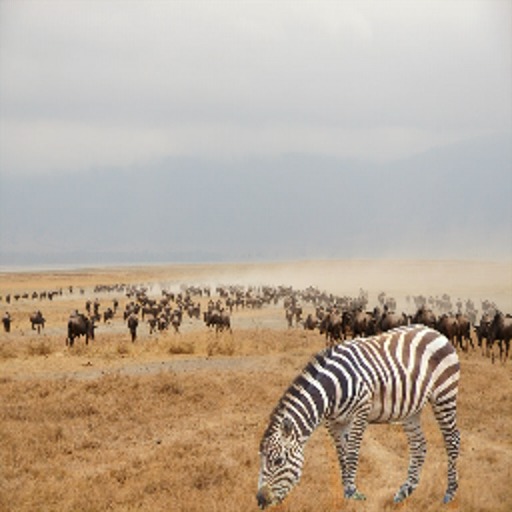}\\

 \end{tabular}
  \caption{COCO classes \emph{Giraffe} and \emph{Zebra}. \emph{First row:} Multiple generated masks conditioned on the background image and bounding box (inset). \emph{Second row:} Our textured results. \emph{Third row:} Deterministic image generation by Hong et al.'s algorithm~\cite{HierarchicalSemanticImageManpulation}. \emph{Fourth row:} FineGAN's whole-image generation~\cite{finegan}. \emph{Fifth row:} Copy and Paste + deep harmonization baselines.
  }
 \label{fig:giraffezebra}
\end{figure*}

\begin{figure*}[tb]

\centering
\newcommand{\tableTitle}[1]{\scriptsize{#1}}%

    \setlength{\imwidth}{0.15\linewidth}%
\setlength{\imwidth}{0.16\linewidth}%

\hfill%
\setlength{\tabcolsep}{0.3pt}%
\setbox1=\hbox{\includegraphics[width=\imwidth]{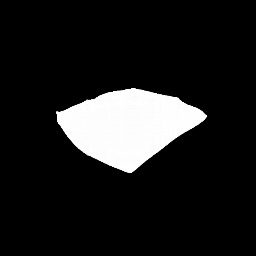}}%
\begin{tabular}{cccccc}


\includegraphics[width=\imwidth]{figures_cvpr/figpizza/m_41_2.jpg}\llap{\makebox[\wd1][l]{\raisebox{0.5\imwidth}{\includegraphics[cfbox=white,width=0.5\imwidth]{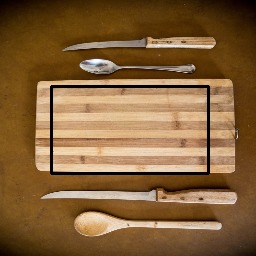}}}}&
\includegraphics[width=\imwidth]{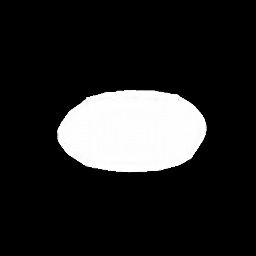} &
\includegraphics[width=\imwidth]{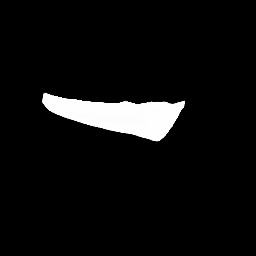}\llap{\makebox[\wd1][l]{\raisebox{0.5\imwidth}{\includegraphics[cfbox=white,width=0.5\imwidth]{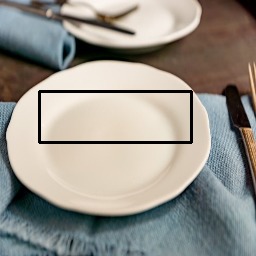}}}}&
\includegraphics[width=\imwidth]{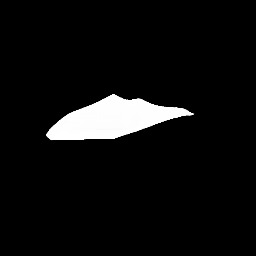}& 
\includegraphics[width=\imwidth]{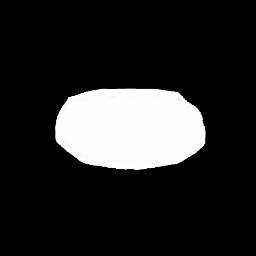}\llap{\makebox[\wd1][l]{\raisebox{0.5\imwidth}{\includegraphics[cfbox=white,width=0.5\imwidth]{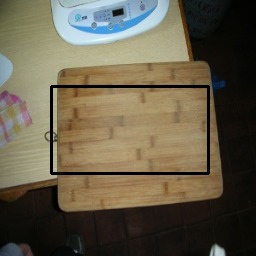}}}}&
\includegraphics[width=\imwidth]{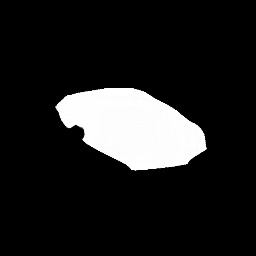} \\


\includegraphics[width=\imwidth]{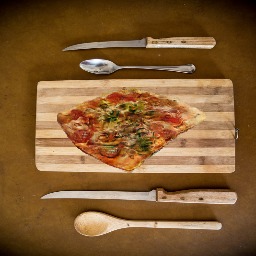}& 
\includegraphics[width=\imwidth]{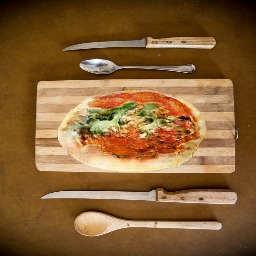}&
\includegraphics[width=\imwidth]{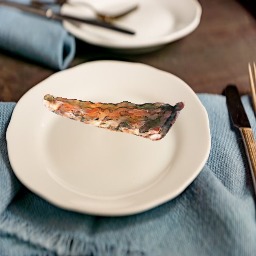}& 
\includegraphics[width=\imwidth]{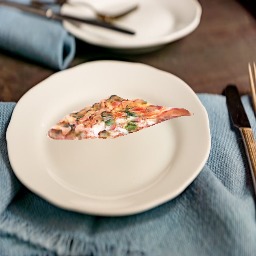}& 
\includegraphics[width=\imwidth]{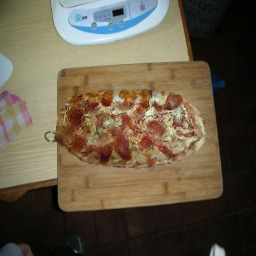}&
\includegraphics[width=\imwidth]{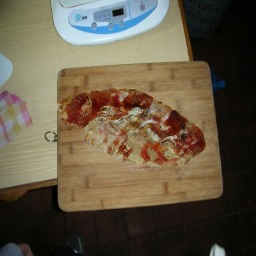} \\

\includegraphics[width=\imwidth]{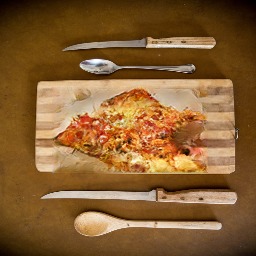}& 
\includegraphics[width=\imwidth]{figures_cvpr/figpizza/im0_template_42.jpg}&
\includegraphics[width=\imwidth]{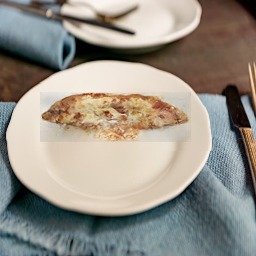}& 
\includegraphics[width=\imwidth]{figures_cvpr/figpizza/im1_template_133.jpg}& 
\includegraphics[width=\imwidth]{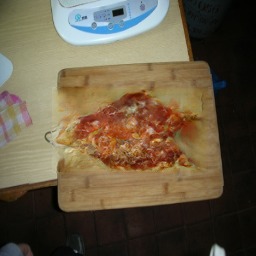}&
\includegraphics[width=\imwidth]{figures_cvpr/figpizza/im2_template_298.jpg} \\

\includegraphics[width=\imwidth]{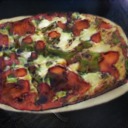}&
\includegraphics[width=\imwidth]{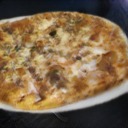}&
\includegraphics[width=\imwidth]{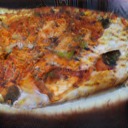}&
\includegraphics[width=\imwidth]{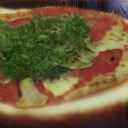}&
\includegraphics[width=\imwidth]{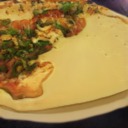}&
\includegraphics[width=\imwidth]{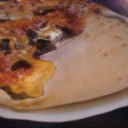}\\

\includegraphics[width=\imwidth]{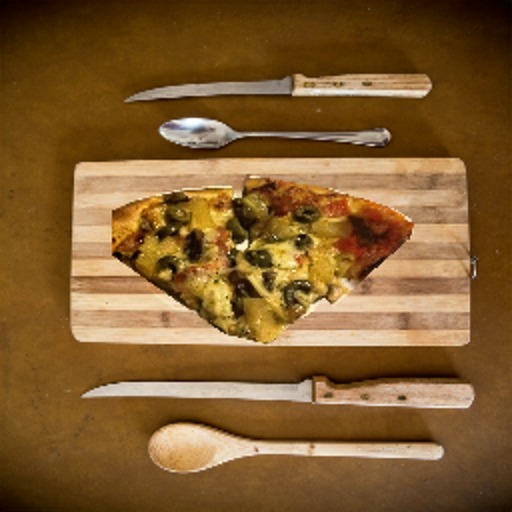}& 
\includegraphics[width=\imwidth]{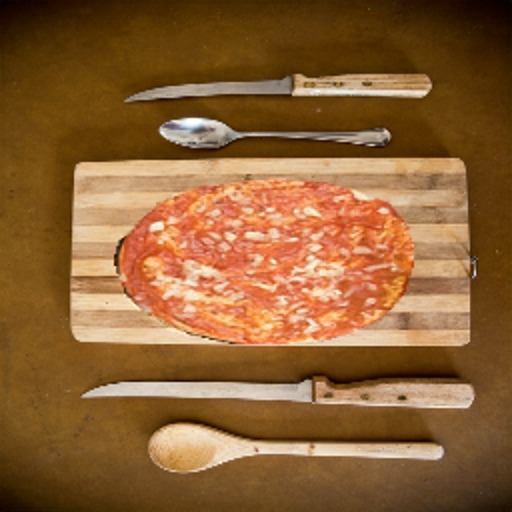}&
\includegraphics[width=\imwidth]{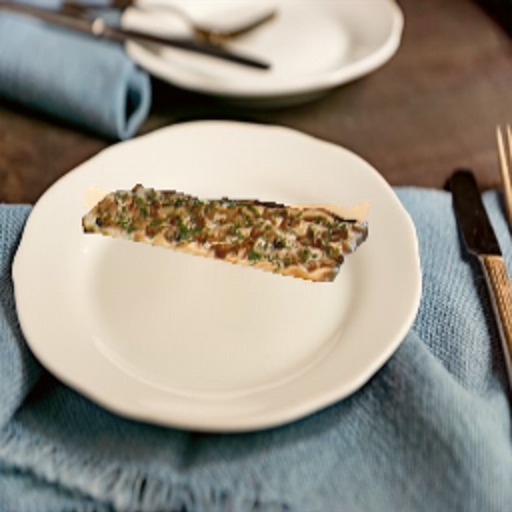}& 
\includegraphics[width=\imwidth]{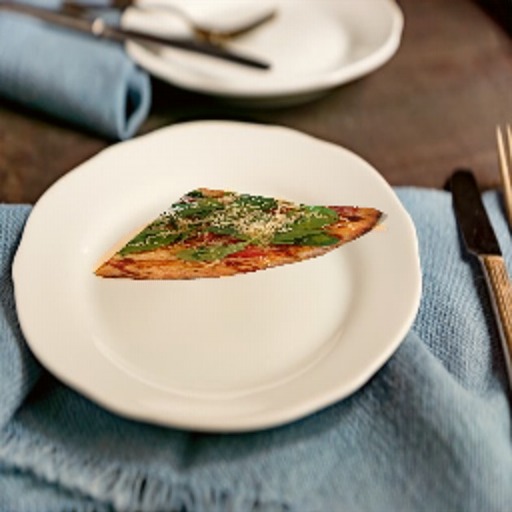}& 
\includegraphics[width=\imwidth]{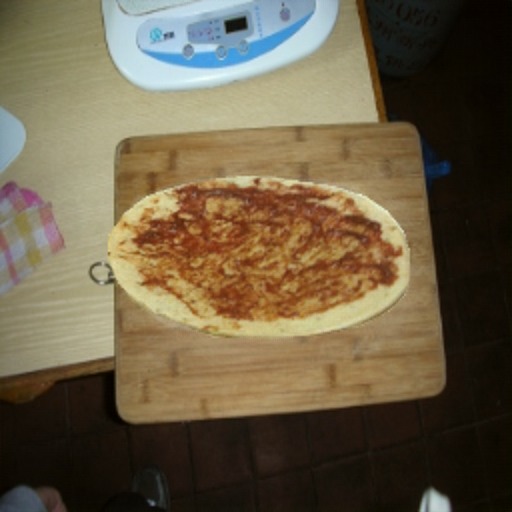}&
\includegraphics[width=\imwidth]{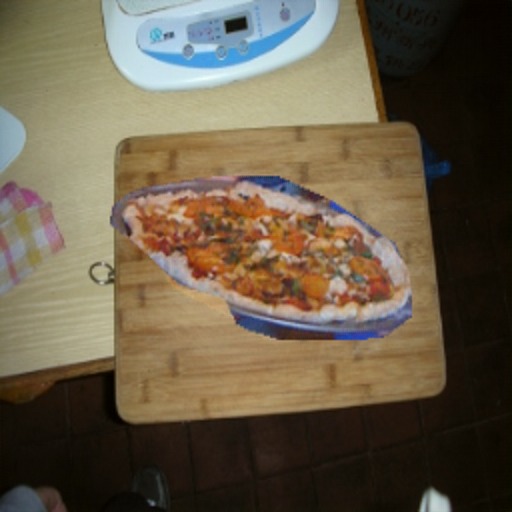} \\
 \end{tabular}\hfill\null
  \caption{COCO class \emph{Pizza}. \emph{First row:} Multiple generated masks conditioned on the background image and bounding box (inset). \emph{Second row:} Our textured results. \emph{Third row:} Deterministic image generation by Hong et al.'s algorithm~\cite{HierarchicalSemanticImageManpulation}. \emph{Fourth row:} FineGAN's whole-image generation~\cite{finegan}. \emph{Fifth row:} Copy and Paste + deep harmonization baselines.
  }
 \label{fig:pizza}
\end{figure*}

\paragraph{Datasets:} 
We extract three classes from COCO: \emph{Giraffe} (2,205 images), \emph{Zebra} (2,306 images), and \emph{Pizza} (2,623 images).
The first two classes have high shape variation, while the last class has high texture variation. 
We exclude all instances that are smaller than 1\% of the entire image, all that contain multiple separate components, and all that intersect any image border.
We collect background images for stamping from the Internet by searching relevant queries under a `free to use and modify' license; we will release these alongside our code.

\paragraph{Training:} We train our shape generation network for 1000 epochs on all datasets, and our texture network for 400 epochs. We use a batch size of 4 and train all our models on 4 GPUs (NVIDIA GTX 1080 Ti).
We use instance normalization in all hidden layers of both generators. 
We use the ADAM optimizer with a learning rate of 0.0002, $\beta_1 = 0.5$, and $\beta_2 = 0.99$. 
We decrease the learning rate linearly towards 0 at half training.
In our loss, all $\lambda$ hyperparameters are fixed to 10, apart from $\lambda_{KL}^T$ which is set to 0.05. We set random vector sizes to $|\z_m|=128$ and $|\z_i|= |\z'_i| =8$. 


\subsection{Method comparisons}
We are not aware of any methods that addresses our exact problem, and so we adapt related methods for comparison.

\paragraph{Semantic image manipulation of Hong et al.~\cite{HierarchicalSemanticImageManpulation}:}
This approach uses multi-class pixel-wise segmentation maps to exploit scene context, from which we extract per-class binary masks.
In this setting, their method produces less distinct shapes and blurrier textures than our approach (Fig.~\ref{fig:giraffezebra}).  
Further, Hong et al.'s algorithm is deterministic: it insert only one instance given a bounding box, whereas our approach can create multiple shapes and textures from the same bounding box.
Finally, Hong et al.~generate content for the entire bounding box, which requires solving the more difficult task of also generating background regions which match seamlessly with the rest of the content (Fig.~\ref{fig:pizza}).

\paragraph{FineGAN~\cite{finegan}:} This whole-image approach decomposes generation into first generating the background and then the foreground. 
It produces convincing results on low variance datasets such as bird images of CUB~\cite{WelinderEtal2010}. 
However, when trained on our classes from COCO, performance tend to decrease (Figs.~\ref{fig:giraffezebra} \& \ref{fig:pizza}), or collapse altogether.
Further, while FineGAN allows for stochastic texture generation, it cannot generate localized instances for user control over the scene layout, \eg, generated foregrounds are often in the center of the image.

\begin{table}[t]
\centering%
\caption{%
Kernel Inception Distance$\times 100 \pm \text{std} \times 100$ for the texture generation on the class Pizza. Lower is better.}
\label{tab:KIDtexture}
\begin{tabular}{lrr}
\toprule
Models & \emph{KID} & \emph{Count}\\
\midrule
Ours & \textbf{-0.046 $\pm$ 0.075} & \textbf{80\%} \\
$-$ Noise & 0.002 $\pm$ 0.084 & 14\% \\
$-$ FM & 0.117 $\pm$ 0.052 & 6\%\\
$-$ VGG & 0.408 $\pm$ 0.157  & 0\%\\
$-$ Bicycle & 0.170 $\pm$ 0.123 & 0\% \\
\bottomrule
\end{tabular}
\end{table}
\begin{table}[t]
\centering%
\caption{%
Kernel Inception Distance$\times 100 \pm \text{std} \times 100$ for the mask generation on the class Giraffe. Lower is better.}
\label{tab:KIDmask}
\begin{tabular}{lrr}
\toprule
Models & \emph{KID} & \emph{Count}\\
\midrule
Ours & \textbf{2.723 $\pm$ 0.405} & \textbf{68\%} \\
Ours-FM &  2.939 $\pm$ 0.369 & 30\%\\
Ours-mrecon & 3.381 $\pm$ 0.410  & 2\%\\
Ours-bgaware & 5.576 $\pm$ 0.525 & 0\% \\
\bottomrule
\end{tabular}
\end{table}\textbf{}
\begin{table}[t]
\centering%
\caption{%
Kernel Inception Distance$\times 100 \pm \text{std} \times 100$ across all datasets used. Lower is better.}
\label{tab:KIDcompare}
\resizebox{1.\linewidth}{!}{
\begin{tabular}{lrrrrrr}
\toprule
Models & \emph{Giraffe} & \emph{count}& \emph{Zebra} & \emph{count} & \emph{Pizza} & \emph{count}  \\
\midrule
SIM & 8.43 $\pm$ 0.59 &  0\% & 6.78 $\pm$ 0.79& 4\% & 9.80 $\pm$ 0.73 & 36\% \\
Ours & \textbf{4.87 $\pm$ 0.45}  &  \textbf{100\%} &  \textbf{5.12 $\pm$ 0.67} & \textbf{96\%} & \textbf{9.35 $\pm$ 0.68} & \textbf{64\%}   \\
\midrule
FineGan & 17.66 $\pm$ 0.86  & 0\% & 12.51 $\pm$ 0.60  & 0\%  & 11.21 $\pm$ 0.29 & 0\%  \\
Ours & \textbf{1.37 $\pm$ 0.24} & \textbf{100\%} & \textbf{1.28 $\pm$ 0.24} & \textbf{100\%} & \textbf{1.76 $\pm$ 0.17} & 100\% \\
\bottomrule
\end{tabular}
}
\end{table}

\begin{figure*}[t]
\centering
{\newcommand{\mysize}{.155\linewidth}
\begin{tabular}{c@{\hspace{2mm}}*{5}{c@{\hspace{1mm}}}}
 Real images & \multicolumn{5}{c}{Varied object retexturing} \\
 \includegraphics[width=\mysize]{figures_cvpr/instanceReplacement/000000173371.jpg} &
 \includegraphics[width=\mysize]{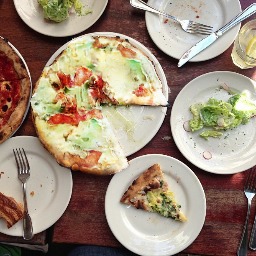} &
 \includegraphics[width=\mysize]{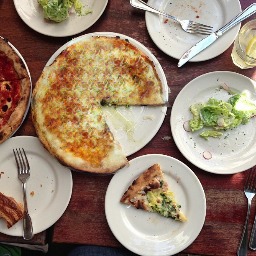} &
 \includegraphics[width=\mysize]{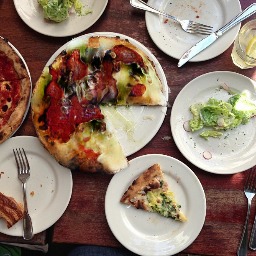} &
 \includegraphics[width=\mysize]{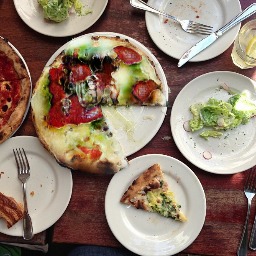} &
 \includegraphics[width=\mysize]{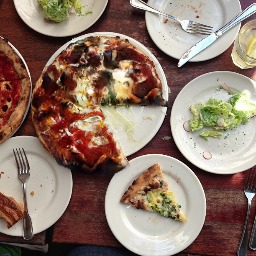}
 \\
 \includegraphics[width=\mysize]{figures_cvpr/instanceReplacement/000000092660.jpg} &
 \includegraphics[width=\mysize]{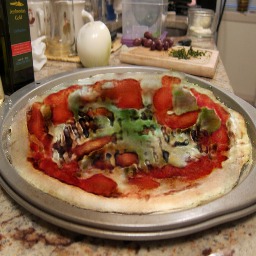} &
 \includegraphics[width=\mysize]{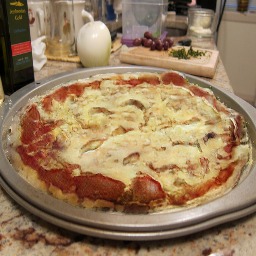} &
 \includegraphics[width=\mysize]{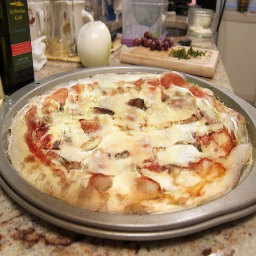} &
  \includegraphics[width=\mysize]{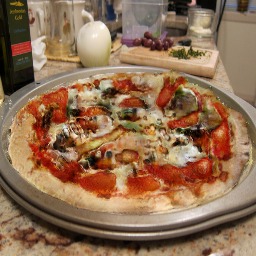} &
  \includegraphics[width=\mysize]{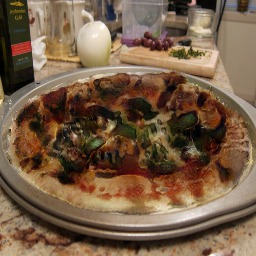}  
  \\
  \includegraphics[width=\mysize]{figures_cvpr/instanceReplacement/000000537053.jpg} &
 \includegraphics[width=\mysize]{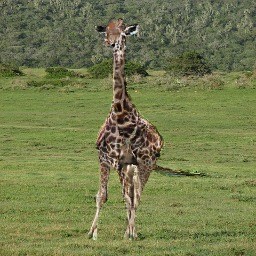} &
 \includegraphics[width=\mysize]{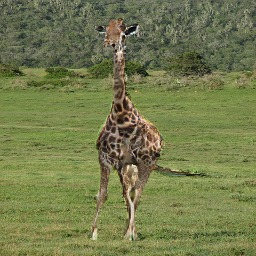} &
 \includegraphics[width=\mysize]{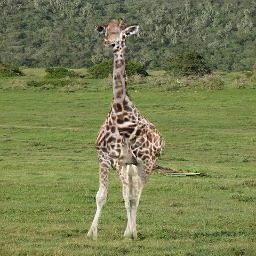} &
 \includegraphics[width=\mysize]{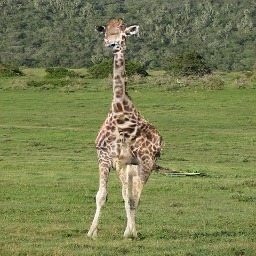} &
 \includegraphics[width=\mysize]{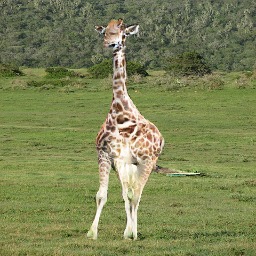}
 \\

\end{tabular}}
 \caption{Results for \emph{object retexturing}. The first column contains the input image, with remaining columns generated by sampling $z_t$.}
  \label{fig:instanceReplacement}
\end{figure*}

\begin{figure*}[t]
\centering
{\newcommand{\mysize}{.155\linewidth}
\begin{tabular}{c@{\hspace{2mm}}*{5}{c@{\hspace{1mm}}}}
 Real masks & \multicolumn{5}{c}{Varied object insertion given mask shape} \\
 \includegraphics[width=\mysize]{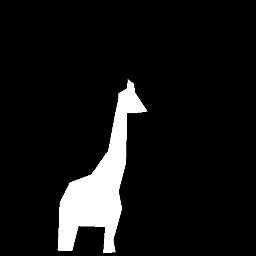} &
 \includegraphics[width=\mysize]{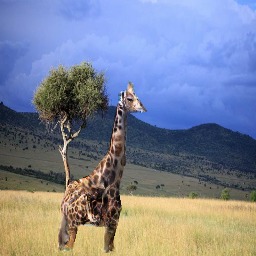} &
 \includegraphics[width=\mysize]{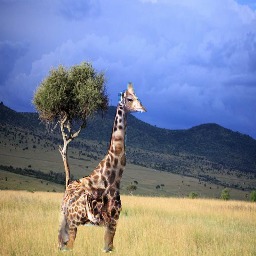} &
 \includegraphics[width=\mysize]{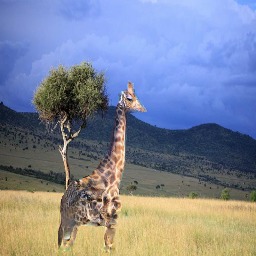} &
 \includegraphics[width=\mysize]{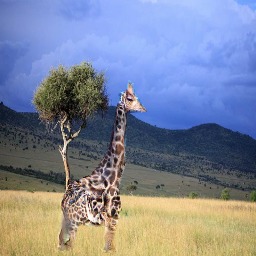} &
 \includegraphics[width=\mysize]{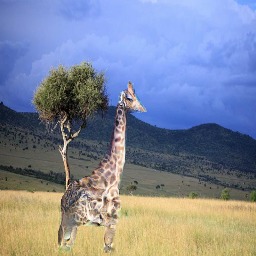}
 \\
 \includegraphics[width=\mysize]{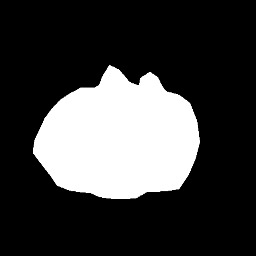} &
 \includegraphics[width=\mysize]{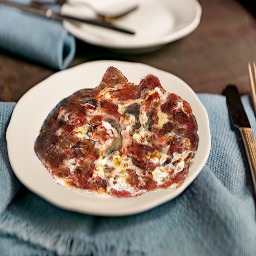} &
 \includegraphics[width=\mysize]{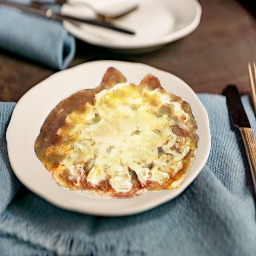} &
 \includegraphics[width=\mysize]{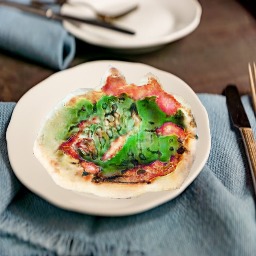} &
 \includegraphics[width=\mysize]{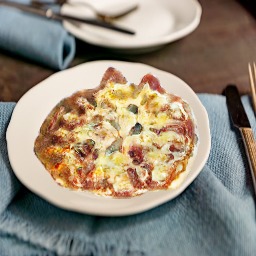} &
 \includegraphics[width=\mysize]{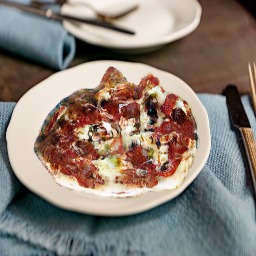}
  \\
  
  \includegraphics[width=\mysize]{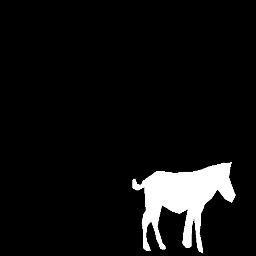} &
 \includegraphics[width=\mysize]{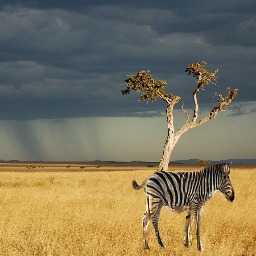} &
 \includegraphics[width=\mysize]{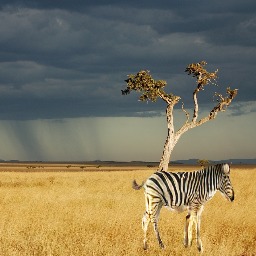} &
 \includegraphics[width=\mysize]{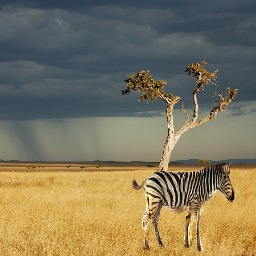} &
 \includegraphics[width=\mysize]{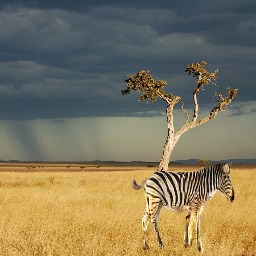} &
 \includegraphics[width=\mysize]{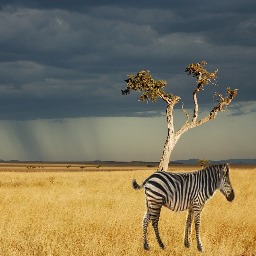}
  \\
\end{tabular}}
 \caption{Results for \emph{object insertion}. The first column contains the input mask image, with remaining columns generated by sampling $z_i$.}
  \label{fig:instanceInsertion}
\end{figure*}

\paragraph{Copy and Paste + Deep Harmonization~\cite{tsaiCVPR2017}:}  
We also compare with a copy and paste baseline. First, we find masks in the training set that most resemble generated masks from our algorithm.
Rather than a simple nearest neighbor search, we crop each training mask and re-scale it to fit the size of our generated mask. Then, and as is usual for binary data, we compute the cosine distance between the masks and select the training set mask with the smallest distance. 
Second, to make this baseline more realistic, we use a state of the art harmonization approach~\cite{tsaiCVPR2017}. 
Results in Figures \ref{fig:giraffezebra} and \ref{fig:pizza} show that this baseline often struggles to form a convincing composite. 
Furthermore, this approach has three other limitations: 1) Without using our model's generated mask to query the database, the user would be required to provide an actual mask rather than a bounding box; 2) It requires user time to search through the training dataset; and 3) It is not able to disentangle shape from texture.

\subsection{Additional Applications} 

Beyond \emph{object stamps}, our algorithm can \emph{retexture} an existing object instance in an image (Fig.~\ref{fig:instanceReplacement}).
With these `ground truth' masks, our algorithm generates realistic textures that blend with their respective backgrounds, \eg, the sides of the generated pizzas contain doughy edges, as real pizza has.
Alternately, our method supports applications where a user would like to insert an object of a specific shape into a given background, and have it be textured realistically (Fig~\ref{fig:instanceInsertion}). 

\subsection{Quantitative Evaluation}

We use KID~\cite{bisuargr18} as an evaluation metric as it has been shown to give more consistent results than Frechet Inception Distance (FID), especially with small numbers of generated samples~\cite{heusel2017gans}.

\paragraph{Semantic image manipulation of Hong et al.~\cite{HierarchicalSemanticImageManpulation}:} As this approach is deterministic, we generate one instance per bounding box and per scene. We do the same for our method using the same bounding boxes and background images. We report KID results on 50 random subsamples in Table~\ref{tab:KIDcompare}. Our approach consistently performs better, which agrees with the visual results (Fig.~\ref{fig:giraffezebra}). 

\paragraph{FineGAN~\cite{finegan}:} We trained FineGAN on the COCO classes using our image resolution target (256$\times$256), which is higher than used in their paper. At this increased resolution, FineGAN mode collapsed, and so we used their original resolution (128$\times$128) and bilinearly upsample the generated images to 256$\times$256.

FineGAN is a whole image generation approach, and so a comparison using our composited foregrounds would not be fair. Instead, we compute KID using only generated foregrounds without compositing them into the backgrounds.
Table~\ref{tab:KIDcompare} shows the corresponding scores. FineGAN produces substantially worse KID scores than our method (Figure~\ref{fig:giraffezebra}). Running the algorithm on classes such as `giraffe' and `zebra' results in a significant reduction in generated diversity in that the same instance is generated most of the time.
Further, resizing the images to 256x256 due to mode collapses in training at higher resolutions causes a lack of fine detail that is captured by KID. Finally, some foreground generations are not successful (fourth column).



\paragraph{Ablation studies:}

First, we evaluate which components of our method contribute to the quality of texture generation (Tab~\ref{tab:KIDtexture}). 
We generate a set of 10 different synthesized foregrounds for each instance mask in the validation set.
Then, we create 50 random subsets of real and generated images, and compute KID scores.
We perform this evaluation on the `pizza' class due to the high texture variety. 
Our full method performs best; removing the noise input (Noise) causes a small reduction; removing the feature matching loss (FM, Eq.~\ref{eq:Loss_T_FM}) causes a larger reduction.
Removing the bicycle loss yields realistic results but without diversity, which accounts for the higher KID score, and removing the VGG feature loss has the largest impact on result quality. 

We similarly ablate our mask generation; in this experiment, we use the `giraffe' class due to the high shape variation. We generate 10 masks per bounding box in the validation set, then we construct 50 subsets to use for KID computation. Results are displayed in Table~\ref{tab:KIDmask}. First, we remove the contribution of the feature matching loss in Eq.~\ref{eq:Loss_M_FM} and notice a score increase indicating that the FM loss helps in this task. Second, we try to reconstruct $\z_m$ from the generated mask directly instead from the predicted AdaIN parameters. In this setting (`Ours-mrecon'), our generator ignores the latent vector $\z_m$ which limits shape diversity and so increases KID values. 
Finally, we condition the mask discriminator using $s$ on top of $\m$ and $b$ (`Ours-bgcond'), but we did not notice any significant improvement (also discussed in \cite{zhu2017bicycle, pathak2016context}).

The high standard deviations in Tables~\ref{tab:KIDtexture},~\ref{tab:KIDmask}, and \ref{tab:KIDcompare} are caused by the variation in images across the 50 subsets.
As such, in each table, we compute as `Count' the number of times across subsets that each approach achieves the best KID score (shown as a percent). Our approach achieves the highest count.

\section{Discussion}

Our method takes one step in the direction of easy to use and diverse guided image creation. 
However, this is a challenging problem, and there are still a number of limitations to this approach (see failure case figure in the appendix). 

First, we note that the COCO dataset is especially challenging, including issues with its instance annotations. 
For example, some instances are occluded, which causes holes in the masks. As such, sampling from the learned mask generator can create shapes with irregularities.
Some instances also only show parts of the objects, e.g., only the head of the giraffe (Fig.~\ref{fig:cocodiversity}), which can lead to some unexpected masks at test time (Fig.~\ref{fig:failures}).

Finally, with datasets containing complicated and diverse textures, texture generation can also be challenging (see the class \emph{Bus} in Fig.~\ref{fig:fig_bus_supp} in the appendix), where complex structured appearance with transparency through the windows is difficult to model.
For less complex classes, our approach allows users to sample the generator to obtain desirable object shapes and textures.


\section{Conclusion}

Inspired by curriculum learning, we have presented a method to generate object stamps by splitting the process into generating an object mask and object texture. Our approach carefully considers how to promote both detail and diverse generation in both shapes and texture, through feature matching, perceptual, and auxiliary losses via conditioning vector in encoder and decoders.
Further, we condition generation on background content with a way to train our approach on pre-existing segmentation datasets which \emph{include} object instances already but do not confuse the generator. Put together, this provides flexible generation across shape and texture via a simple bounding box interface.

\paragraph{Acknowledgements:} Youssef~A.~Mejjati thanks the European Union’s Horizon 2020 research and innovation program under 
the Marie Sklodowska-Curie grant agreement No 665992, and the UK’s EPSRC Center for Doctoral Training in Digital Entertainment (CDE), EP/L016540/1. James Tompkin and Kwang In Kim thank gifts from Adobe.

{\small
\bibliographystyle{plain}
\bibliography{bib}

\begin{thebibliography}{10}

\bibitem{agarwala2004interactive}
A.~Agarwala, M.~Dontcheva, M.~Agrawala, S.~Drucker, A.~Colburn, B.~Curless,
  D.~Salesin, and M.~Cohen.
\newblock Interactive digital photomontage.
\newblock {\em ACM Trans. Graphics (ToG)}, 23(3):294--302, 2004.

\bibitem{bisuargr18}
M.~Bi{\'n}kowski, D.~Sutherland, M.~Arbel, and A.~Gretton.
\newblock Demystifying {MMD GANs}.
\newblock In {\em ICLR}, 2018.

\bibitem{chu2017cyclesteg}
Casey Chu, Andrey Zhmoginov, and Mark Sandler.
\newblock Cyclegan, a master of steganography.
\newblock In {\em NeurIPS, workshop on Machine Deception}, 2017.

\bibitem{Geirhos2018}
R.~Geirhos, P.~Rubisch, C.~Michaelis, M.~Bethge, F.~A. Wichmann, and
  W.~Brendel.
\newblock {ImageNet-trained CNNs} are biased towards texture; increasing shape
  bias improves accuracy and robustness.
\newblock In {\em ICLR}, 2019.

\bibitem{Gokaslan2018}
Aaron Gokaslan, Vivek Ramanujan, Daniel Ritchie, Kwang~In Kim, and James
  Tompkin.
\newblock Improved shape deformation in unsupervised image to image
  translation.
\newblock In {\em ECCV}, 2018.

\bibitem{goodfellow2014generative}
I.~Goodfellow, J.~Pouget-Abadie, M.~Mirza, B.~Xu, D.~Warde-Farley, S.~Ozair,
  A.~Courville, and Y.~Bengio.
\newblock Generative adversarial nets.
\newblock In {\em NeuIPS}, 2014.

\bibitem{heusel2017gans}
M.~Heusel, H.~Ramsauer, T.~Unterthiner, B.~Nessler, and S.~Klambauer.
\newblock {GANs} trained by a two time-scale update rule converge to a {Nash}
  equilibrium.
\newblock In {\em NeuIPS}, 2017.

\bibitem{HierarchicalSemanticImageManpulation}
S.~Hong, X.~Yan, T.~Huang, and H.~Lee.
\newblock Learning hierarchical semantic image manipulation through structured
  representations.
\newblock In {\em NeurIPS}, 2018.

\bibitem{huang2017adain}
X.~Huang and S.~Belongie.
\newblock Arbitrary style transfer in real-time with adaptive instance
  normalization.
\newblock In {\em ICCV}, 2017.

\bibitem{multimodal2018}
X.~Huang, M.~Liu, S.~Belongie, and J.~Kautz.
\newblock Multimodal unsupervised image-to-image translation.
\newblock In {\em ECCV}, 2018.

\bibitem{johnson2016perceptual}
Justin Johnson, Alexandre Alahi, and Li~Fei-Fei.
\newblock Perceptual losses for real-time style transfer and super-resolution.
\newblock In {\em ECCV}, 2016.

\bibitem{karras2018style}
T.~Karras, S.~Laine, and T.~Aila.
\newblock A style-based generator architecture for generative adversarial
  networks.
\newblock {\em CVPR}, 2019.

\bibitem{kim2017learning}
T.~Kim, M.~Cha, H.~Kim, J.~Lee, and J.~Kim.
\newblock Learning to discover cross-domain relations with generative
  adversarial networks.
\newblock {\em JMLR}, 2017.

\bibitem{kingma2013auto}
Diederik~P Kingma and Max Welling.
\newblock Auto-encoding variational bayes.
\newblock {\em ICLR}, 2014.

\bibitem{kwak2016generating}
H.~Kwak and B.-T. Zhang.
\newblock Generating images part by part with composite generative adversarial
  networks.
\newblock {\em arXiv preprint arXiv:1607.05387}, 2016.

\bibitem{lim2017geometric}
J.~H. Lim and J.~C. Ye.
\newblock Geometric {GAN}.
\newblock {\em arXiv preprint arXiv:1705.02894}, 2017.

\bibitem{lin2018st}
C.-H. Lin, E.~Yumer, O.~Wang, E.~Shechtman, and S.~Lucey.
\newblock {ST-GAN}: Spatial transformer generative adversarial networks for
  image compositing.
\newblock In {\em CVPR}, 2018.

\bibitem{MSCOCO2014}
T.-Y. Lin, M.~Maire, S.~J. Belongie, L.~D. Bourdev, R.~B. Girshick, J.~Hays,
  P.~Perona, D.~Ramanan, P.~Doll{\'{a}}r, and C.~L. Zitnick.
\newblock Microsoft {COCO:} common objects in context.
\newblock In {\em ECCV}, 2014.

\bibitem{mejjati2018unsupervised}
Youssef~A Mejjati, Christian Richardt, James Tompkin, Darren Cosker, and
  Kwang~In Kim.
\newblock Unsupervised attention-guided image to image translation.
\newblock In {\em NeurIPS}, 2018.

\bibitem{miyato2018spectral}
T.~Miyato, T.~Kataoka, M.~Koyama, and Y.~Yoshida.
\newblock Spectral normalization for generative adversarial networks.
\newblock 2018.

\bibitem{ostyakov2018seigan}
P.~Ostyakov, R.~Suvorov, E.~Logacheva, O.~Khomenko, and S.~I. Nikolenko.
\newblock {SEIGAN}: Towards compositional image generation by simultaneously
  learning to segment, enhance, and inpaint.
\newblock {\em arXiv preprint arXiv:1811.07630}, 2018.

\bibitem{park2019SPADE}
T.~Park, M.-Y. Liu, T.-C. Wang, and J.-Y. Zhu.
\newblock Semantic image synthesis with spatially-adaptive normalization.
\newblock In {\em CVPR}, 2019.

\bibitem{pathak2016context}
Deepak Pathak, Philipp Krahenbuhl, Jeff Donahue, Trevor Darrell, and Alexei~A
  Efros.
\newblock Context encoders: Feature learning by inpainting.
\newblock In {\em CVPR}, 2016.

\bibitem{perez2003poisson}
P.~P{\'e}rez, M.~Gangnet, and A.~Blake.
\newblock Poisson image editing.
\newblock {\em TOG}, 2003.

\bibitem{salimans2016improved}
T.~Salimans, I.~Goodfellow, W.~Zaremba, V.~Cheung, A.~Radford, and X.~Chen.
\newblock Improved techniques for training {GANs}.
\newblock In {\em NeuIPS}, 2016.

\bibitem{finegan}
K.~K. Singh, U.~Ojha, and Y.~J. Lee.
\newblock Finegan: Unsupervised hierarchical disentanglement for fine-grained
  object generation and discovery.
\newblock In {\em CVPR}, 2019.

\bibitem{tran2017deep}
D.~Tran, R.~Ranganath, and D.~M. Blei.
\newblock Deep and hierarchical implicit models.
\newblock {\em arXiv preprint arXiv:1702.08896}, 7, 2017.

\bibitem{tsaiCVPR2017}
Y.-H. Tsai, X.~Shen, Z.~Lin, K.~Sunkavalli, X.~Lu, and M.-H. Yang.
\newblock Deep image harmonization.
\newblock In {\em CVPR}, 2017.

\bibitem{WelinderEtal2010}
P.~Welinder, S.~Branson, T.~Mita, C.~Wah, F.~Schroff, S.~Belongie, and
  P.~Perona.
\newblock {Caltech-UCSD Birds 200}.
\newblock Technical Report CNS-TR-2010-001, California Institute of Technology,
  2010.

\bibitem{wu2019transgaga}
Wayne Wu, Kaidi Cao, Cheng Li, Chen Qian, and Chen~Change Loy.
\newblock Transgaga: Geometry-aware unsupervised image-to-image translation.
\newblock In {\em CVPR}, 2019.

\bibitem{yang2017lr}
J.~Yang, A.~Kannan, D.~Batra, and D.~Parikh.
\newblock {LR-GAN}: Layered recursive generative adversarial networks for image
  generation.
\newblock In {\em ICLR}, 2017.

\bibitem{zhan2019adaptive}
F.~Zhan, J.~Huang, and S.~Lu.
\newblock Adaptive composition {GAN} towards realistic image synthesis.
\newblock {\em arXiv preprint arXiv:1905.04693}, 2019.

\bibitem{zhang2018unreasonable}
R.~Zhang, P.~Isola, A.~A. Efros, E.~Shechtman, and O.~Wang.
\newblock The unreasonable effectiveness of deep features as a perceptual
  metric.
\newblock In {\em CVPR}, pages 586--595, 2018.

\bibitem{zhu2017unpaired}
J.~Zhu, T.~Park, P.~Isola, and A.~Efros.
\newblock Unpaired image-to-image translation using cycle-consistent
  adversarial networks.
\newblock In {\em ICCV}, 2017.

\bibitem{zhu2017bicycle}
J.-Y. Zhu, R.~Zhang, D.~Pathak, T.~Darrell, A.~A. Efros, O.~Wang, and
  E.~Shechtman.
\newblock Toward multimodal image-to-image translation.
\newblock In {\em NeuIPS}, 2017.

\bibitem{zhu2017toward}
Jun-Yan Zhu, Richard Zhang, Deepak Pathak, Trevor Darrell, Alexei~A Efros,
  Oliver Wang, and Eli Shechtman.
\newblock Toward multimodal image-to-image translation.
\newblock In {\em NeurIPS}, 2017.

\end{thebibliography}
}

\clearpage
\appendix
\section*{Appendix}

\section{Diversity of instances in the COCO dataset}
We train our generator on images `in the wild' using the diverse COCO dataset. This dataset is complex: object instances are at multiple sizes, and are not necessarily centered in the image. Object instances are under occlusion from the edge of the image, for example, when only the head of a giraffe is visible (Figure~\ref{fig:cocodiversity}). Further, some object instances are severely occluded by other image content. For example, in the second image in the first row of Figure~\ref{fig:cocodiversity}, we see only the body of a baby giraffe. Finally, the dataset often contains incorrect annotations, such as the crepe and the pancakes in the last row of Figure~\ref{fig:cocodiversity}, which are annotated as being from the class `pizza'.

\newcommand{\mysizew}{.23\linewidth}
\newcommand{\mysizeh}{.18\linewidth}
\setlength{\tabcolsep}{1pt}
\begin{table*}[t]
  \centering
\begin{tabular}{cccc}
 \includegraphics[width=\mysizew, height=\mysizeh]{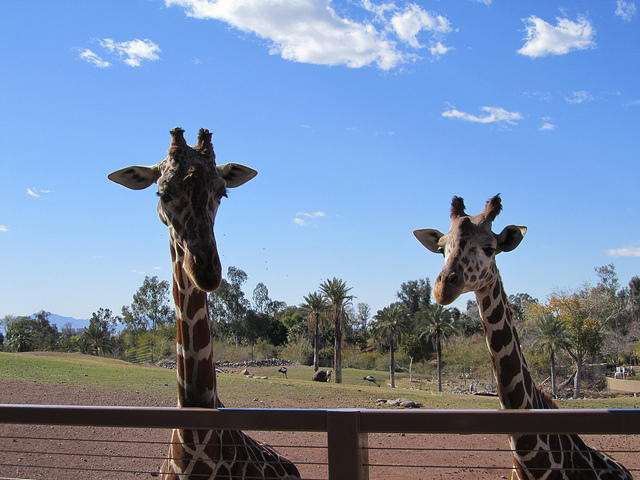} &
 \includegraphics[width=\mysizew, height=\mysizeh]{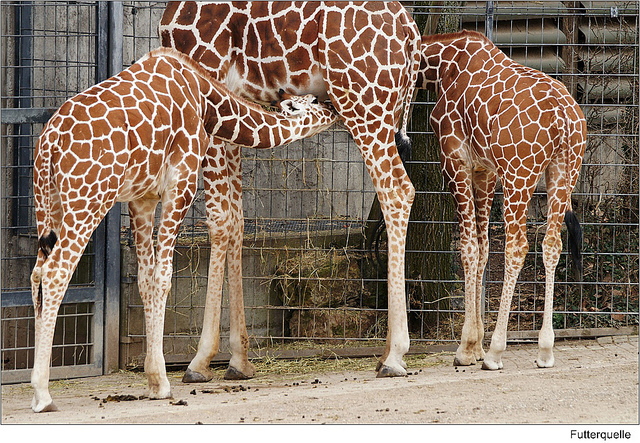} &
  \includegraphics[width=\mysizew, height=\mysizeh]{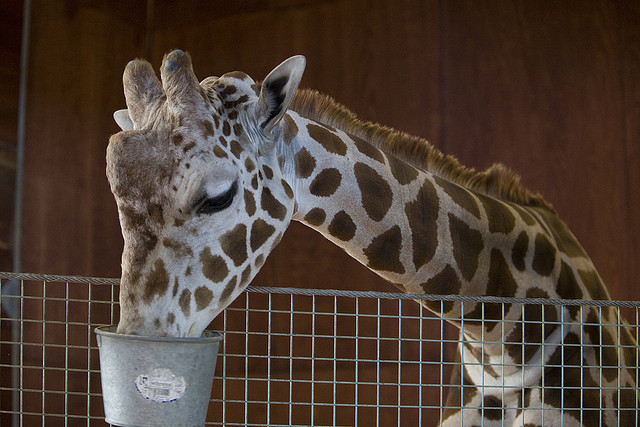} &
  \includegraphics[width=\mysizew, height=\mysizeh]{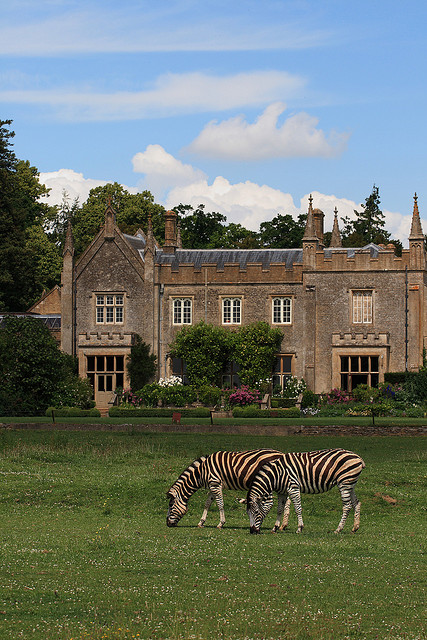} \\
 
  \includegraphics[width=\mysizew, height=\mysizeh]{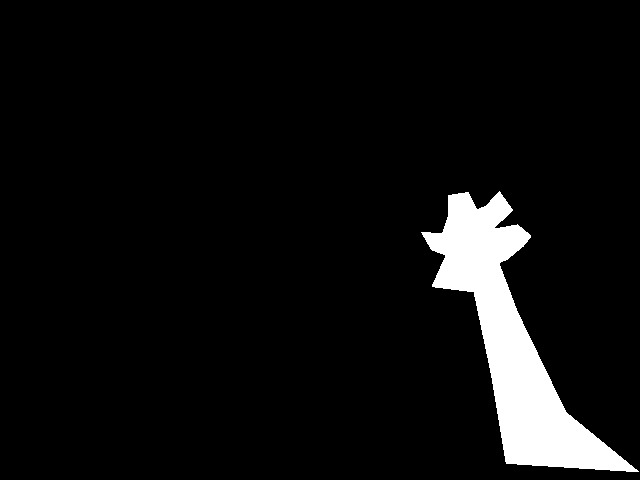} &
 \includegraphics[width=\mysizew, height=\mysizeh]{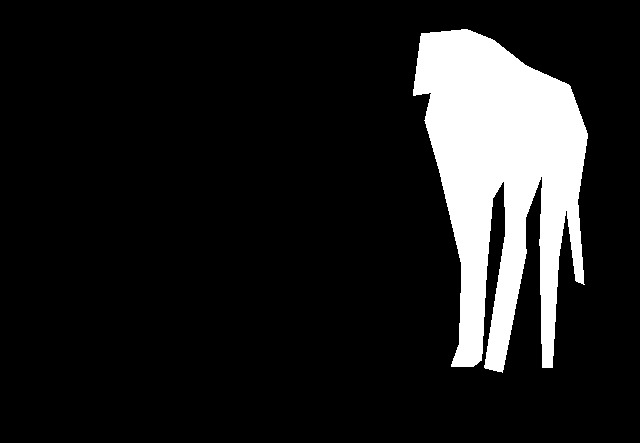} &
  \includegraphics[width=\mysizew, height=\mysizeh]{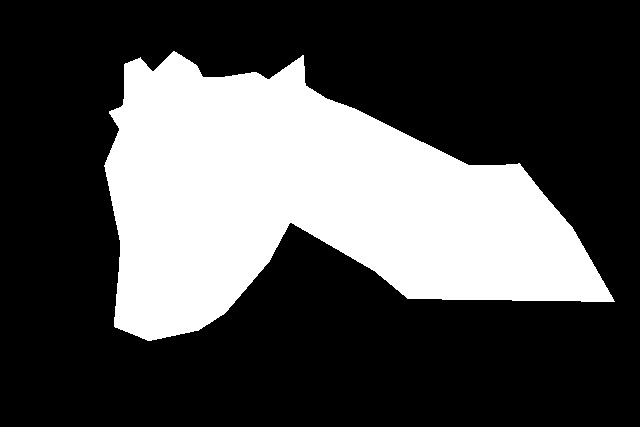} &
    \includegraphics[width=\mysizew, height=\mysizeh]{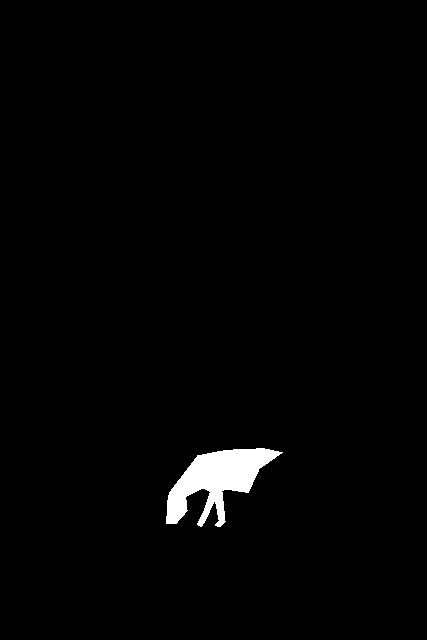}\\

  \includegraphics[width=\mysizew, height=\mysizeh]{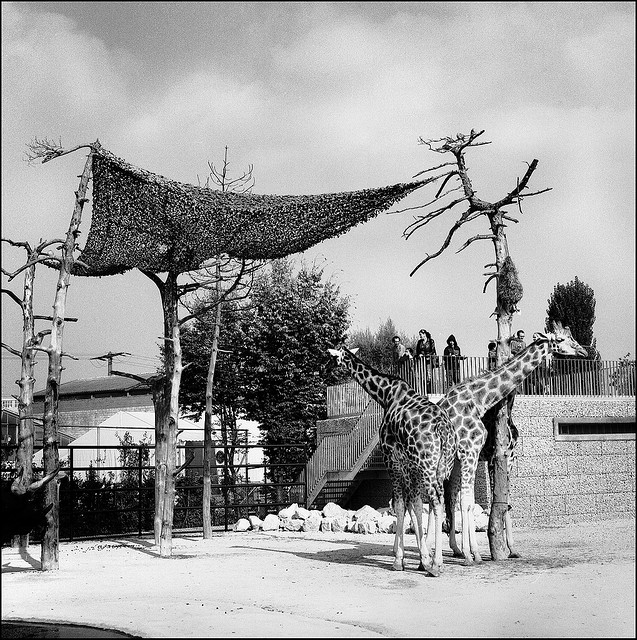} &
 \includegraphics[width=\mysizew, height=\mysizeh]{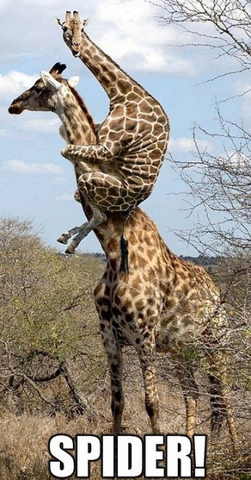} &
  \includegraphics[width=\mysizew, height=\mysizeh]{figures_cvpr/cocodiversity/000000559051.jpg} &
  \includegraphics[width=\mysizew, height=\mysizeh]{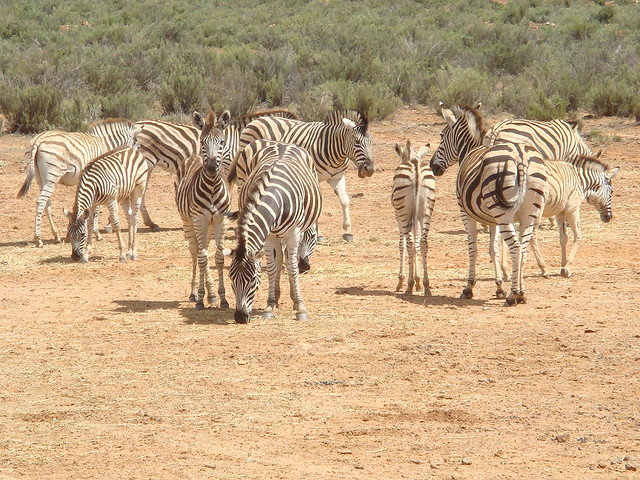} \\
 
  \includegraphics[width=\mysizew, height=\mysizeh]{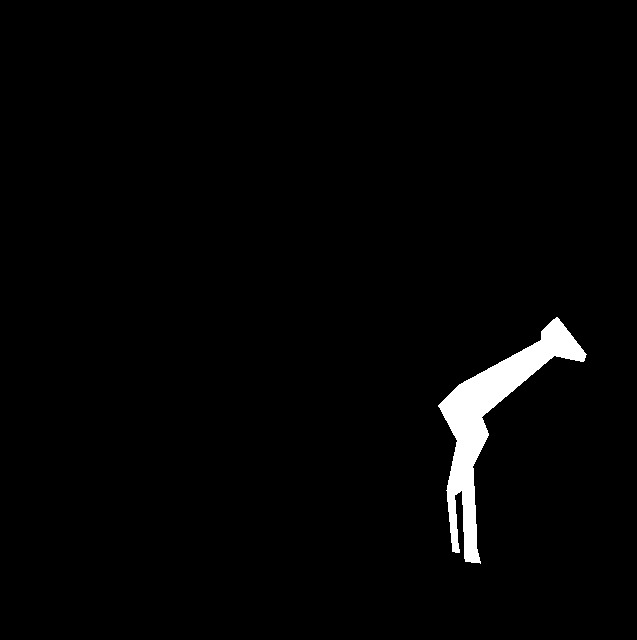} &
 \includegraphics[width=\mysizew, height=\mysizeh]{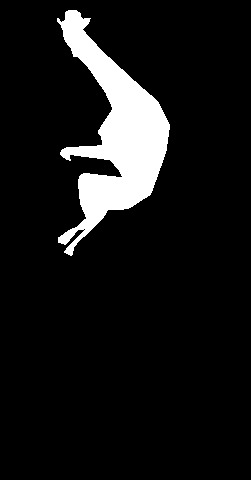} &
  \includegraphics[width=\mysizew, height=\mysizeh]{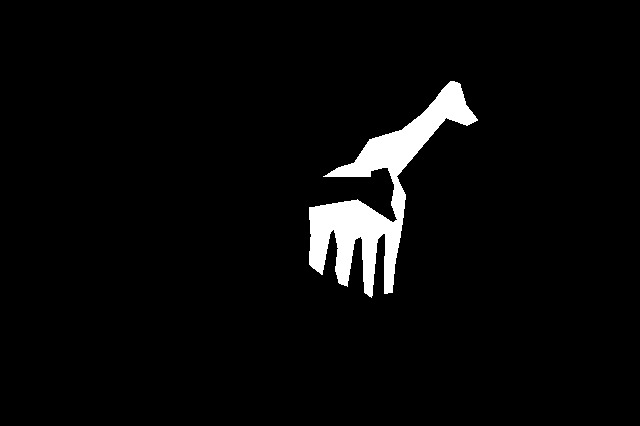} &
     \includegraphics[width=\mysizew, height=\mysizeh]{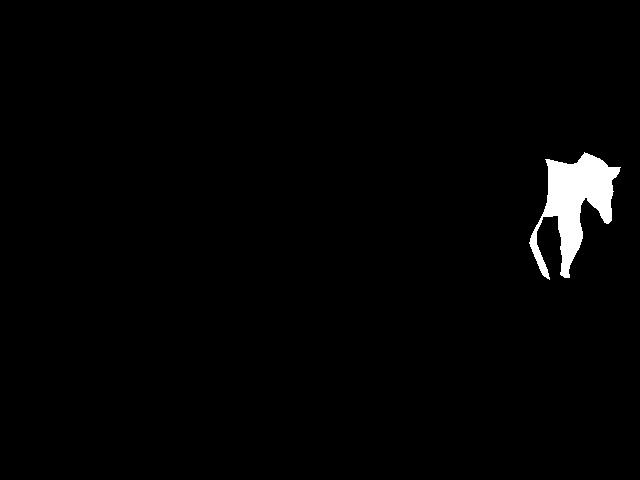} \\

  \includegraphics[width=\mysizew, height=\mysizeh]{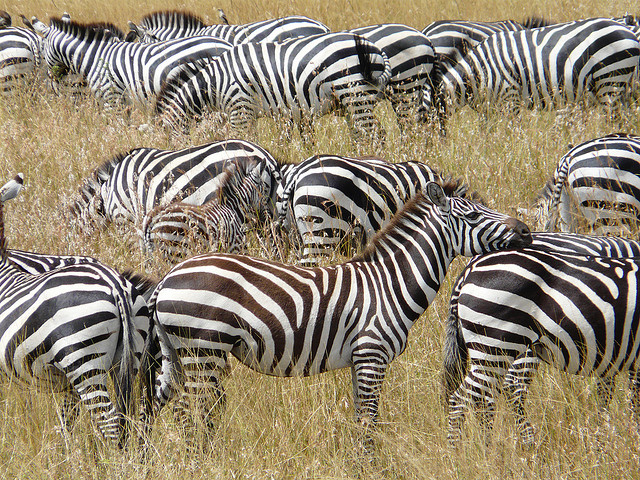} &
 \includegraphics[width=\mysizew, height=\mysizeh]{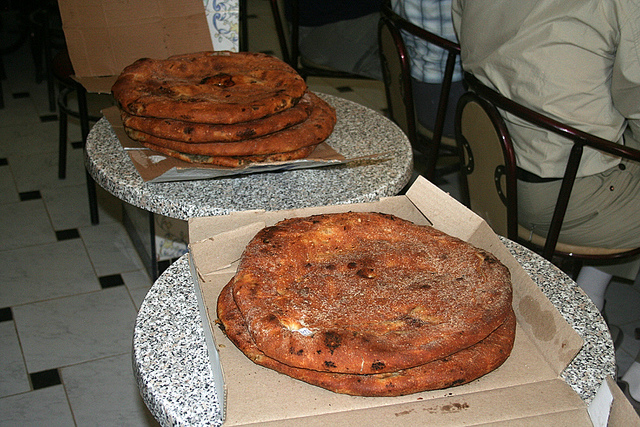} &
  \includegraphics[width=\mysizew, height=\mysizeh]{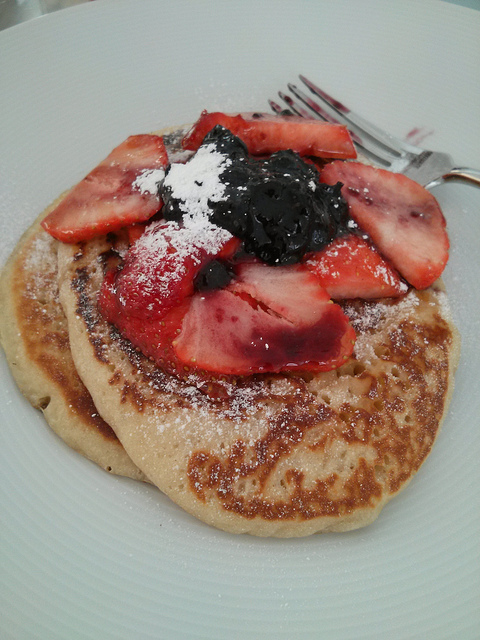} &
     \includegraphics[width=\mysizew, height=\mysizeh]{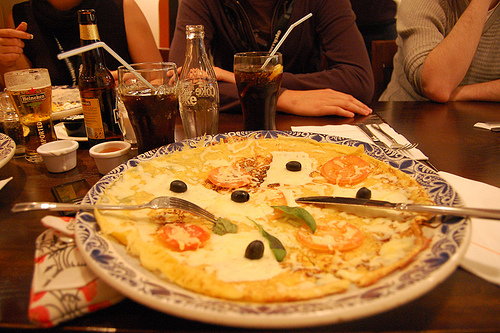} \\
     
        \includegraphics[width=\mysizew, height=\mysizeh]{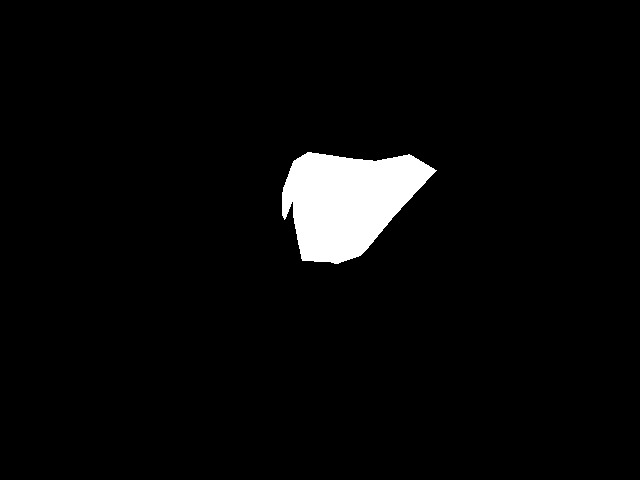} &
 \includegraphics[width=\mysizew, height=\mysizeh]{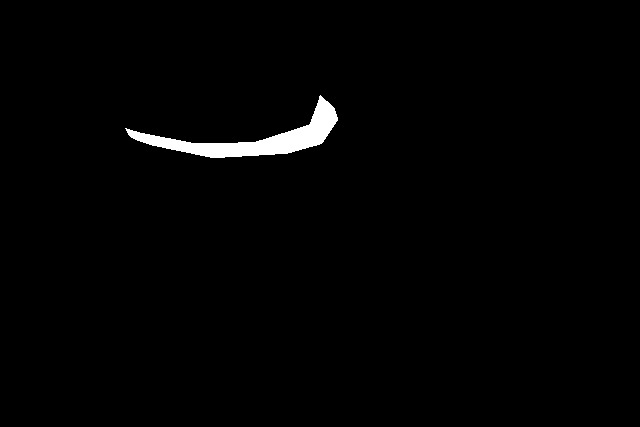} &
  \includegraphics[width=\mysizew, height=\mysizeh]{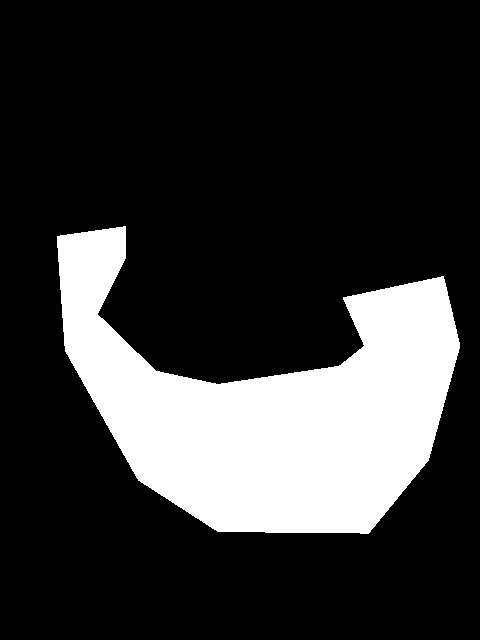} &
     \includegraphics[width=\mysizew, height=\mysizeh]{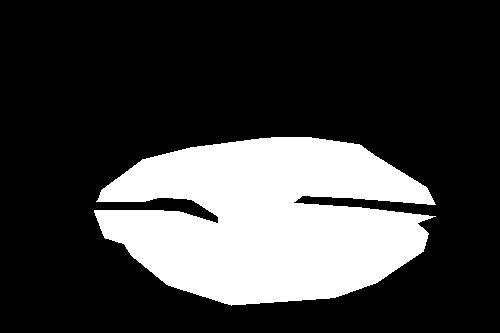} \\

\end{tabular}

 \captionof{figure}{Complexity of COCO dataset. COCO contains images `in the wild' which are severely occluded and noisy, e.g., occluded zebras and giraffes in the first and second rows. The dataset also contains unusual poses, such as in the third row, second column example. Noise also comes from mistakes in the annotation itself. For example, the pancakes and the crepe in the last row are mistakenly annotated as pizzas (on top of being occluded). All these issues make our task of generating realistic shapes more challenging.}
  \label{fig:cocodiversity}
 \end{table*}

\section{Interpolations}

\paragraph{Background brightness} Figure~\ref{fig:fig_change_gamma_gir_supp} shows how varying the background brightness influences the texture generation: our generator is conditioned upon the background image, and so smoothly adapts the foreground brightness to fit the background illumination. In this experiment, we fix the shape mask and texture latent codes.

\paragraph{Texture latent code} 
Figures~\ref{fig:fig_instance_interpolation_pizza_supp} and \ref{fig:fig_instance_interpolation_gir_supp} show interpolations using $\mathbf{z}_t$ while fixing the mask. The first and last images in each figure are the interpolation extremities, with the interpolation proceeding in Western order. Our network learns a smooth texture space, e.g., smoothly varying the pizza toppings in Figure~\ref{fig:fig_instance_interpolation_pizza_supp}, and gradually adding shadows on the left side of the giraffe in Figure~\ref{fig:fig_instance_interpolation_gir_supp}.

\paragraph{Shape mask latent code} 
Figures~\ref{fig:fig_mask_interpolation_zebra_supp} and \ref{fig:fig_mask_interpolation_gir_supp} show interpolations of $\mathbf{z}_m$ while fixing the texture latent code $\mathbf{z}_t$. Again, the first and last images in each figure are the interpolation extremities, with the interpolation proceeding in Western order. As perhaps expected given the high variability of shapes/masks in COCO, our network learns a less smooth shape space compared to the texture space. Further, with the bounding box size fixed, there are no guarantees that the intermediate generated images remain realistic, as shown on the second row of Figure~\ref{fig:fig_mask_interpolation_gir_supp}; e.g., a true `object rotation' would require the bounding box to change shape.

\section{Additional results}
We present additional \emph{object stamp} application results for the classes `giraffe', `zebra', and `pizza' in Figures~\ref{fig:fig_giraffe_supp}--\ref{fig:fig_pizza_supp}. Further, we present additional \emph{object re-texturing} application results for the classes `giraffe', `pizza' and `donut' in Figures~\ref{fig:fig_giraffe_retexturing_supp}--\ref{fig:fig_donut_retexturing_supp}.

\section{Failure cases}
 
Learning with such noisy data makes our shape and texture generation task more challenging. As a result, we obtain several failure cases. We highlight these in Figure~\ref{fig:failures}.

Column 1 shows failure cases due to occlusions in the training set. 
Column 2 shows failure cases 
where there is a mismatch between the foreground and background lighting. In settings with user control, both occlusion and lighting mismatch failure cases can be avoided by re-sampling from the generator.
Column 3 shows how our algorithm 
fails to generate fine details, for example the heads of the giraffe and zebra lack details and so are unrealistic. This problem is challenging to overcome as it requires that our algorithm learns the notion of animal `parts' at multiple scales under occlusion. 

Columns 4 and 5 show failure cases due to the input bounding box shape: Column 4 shows that if the bounding box is too large, then this can lead to unrealistic shapes as the contribution of the background in the shape generation stage is reduced. Column 5 shows that requesting unusual bounding boxes can also lead to unrealistic shapes, e.g., small but wide bounding boxes for giraffes, and tall but thin bounding boxes for zebra. This is somewhat expected as such bounding box shapes are rarer in the corresponding training sets.

An additional interesting failure case can be found in the last row of Figure~\ref{fig:fig_giraffe_retexturing_supp}. In this example, the giraffe is behind a fence. However the corresponding mask does not hide the giraffe parts occluded by the fence. As such, our algorithm infills the giraffe on top of the fence, making the final result uncanny.

Finally, we generate results for the class `bus' and show a sample in Figure~\ref{fig:fig_bus_supp}. This is a difficult class with complex and diverse structured texture, including advertisements on the sides of busses, transparency from windows, different colored busses independent of lighting and shadow, and strong perspective effects from the rigid man-made object. Our algorithm fails on this class: it fails to create realistic shapes which adhere to the perspective of the background image, it fails to create convincing texture details, and it has trouble capturing the appearance diversity within this class. Future work could investigate more-explicit modeling of these parts of the class appearance space to improve generation.
 
\newcommand{\mysizewf}{.19\linewidth}
\setlength{\tabcolsep}{1pt}
\begin{table*}[t]
  \centering
\begin{tabular}{ccccc}
 \includegraphics[width=\mysizewf]{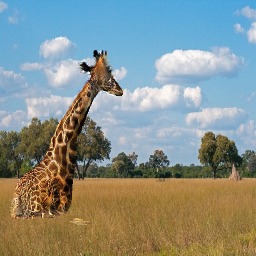} &
    \includegraphics[width=\mysizewf]{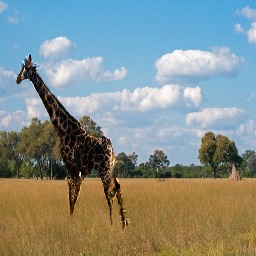} &
      \includegraphics[width=\mysizewf]{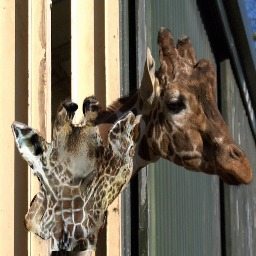} &
     \includegraphics[width=\mysizewf]{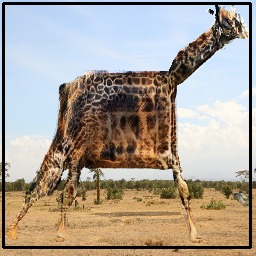} &
        \includegraphics[width=\mysizewf]{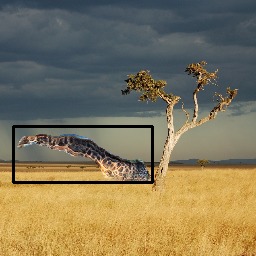}\\

 \includegraphics[width=\mysizewf]{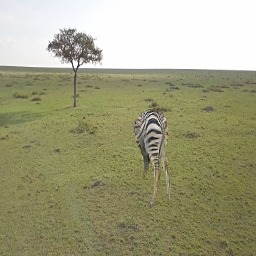} &
      \includegraphics[width=\mysizewf]{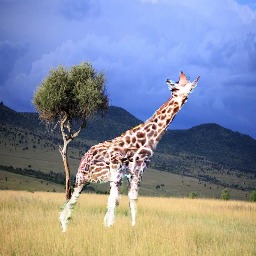} &
        \includegraphics[width=\mysizewf]{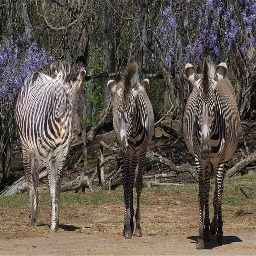} &
        \includegraphics[width=\mysizewf]{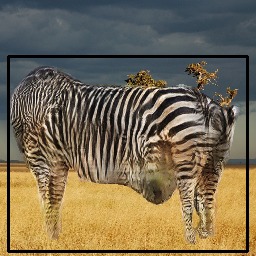} &
         \includegraphics[width=\mysizewf]{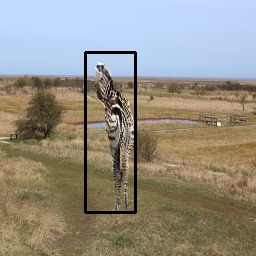} \\

\end{tabular}
 \captionof{figure}{Failure cases generated by our algorithm. The first column shows failures due to occluded instances in the training set. The second column shows failures due to mismatch between the foreground and background illumination. Column three shows failures due to the inability of our algorithm to generate fine details at multiple scales. Column four shows failure cases due to large input bounding boxes. The last column shows failures due to unusual bounding box shapes (e.g. high and thin bounding box for the class `zebra'.)}
  \label{fig:failures}
 \end{table*}
  
\setlength{\tabcolsep}{1pt}
\begin{table*}[t]
  \centering
\begin{tabular}{ccccc}
 \includegraphics[width=\mysizewf]{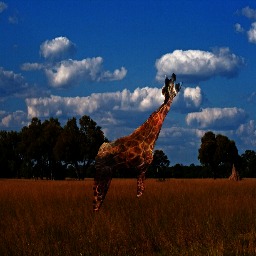} &
    \includegraphics[width=\mysizewf]{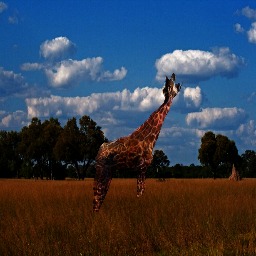} &
    \includegraphics[width=\mysizewf]{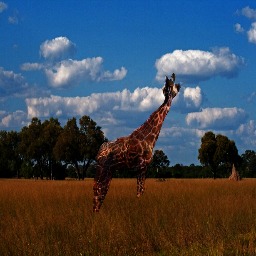} &
      \includegraphics[width=\mysizewf]{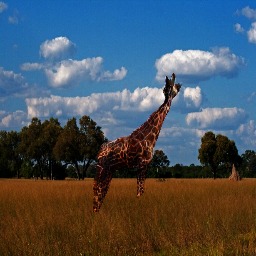} &
     \includegraphics[width=\mysizewf]{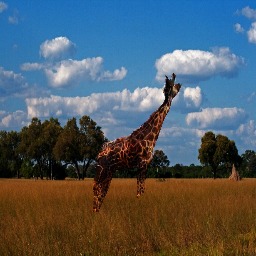} \\
     
      \includegraphics[width=\mysizewf]{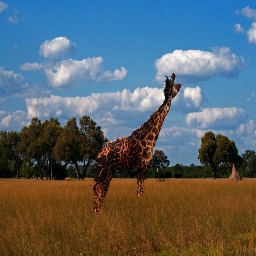} &
    \includegraphics[width=\mysizewf]{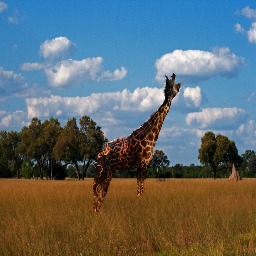} &
    \includegraphics[width=\mysizewf]{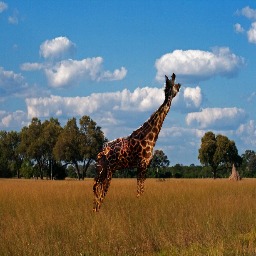} &
      \includegraphics[width=\mysizewf]{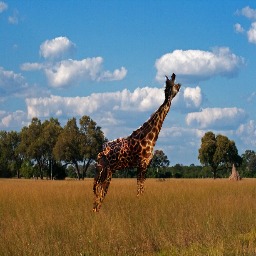} &
     \includegraphics[width=\mysizewf]{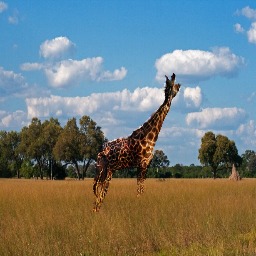} \\
     
      \includegraphics[width=\mysizewf]{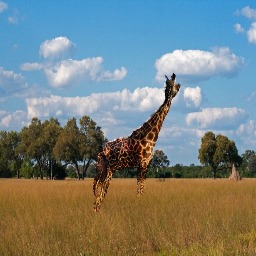} &
    \includegraphics[width=\mysizewf]{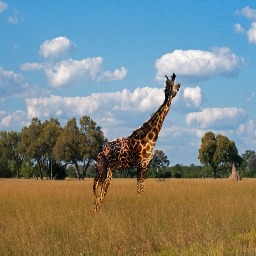} &
    \includegraphics[width=\mysizewf]{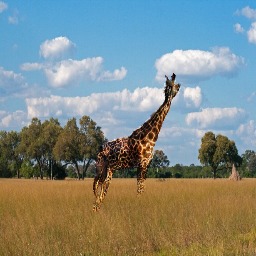} &
      \includegraphics[width=\mysizewf]{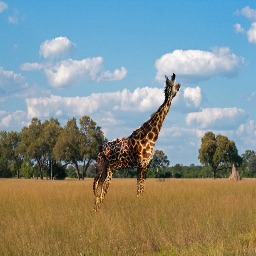} &
     \includegraphics[width=\mysizewf]{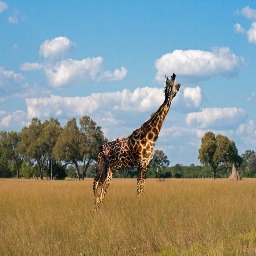} \\
     
      \includegraphics[width=\mysizewf]{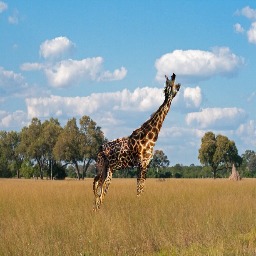} &
    \includegraphics[width=\mysizewf]{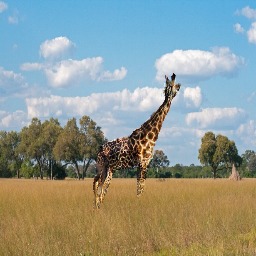} &
    \includegraphics[width=\mysizewf]{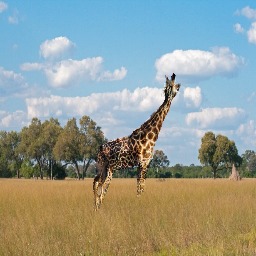} &
      \includegraphics[width=\mysizewf]{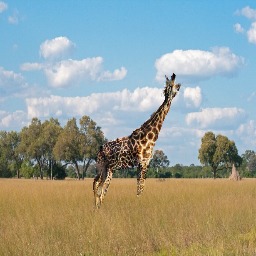} &
     \includegraphics[width=\mysizewf]{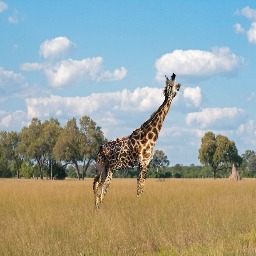} \\
     
      \includegraphics[width=\mysizewf]{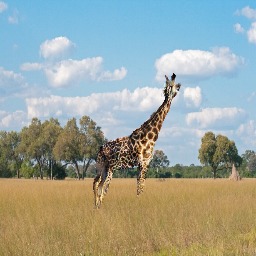} &
    \includegraphics[width=\mysizewf]{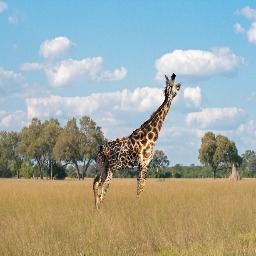} &
    \includegraphics[width=\mysizewf]{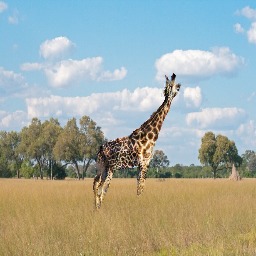} &
      \includegraphics[width=\mysizewf]{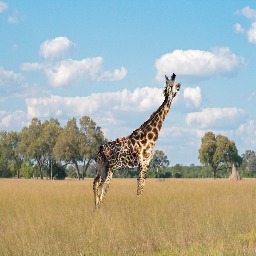} &
     \includegraphics[width=\mysizewf]{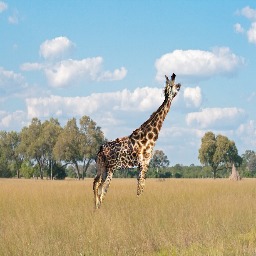} \\
     
      \includegraphics[width=\mysizewf]{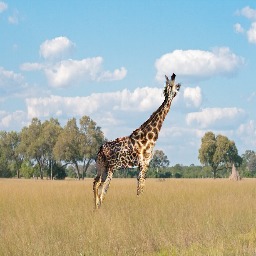} &
    \includegraphics[width=\mysizewf]{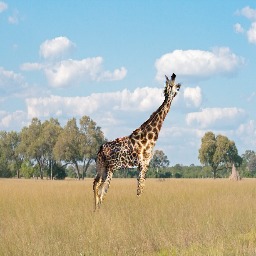} &
    \includegraphics[width=\mysizewf]{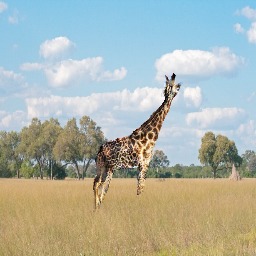} &
      \includegraphics[width=\mysizewf]{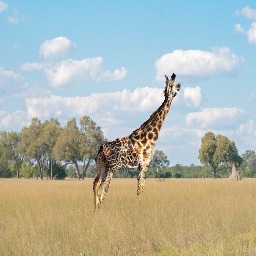} &
     \includegraphics[width=\mysizewf]{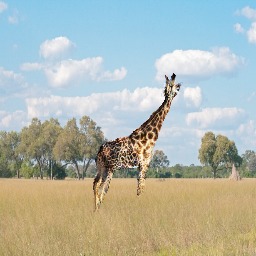} \\

\end{tabular}
 \captionof{figure}{Changing the background brightness while fixing the shape and texture latent codes. Our texture adapts to the background changes in brightness.}
  \label{fig:fig_change_gamma_gir_supp}
 
 \end{table*}
\setlength{\tabcolsep}{1pt}
\begin{table*}[t]
  \centering
\begin{tabular}{ccccc}
 \includegraphics[width=\mysizewf]{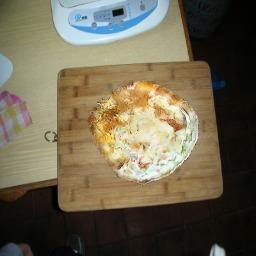} &
    \includegraphics[width=\mysizewf]{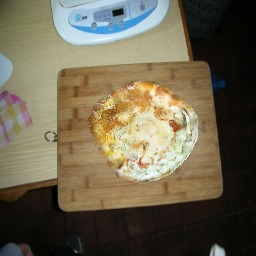} &
    \includegraphics[width=\mysizewf]{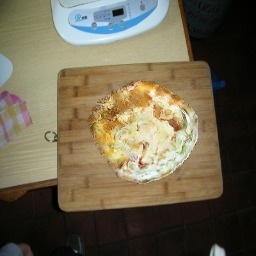} &
      \includegraphics[width=\mysizewf]{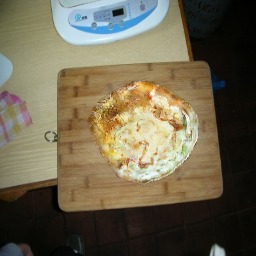} &
     \includegraphics[width=\mysizewf]{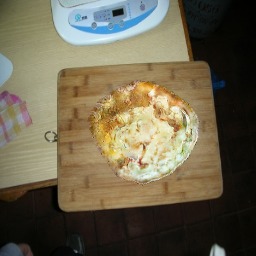} \\
     
      \includegraphics[width=\mysizewf]{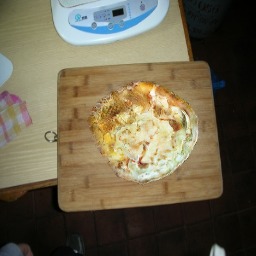} &
    \includegraphics[width=\mysizewf]{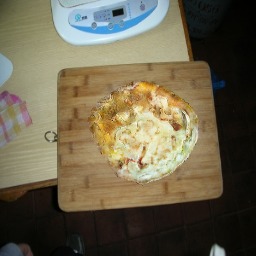} &
    \includegraphics[width=\mysizewf]{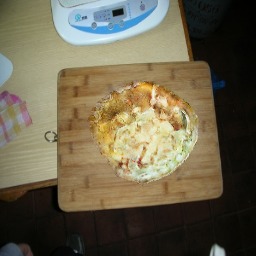} &
      \includegraphics[width=\mysizewf]{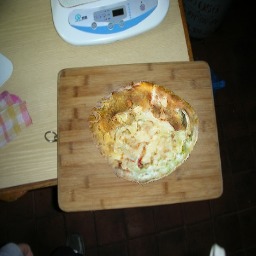} &
     \includegraphics[width=\mysizewf]{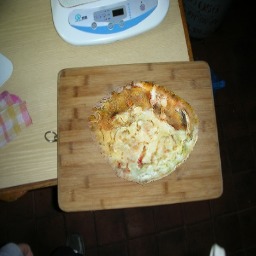} \\
     
      \includegraphics[width=\mysizewf]{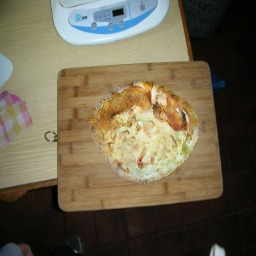} &
    \includegraphics[width=\mysizewf]{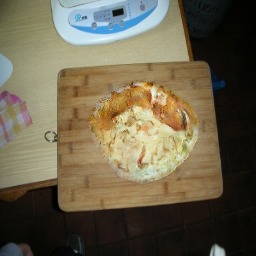} &
    \includegraphics[width=\mysizewf]{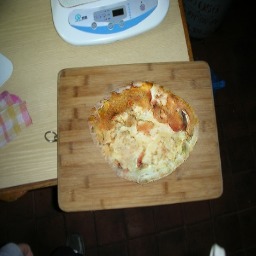} &
      \includegraphics[width=\mysizewf]{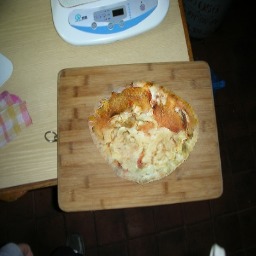} &
     \includegraphics[width=\mysizewf]{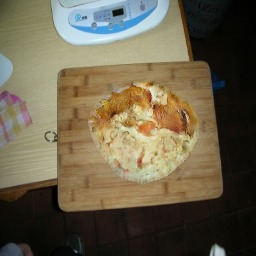} \\
     
      \includegraphics[width=\mysizewf]{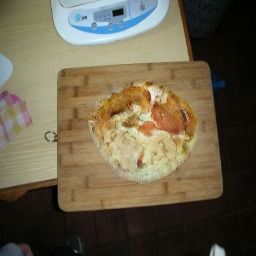} &
    \includegraphics[width=\mysizewf]{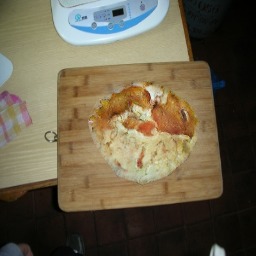} &
    \includegraphics[width=\mysizewf]{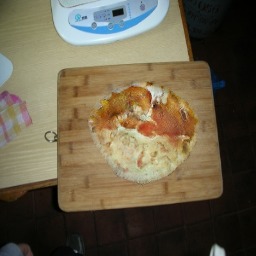} &
      \includegraphics[width=\mysizewf]{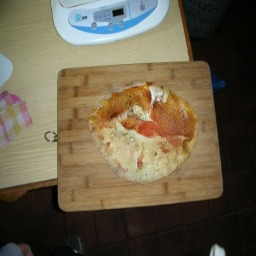} &
     \includegraphics[width=\mysizewf]{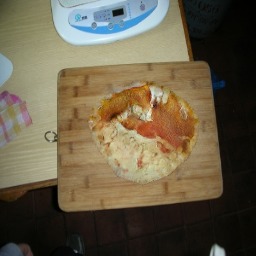} \\
     
      \includegraphics[width=\mysizewf]{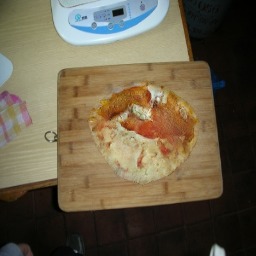} &
    \includegraphics[width=\mysizewf]{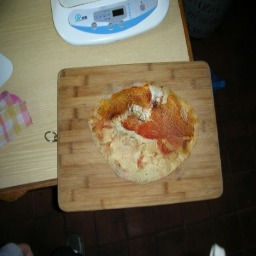} &
    \includegraphics[width=\mysizewf]{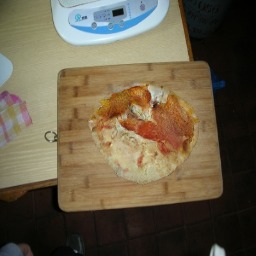} &
      \includegraphics[width=\mysizewf]{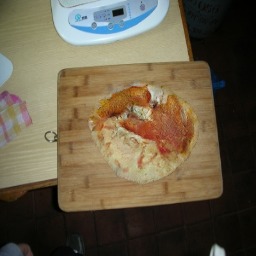} &
     \includegraphics[width=\mysizewf]{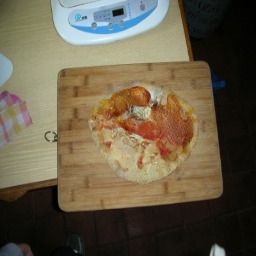} \\
     
      \includegraphics[width=\mysizewf]{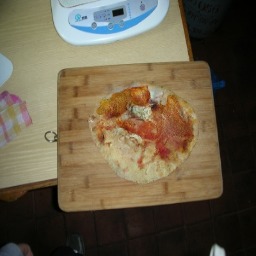} &
    \includegraphics[width=\mysizewf]{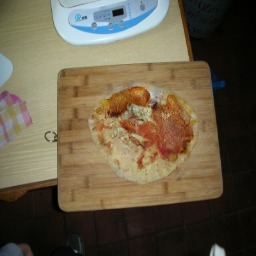} &
    \includegraphics[width=\mysizewf]{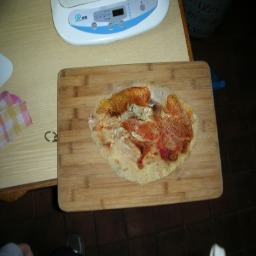} &
      \includegraphics[width=\mysizewf]{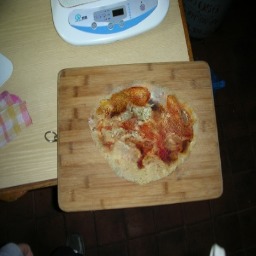} &
     \includegraphics[width=\mysizewf]{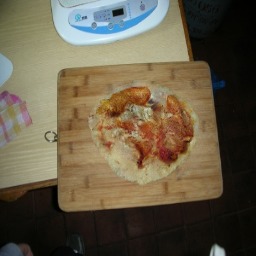} \\

\end{tabular}
 \captionof{figure}{Linear interpolation on the latent vector $\mathbf{z}_t$ controlling the texture space. The first and last images being the interpolation extremes. Our network has learned a smooth texture space.}
  \label{fig:fig_instance_interpolation_pizza_supp}
 
 \end{table*}
\setlength{\tabcolsep}{1pt}
\begin{table*}[t]
  \centering
\begin{tabular}{ccccc}
 \includegraphics[width=\mysizewf]{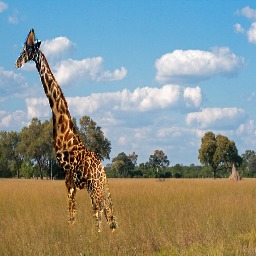} &
    \includegraphics[width=\mysizewf]{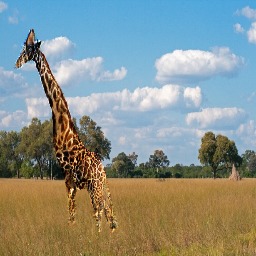} &
    \includegraphics[width=\mysizewf]{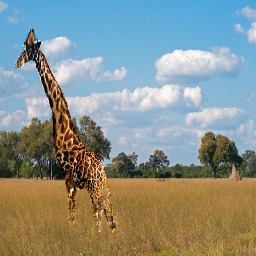} &
      \includegraphics[width=\mysizewf]{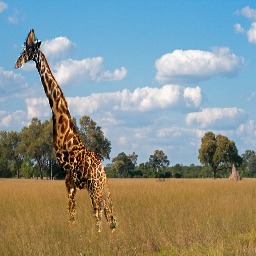} &
     \includegraphics[width=\mysizewf]{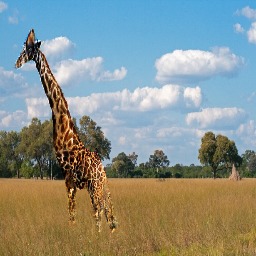} \\
     
      \includegraphics[width=\mysizewf]{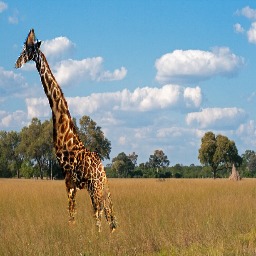} &
    \includegraphics[width=\mysizewf]{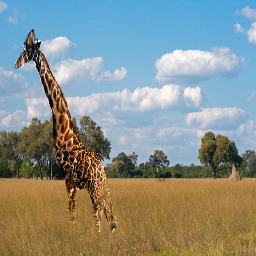} &
    \includegraphics[width=\mysizewf]{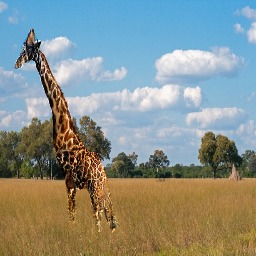} &
      \includegraphics[width=\mysizewf]{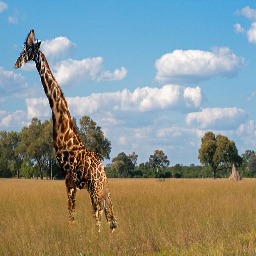} &
     \includegraphics[width=\mysizewf]{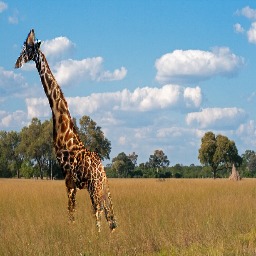} \\
     
      \includegraphics[width=\mysizewf]{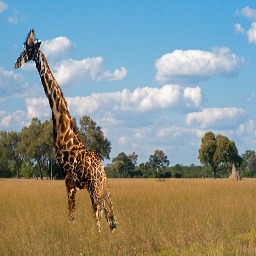} &
    \includegraphics[width=\mysizewf]{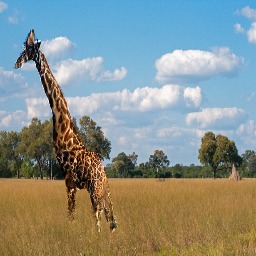} &
    \includegraphics[width=\mysizewf]{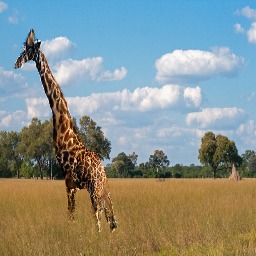} &
      \includegraphics[width=\mysizewf]{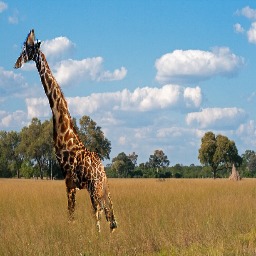} &
     \includegraphics[width=\mysizewf]{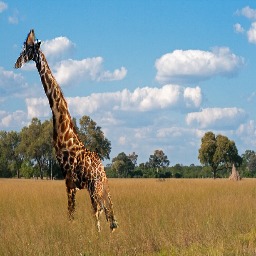} \\
     
      \includegraphics[width=\mysizewf]{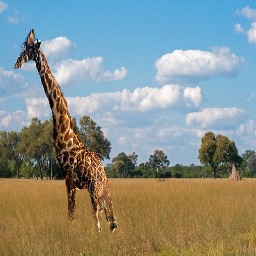} &
    \includegraphics[width=\mysizewf]{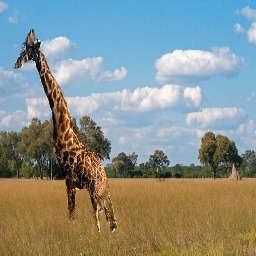} &
    \includegraphics[width=\mysizewf]{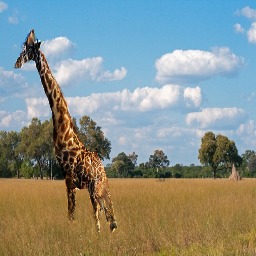} &
      \includegraphics[width=\mysizewf]{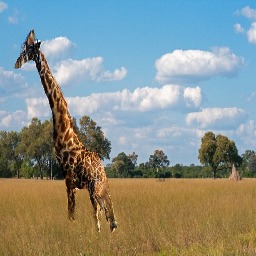} &
     \includegraphics[width=\mysizewf]{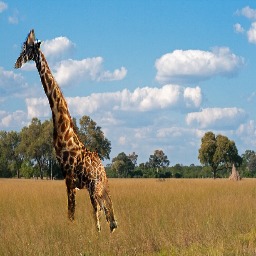} \\
     
      \includegraphics[width=\mysizewf]{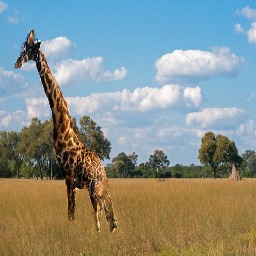} &
    \includegraphics[width=\mysizewf]{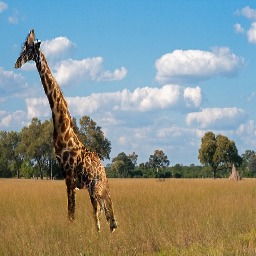} &
    \includegraphics[width=\mysizewf]{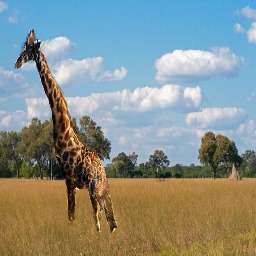} &
      \includegraphics[width=\mysizewf]{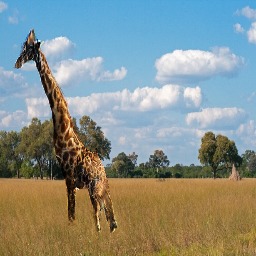} &
     \includegraphics[width=\mysizewf]{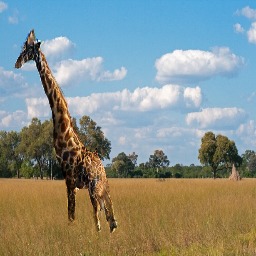} \\
     
      \includegraphics[width=\mysizewf]{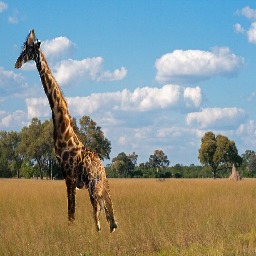} &
    \includegraphics[width=\mysizewf]{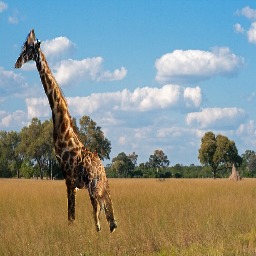} &
    \includegraphics[width=\mysizewf]{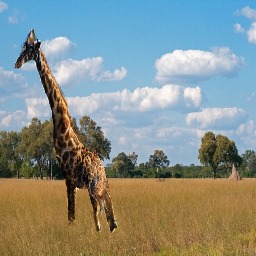} &
      \includegraphics[width=\mysizewf]{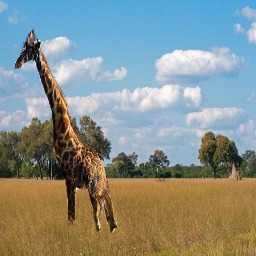} &
     \includegraphics[width=\mysizewf]{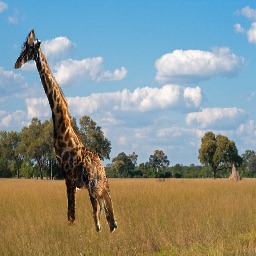} \\

\end{tabular}
 \captionof{figure}{Linear interpolation on the latent vector $\mathbf{z}_t$ controlling the texture space. The first and last images being the interpolation extremes. Our network has learned a smooth texture space.}
  \label{fig:fig_instance_interpolation_gir_supp}
 
 \end{table*}
\setlength{\tabcolsep}{1pt}
\begin{table*}[t]
  \centering
\begin{tabular}{ccccc}
 \includegraphics[width=\mysizewf]{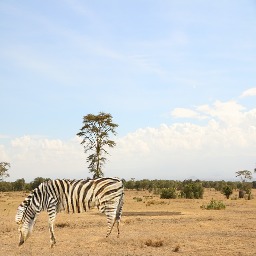} &
    \includegraphics[width=\mysizewf]{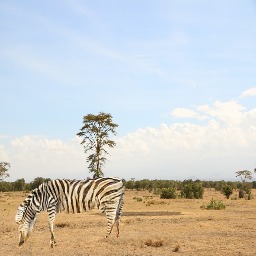} &
    \includegraphics[width=\mysizewf]{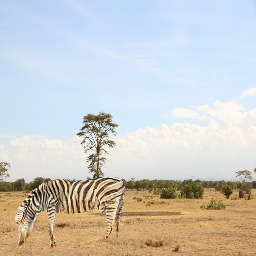} &
      \includegraphics[width=\mysizewf]{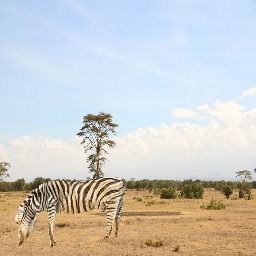} &
     \includegraphics[width=\mysizewf]{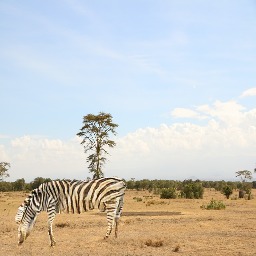} \\
     
      \includegraphics[width=\mysizewf]{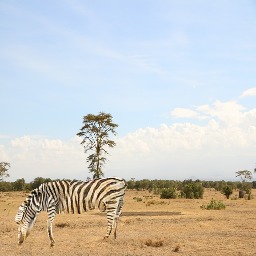} &
    \includegraphics[width=\mysizewf]{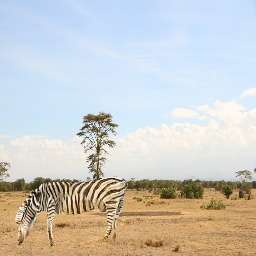} &
    \includegraphics[width=\mysizewf]{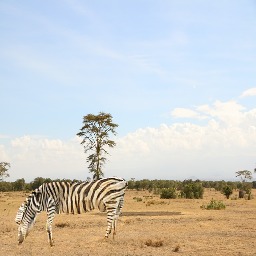} &
      \includegraphics[width=\mysizewf]{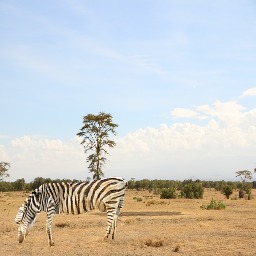} &
     \includegraphics[width=\mysizewf]{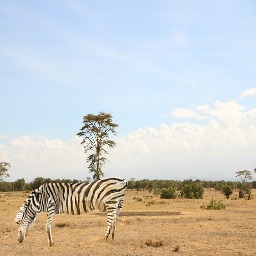} \\
     
      \includegraphics[width=\mysizewf]{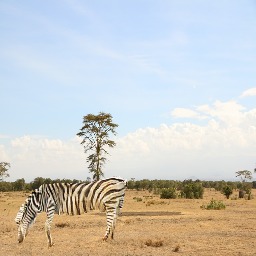} &
    \includegraphics[width=\mysizewf]{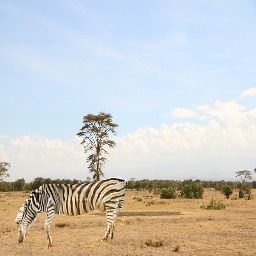} &
    \includegraphics[width=\mysizewf]{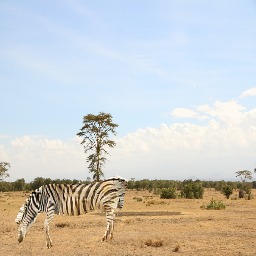} &
      \includegraphics[width=\mysizewf]{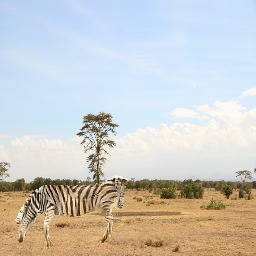} &
     \includegraphics[width=\mysizewf]{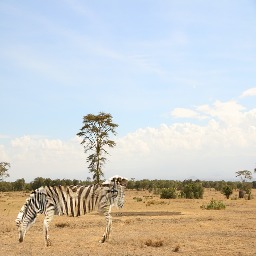} \\
     
      \includegraphics[width=\mysizewf]{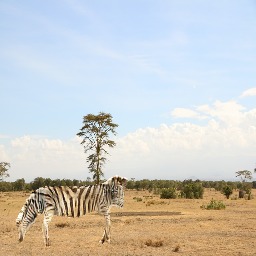} &
    \includegraphics[width=\mysizewf]{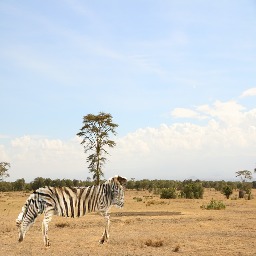} &
    \includegraphics[width=\mysizewf]{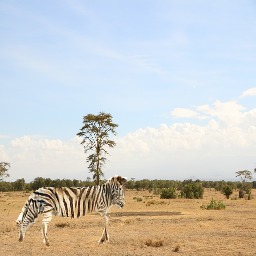} &
      \includegraphics[width=\mysizewf]{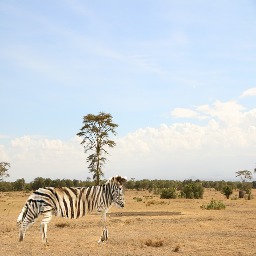} &
     \includegraphics[width=\mysizewf]{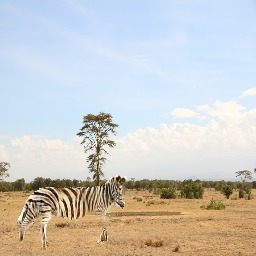} \\
     
      \includegraphics[width=\mysizewf]{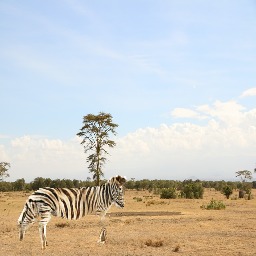} &
    \includegraphics[width=\mysizewf]{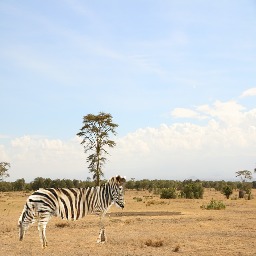} &
    \includegraphics[width=\mysizewf]{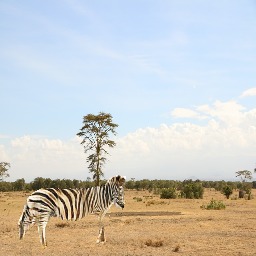} &
      \includegraphics[width=\mysizewf]{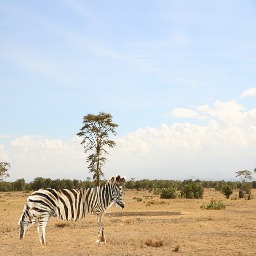} &
     \includegraphics[width=\mysizewf]{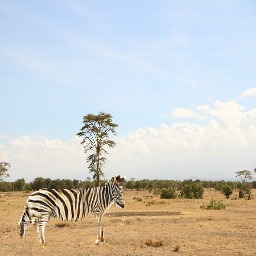} \\
     
      \includegraphics[width=\mysizewf]{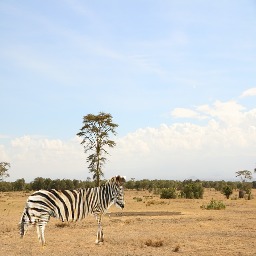} &
    \includegraphics[width=\mysizewf]{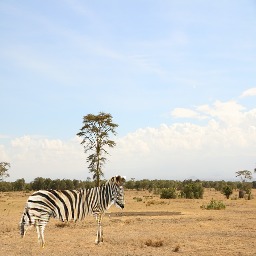} &
    \includegraphics[width=\mysizewf]{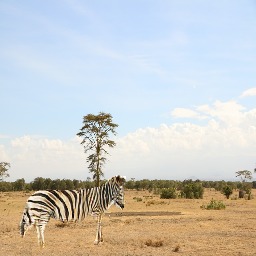} &
      \includegraphics[width=\mysizewf]{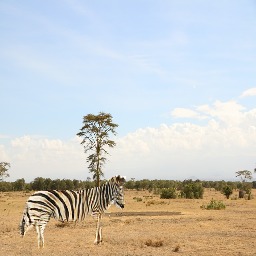} &
     \includegraphics[width=\mysizewf]{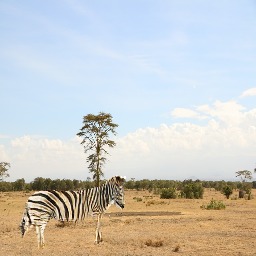} \\

\end{tabular}
 \captionof{figure}{Linear interpolation on the latent vector $\mathbf{z}_m$ controlling the shape generation. The first and last images being the interpolation extremes. Note that the bounding box is fixed throughout the interpolation, as is the texture latent vector $\mathbf{z}_t$. Our network has learned a smooth shape space; however, there is no guarantee that intermediate images will remain realistic. For example, the zebra head on the left slowly disappears and switches to the right. A smaller bounding box shape would be required to generate a `front' or `rear' mask shape.}
  \label{fig:fig_mask_interpolation_zebra_supp}
 
 \end{table*}
\setlength{\tabcolsep}{1pt}
\begin{table*}[t]
  \centering
\begin{tabular}{ccccc}
 \includegraphics[width=\mysizewf]{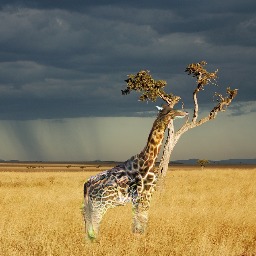} &
    \includegraphics[width=\mysizewf]{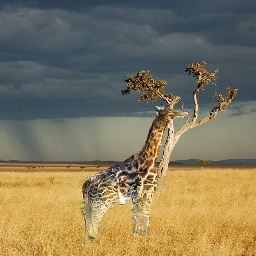} &
    \includegraphics[width=\mysizewf]{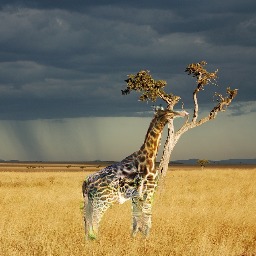} &
      \includegraphics[width=\mysizewf]{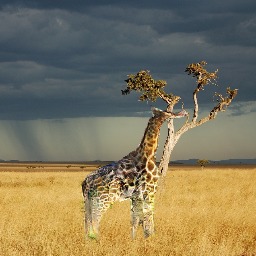} &
     \includegraphics[width=\mysizewf]{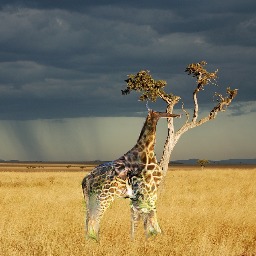} \\
     
      \includegraphics[width=\mysizewf]{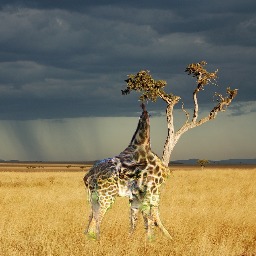} &
    \includegraphics[width=\mysizewf]{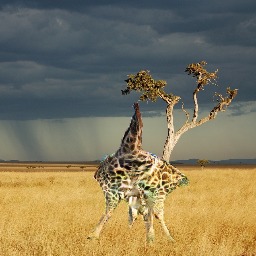} &
    \includegraphics[width=\mysizewf]{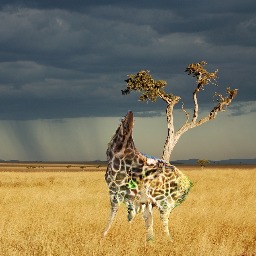} &
      \includegraphics[width=\mysizewf]{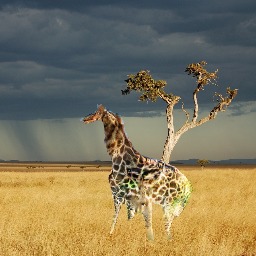} &
     \includegraphics[width=\mysizewf]{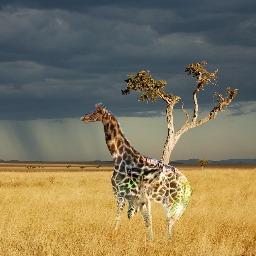} \\
     
      \includegraphics[width=\mysizewf]{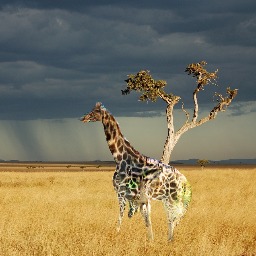} &
    \includegraphics[width=\mysizewf]{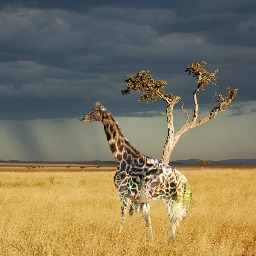} &
    \includegraphics[width=\mysizewf]{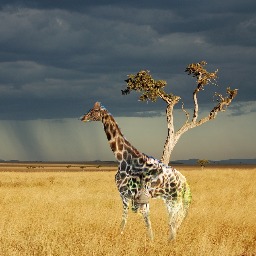} &
      \includegraphics[width=\mysizewf]{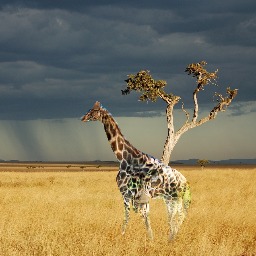} &
     \includegraphics[width=\mysizewf]{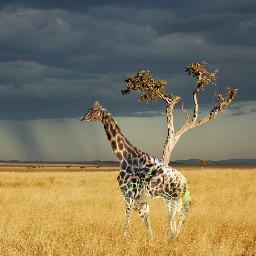} \\
     
      \includegraphics[width=\mysizewf]{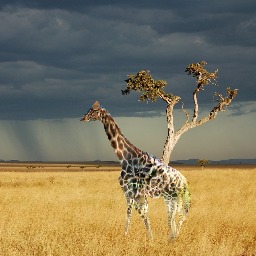} &
    \includegraphics[width=\mysizewf]{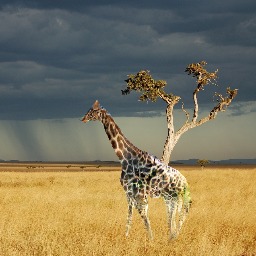} &
    \includegraphics[width=\mysizewf]{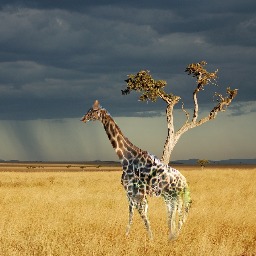} &
      \includegraphics[width=\mysizewf]{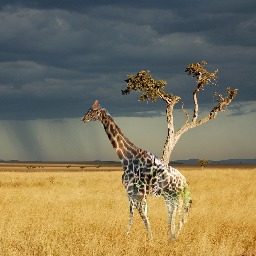} &
     \includegraphics[width=\mysizewf]{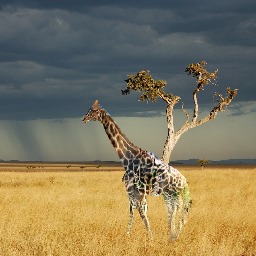} \\
     
      \includegraphics[width=\mysizewf]{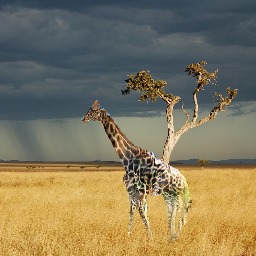} &
    \includegraphics[width=\mysizewf]{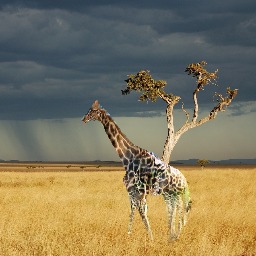} &
    \includegraphics[width=\mysizewf]{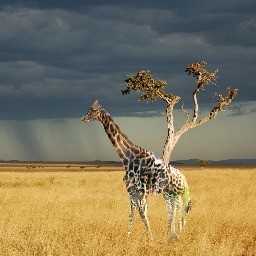} &
      \includegraphics[width=\mysizewf]{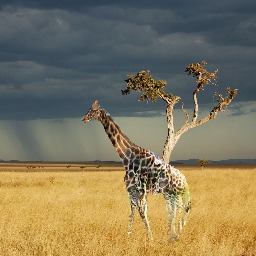} &
     \includegraphics[width=\mysizewf]{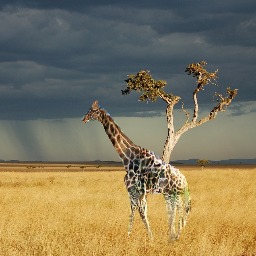} \\
     
      \includegraphics[width=\mysizewf]{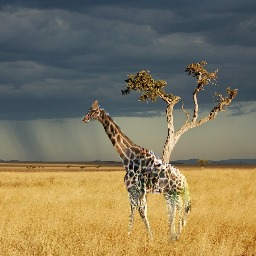} &
    \includegraphics[width=\mysizewf]{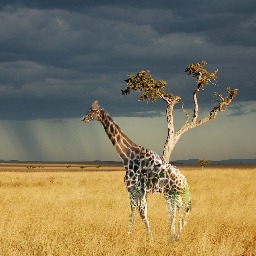} &
    \includegraphics[width=\mysizewf]{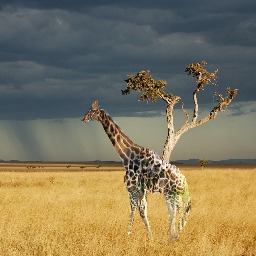} &
      \includegraphics[width=\mysizewf]{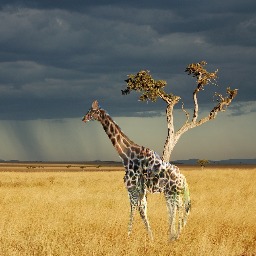} &
     \includegraphics[width=\mysizewf]{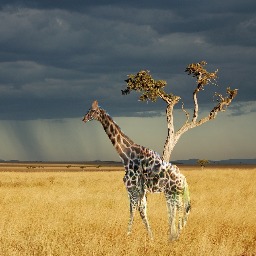} \\

\end{tabular}
 \captionof{figure}{Linear interpolation on the latent vector $\mathbf{z}_m$ controlling the shape generation. The first and last images being the interpolation extremes. Note that the texture latent vector $\mathbf{z}_t$ is fixed throughout the interpolation. Our network has learned a smooth shape space; however, there are no guarantees that intermediate images would remain realistic.}
  \label{fig:fig_mask_interpolation_gir_supp}
 
 \end{table*}
\setlength{\tabcolsep}{1pt}
\begin{table*}[t]
  \centering
\begin{tabular}{ccccc}
 \includegraphics[width=\mysizewf]{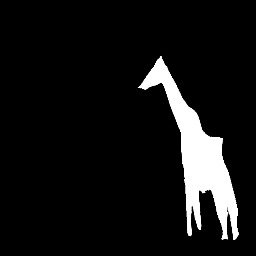} &
    \includegraphics[width=\mysizewf]{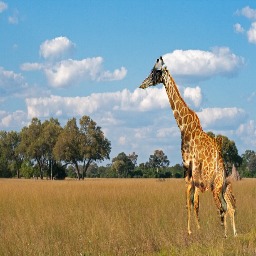} &
    \includegraphics[width=\mysizewf]{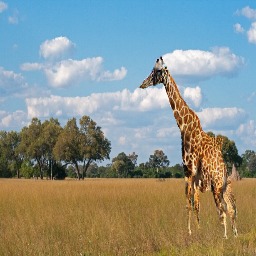} &
      \includegraphics[width=\mysizewf]{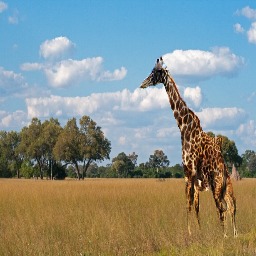} &
     \includegraphics[width=\mysizewf]{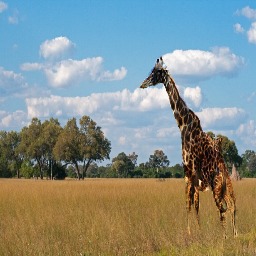} \\
      
       \includegraphics[width=\mysizewf]{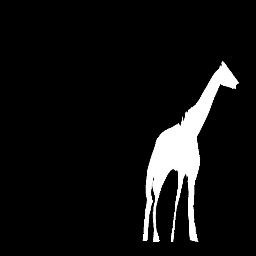} &
    \includegraphics[width=\mysizewf]{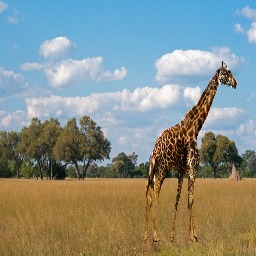} &
    \includegraphics[width=\mysizewf]{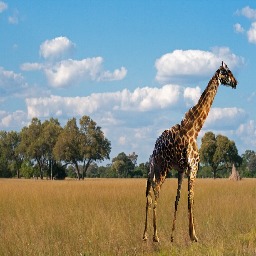} &
      \includegraphics[width=\mysizewf]{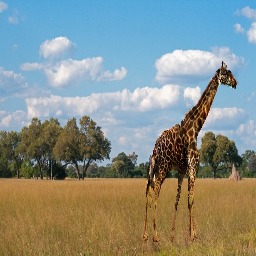} &
     \includegraphics[width=\mysizewf]{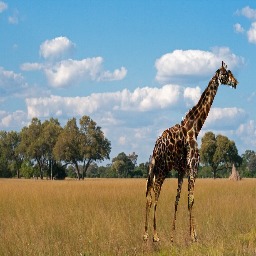} \\
        
       \includegraphics[width=\mysizewf]{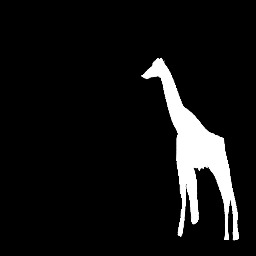} &
             \includegraphics[width=\mysizewf]{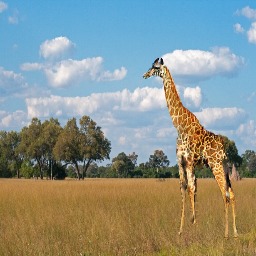} &
     \includegraphics[width=\mysizewf]{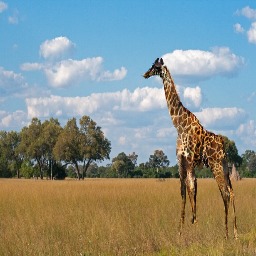} &
    \includegraphics[width=\mysizewf]{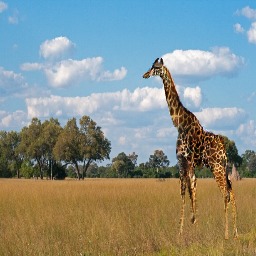} &
        \includegraphics[width=\mysizewf]{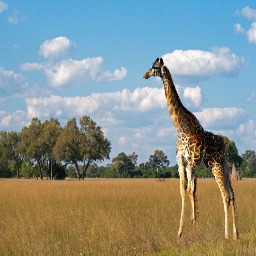}\\

 \includegraphics[width=\mysizewf]{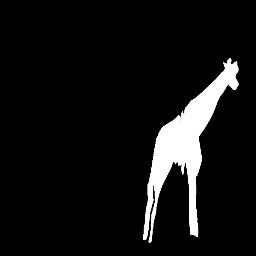} &
      \includegraphics[width=\mysizewf]{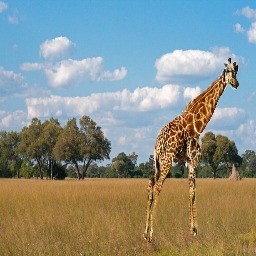} &
     \includegraphics[width=\mysizewf]{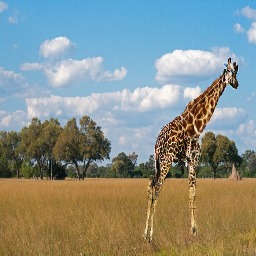} &
         \includegraphics[width=\mysizewf]{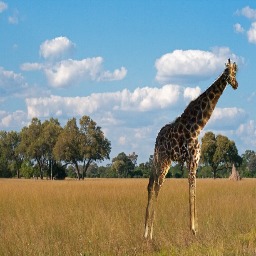} &
        \includegraphics[width=\mysizewf]{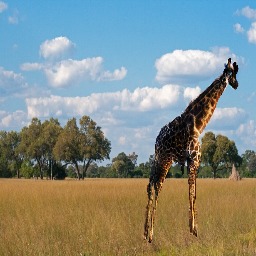}\\
        
 \includegraphics[width=\mysizewf]{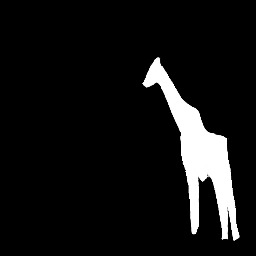} &
         \includegraphics[width=\mysizewf]{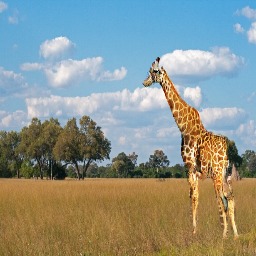}&
         \includegraphics[width=\mysizewf]{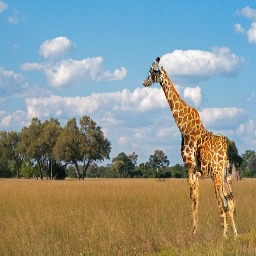} &
        \includegraphics[width=\mysizewf]{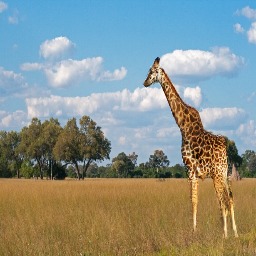} &
        \includegraphics[width=\mysizewf]{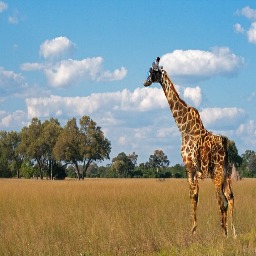}\\
        
 \includegraphics[width=\mysizewf]{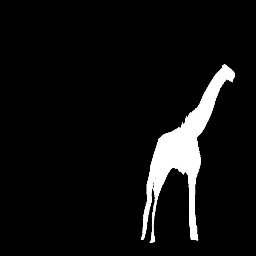} &
      \includegraphics[width=\mysizewf]{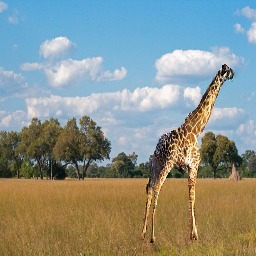} &
        \includegraphics[width=\mysizewf]{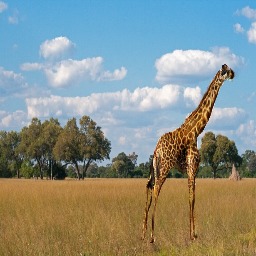}&
    \includegraphics[width=\mysizewf]{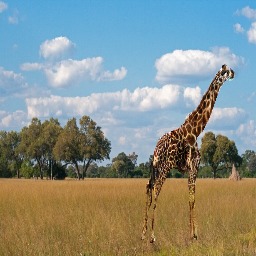} &
      \includegraphics[width=\mysizewf]{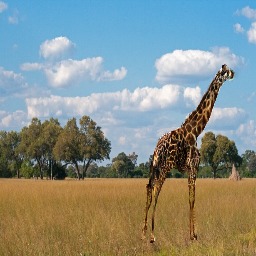} \\

\end{tabular}
 \captionof{figure}{Masks generated for the COCO class `giraffe' and the corresponding synthesized textures. Each row shows four different texture generation results based on the generated shape mask in the first column.}
  \label{fig:fig_giraffe_supp}
 
 \end{table*}
\setlength{\tabcolsep}{1pt}
\begin{table*}[t]
  \centering
\begin{tabular}{ccccc}
 \includegraphics[width=\mysizewf]{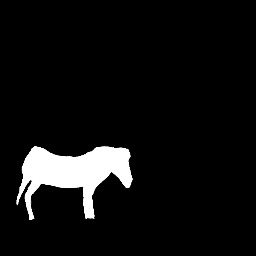} &
    \includegraphics[width=\mysizewf]{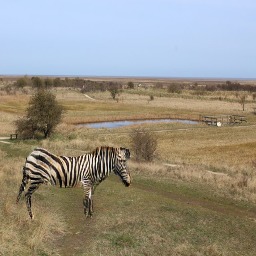} &
    \includegraphics[width=\mysizewf]{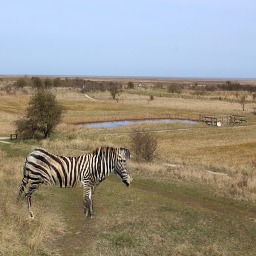} &
    \includegraphics[width=\mysizewf]{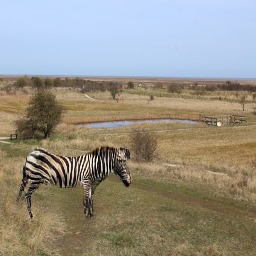} &
    \includegraphics[width=\mysizewf]{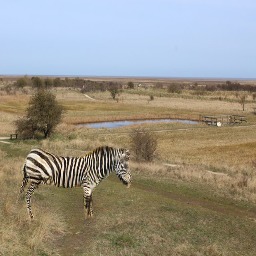} \\
      
       \includegraphics[width=\mysizewf]{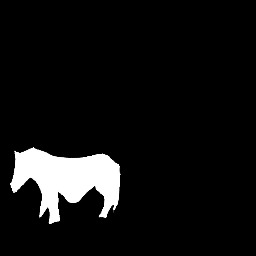} &
    \includegraphics[width=\mysizewf]{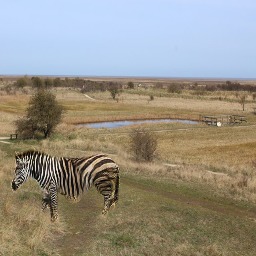} &
    \includegraphics[width=\mysizewf]{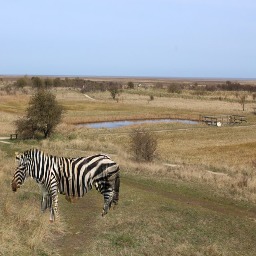} &
    \includegraphics[width=\mysizewf]{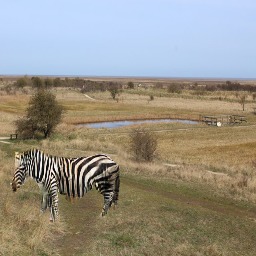} &
    \includegraphics[width=\mysizewf]{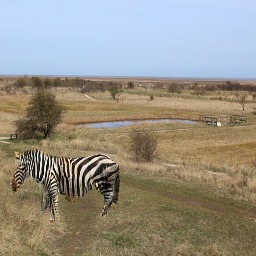} \\
        
       \includegraphics[width=\mysizewf]{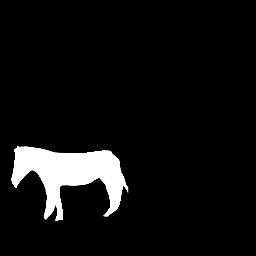} &
    \includegraphics[width=\mysizewf]{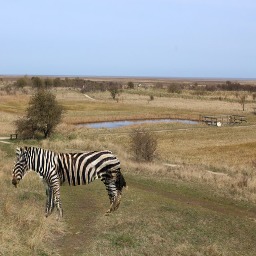} &
    \includegraphics[width=\mysizewf]{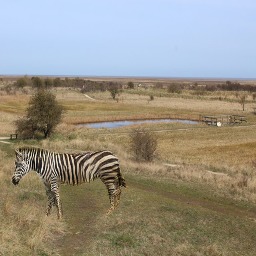} &
    \includegraphics[width=\mysizewf]{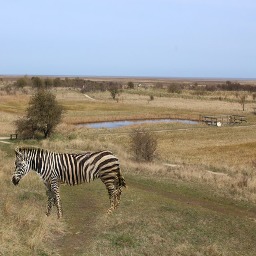} &
    \includegraphics[width=\mysizewf]{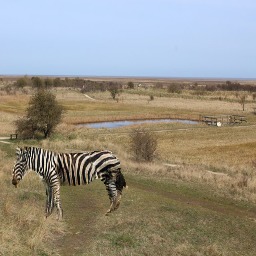} \\
        
       \includegraphics[width=\mysizewf]{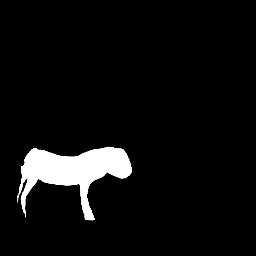} &
    \includegraphics[width=\mysizewf]{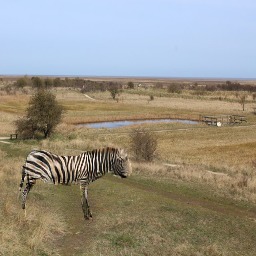} &
    \includegraphics[width=\mysizewf]{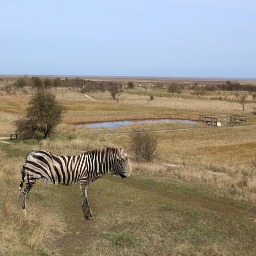} &
    \includegraphics[width=\mysizewf]{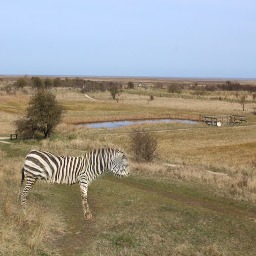} &
    \includegraphics[width=\mysizewf]{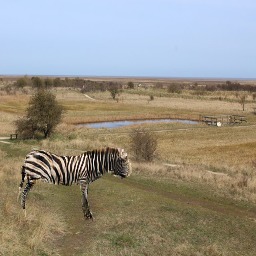} \\
        
       \includegraphics[width=\mysizewf]{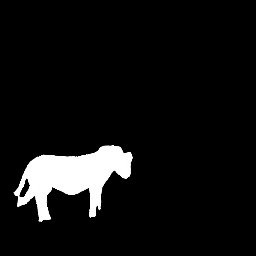} &
    \includegraphics[width=\mysizewf]{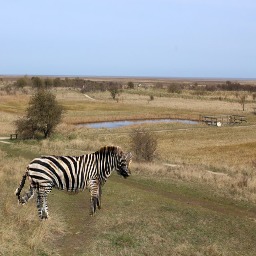} &
    \includegraphics[width=\mysizewf]{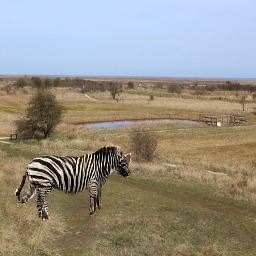} &
    \includegraphics[width=\mysizewf]{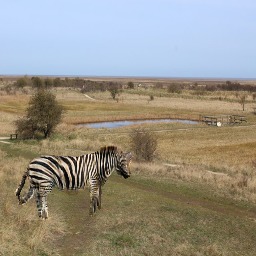} &
    \includegraphics[width=\mysizewf]{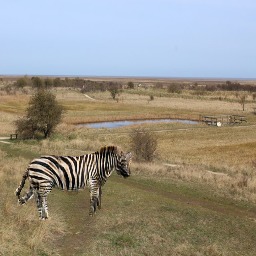} \\
        
       \includegraphics[width=\mysizewf]{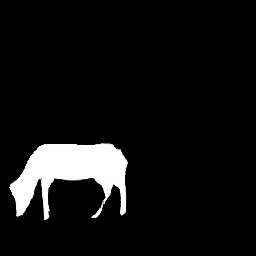} &
    \includegraphics[width=\mysizewf]{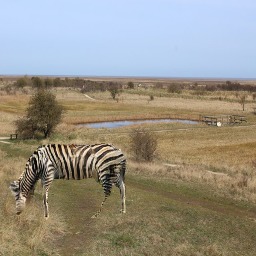} &
    \includegraphics[width=\mysizewf]{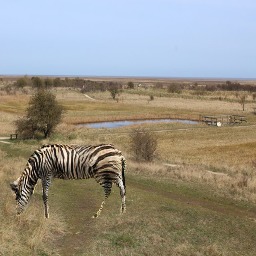} &
    \includegraphics[width=\mysizewf]{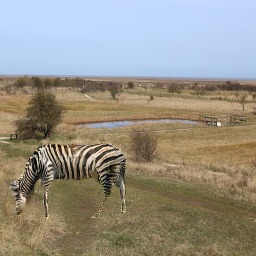} &
    \includegraphics[width=\mysizewf]{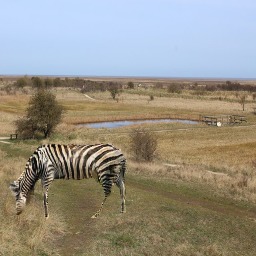} \\

\end{tabular}
 \captionof{figure}{Masks generated for the COCO class `zebra' and the corresponding synthesized textures. Each row shows four different texture generation results based on the generated shape mask in the first column.}
  \label{fig:fig_zebra_supp}
 \end{table*}
\setlength{\tabcolsep}{1pt}
\begin{table*}[t]
  \centering
\begin{tabular}{ccccc}
 \includegraphics[width=\mysizewf]{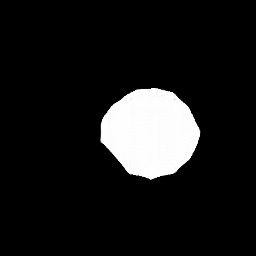} &
    \includegraphics[width=\mysizewf]{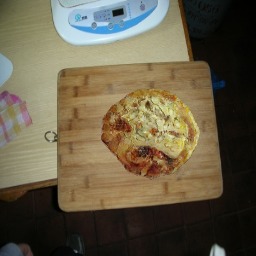} &
    \includegraphics[width=\mysizewf]{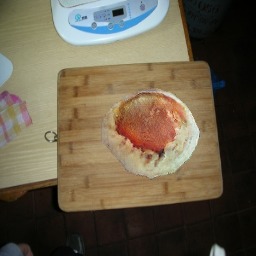} &
    \includegraphics[width=\mysizewf]{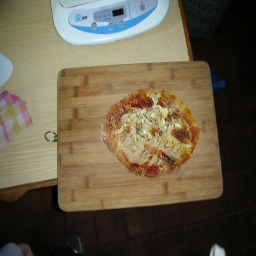} &
    \includegraphics[width=\mysizewf]{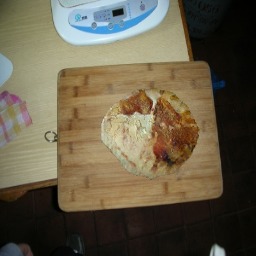} \\
    
     \includegraphics[width=\mysizewf]{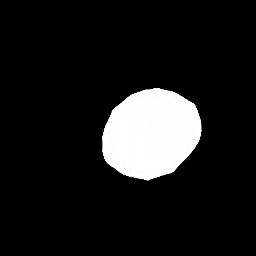} &
    \includegraphics[width=\mysizewf]{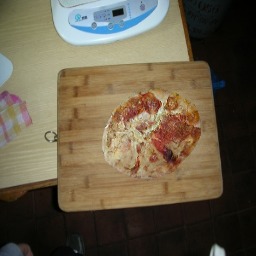} &
    \includegraphics[width=\mysizewf]{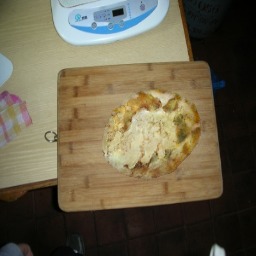} &
    \includegraphics[width=\mysizewf]{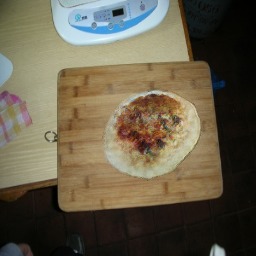} &
    \includegraphics[width=\mysizewf]{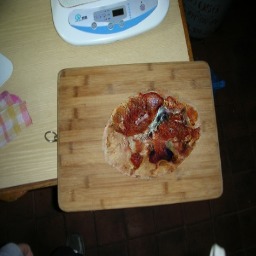} \\
    
         \includegraphics[width=\mysizewf]{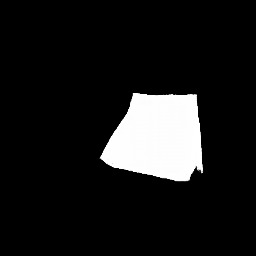} &
    \includegraphics[width=\mysizewf]{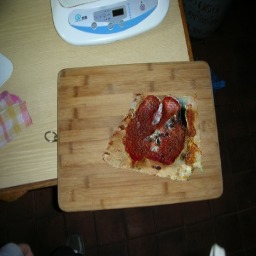} &
    \includegraphics[width=\mysizewf]{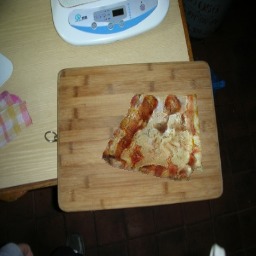} &
    \includegraphics[width=\mysizewf]{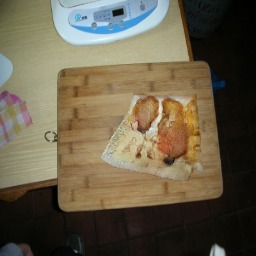} &
    \includegraphics[width=\mysizewf]{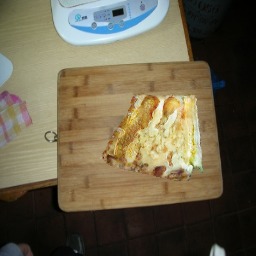} \\
    
         \includegraphics[width=\mysizewf]{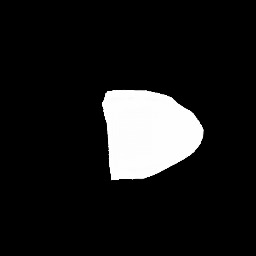} &
    \includegraphics[width=\mysizewf]{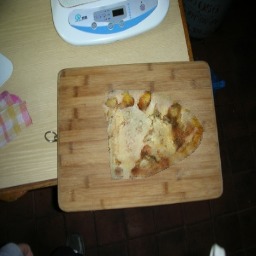} &
    \includegraphics[width=\mysizewf]{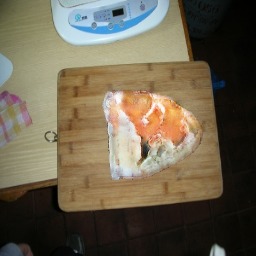} &
    \includegraphics[width=\mysizewf]{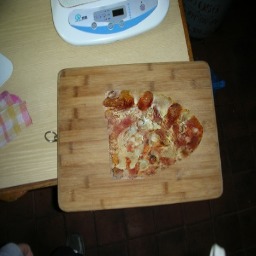} &
    \includegraphics[width=\mysizewf]{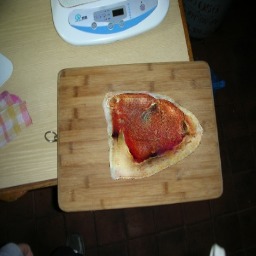} \\
    
         \includegraphics[width=\mysizewf]{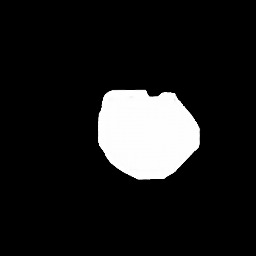} &
    \includegraphics[width=\mysizewf]{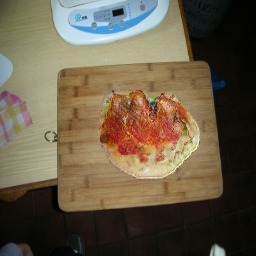} &
    \includegraphics[width=\mysizewf]{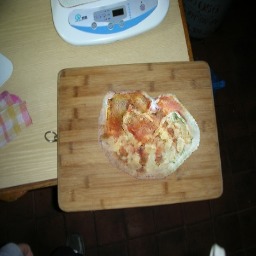} &
    \includegraphics[width=\mysizewf]{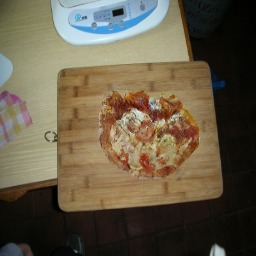} &
    \includegraphics[width=\mysizewf]{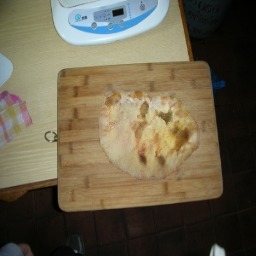} \\
    
         \includegraphics[width=\mysizewf]{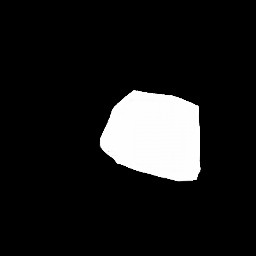} &
    \includegraphics[width=\mysizewf]{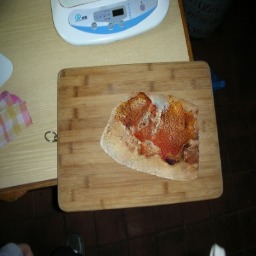} &
    \includegraphics[width=\mysizewf]{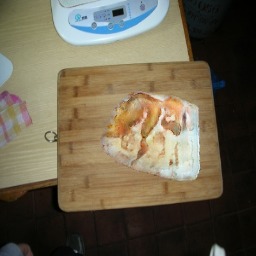} &
    \includegraphics[width=\mysizewf]{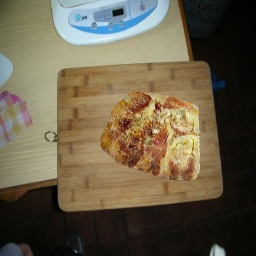} &
    \includegraphics[width=\mysizewf]{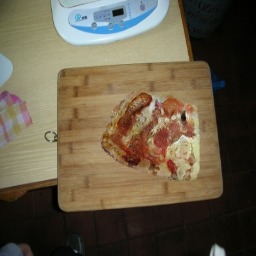} \\

\end{tabular}
 \captionof{figure}{Masks generated for the COCO class `pizza' and the corresponding synthesized textures. Each row shows four different texture generation results based on the generated shape mask in the first column.}
  \label{fig:fig_pizza_supp}
 \end{table*}
\setlength{\tabcolsep}{1pt}
\begin{table*}[t]
  \centering
\begin{tabular}{ccccc}
 \includegraphics[width=\mysizewf]{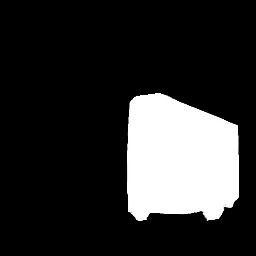} &
    \includegraphics[width=\mysizewf]{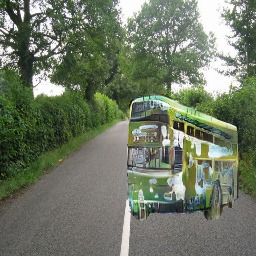} &
    \includegraphics[width=\mysizewf]{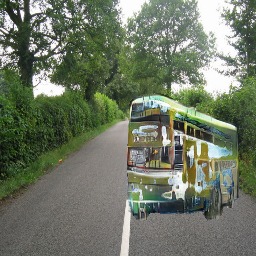} &
    \includegraphics[width=\mysizewf]{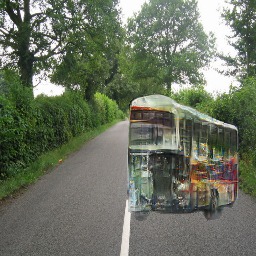} &
    \includegraphics[width=\mysizewf]{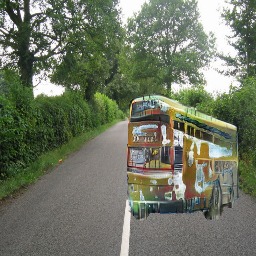} \\
    
     \includegraphics[width=\mysizewf]{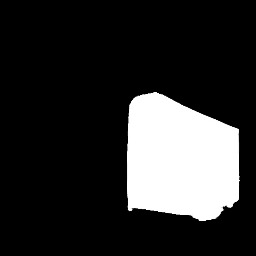} &
    \includegraphics[width=\mysizewf]{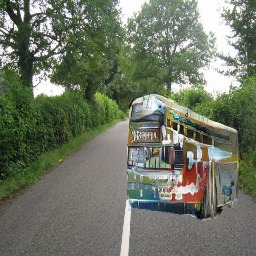} &
    \includegraphics[width=\mysizewf]{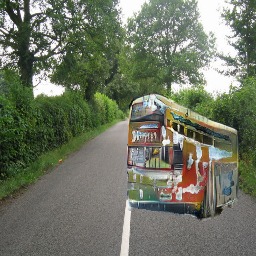} &
    \includegraphics[width=\mysizewf]{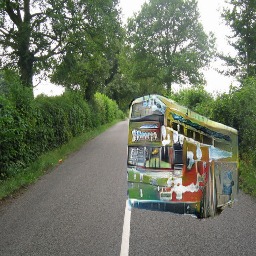} &
    \includegraphics[width=\mysizewf]{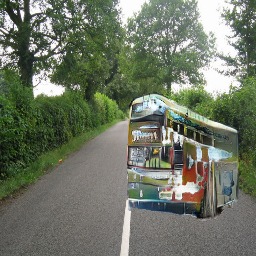} \\
    
         \includegraphics[width=\mysizewf]{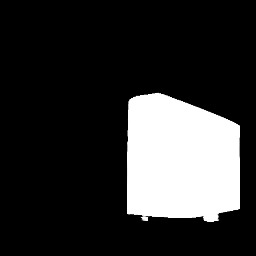} &
    \includegraphics[width=\mysizewf]{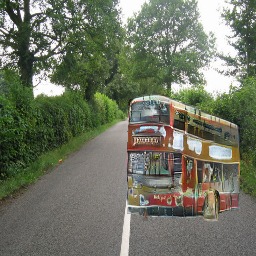} &
    \includegraphics[width=\mysizewf]{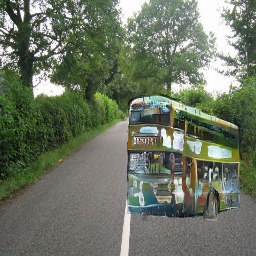} &
    \includegraphics[width=\mysizewf]{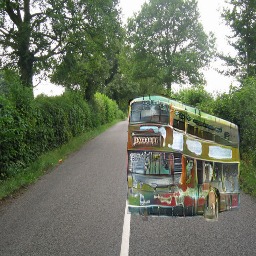} &
    \includegraphics[width=\mysizewf]{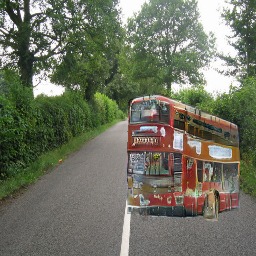} \\
    
         \includegraphics[width=\mysizewf]{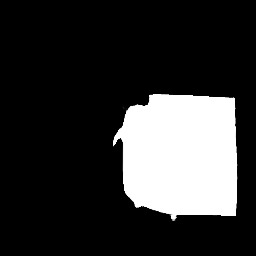} &
    \includegraphics[width=\mysizewf]{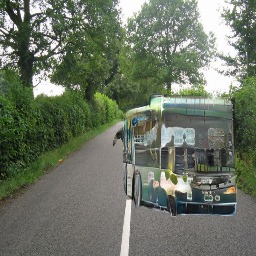} &
    \includegraphics[width=\mysizewf]{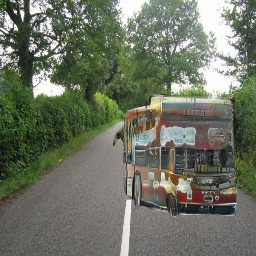} &
    \includegraphics[width=\mysizewf]{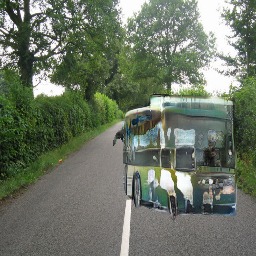} &
    \includegraphics[width=\mysizewf]{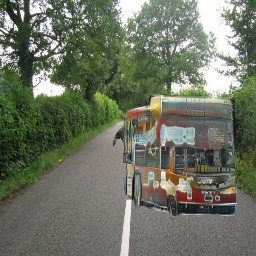} \\
    
         \includegraphics[width=\mysizewf]{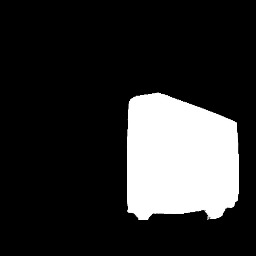} &
    \includegraphics[width=\mysizewf]{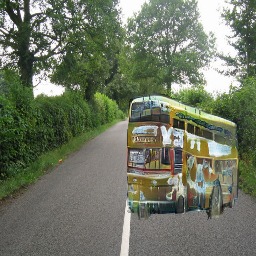} &
    \includegraphics[width=\mysizewf]{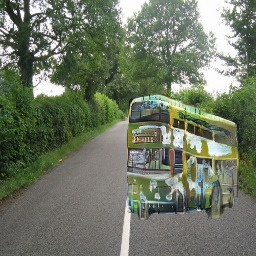} &
    \includegraphics[width=\mysizewf]{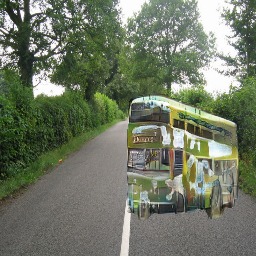} &
    \includegraphics[width=\mysizewf]{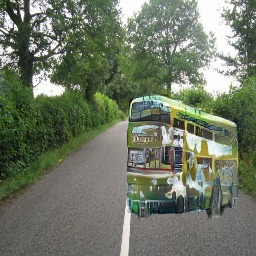} \\
    
         \includegraphics[width=\mysizewf]{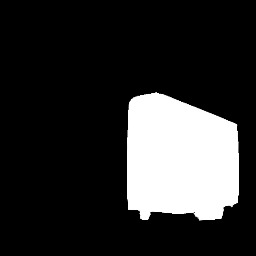} &
    \includegraphics[width=\mysizewf]{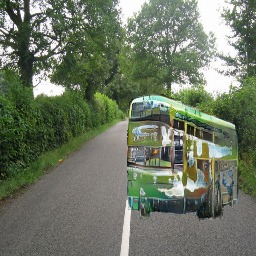} &
    \includegraphics[width=\mysizewf]{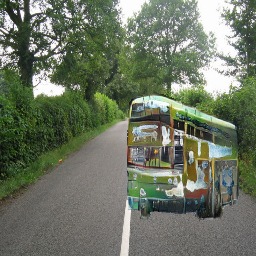} &
    \includegraphics[width=\mysizewf]{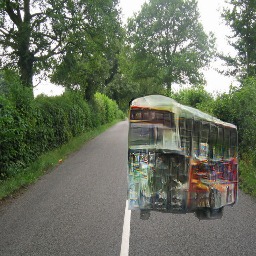} &
    \includegraphics[width=\mysizewf]{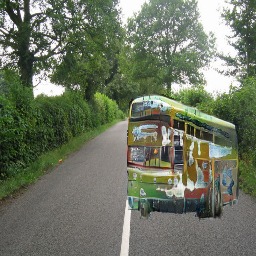} \\

\end{tabular}
 \captionof{figure}{Masks generated for the COCO class `bus' and the corresponding synthesized textures. Our model fails to capture the diversity of this dataset. Moreover, our texture generated is not realistic and fails to account for perspective.}
  \label{fig:fig_bus_supp}
 \end{table*}
\setlength{\tabcolsep}{1pt}
\begin{table*}[t]
  \centering
\begin{tabular}{ccccc}
 \includegraphics[width=\mysizewf, height=\mysizewf]{figures_cvpr/fig_giraffe_retexturing_supp/000000509735.jpg} &
    \includegraphics[width=\mysizewf]{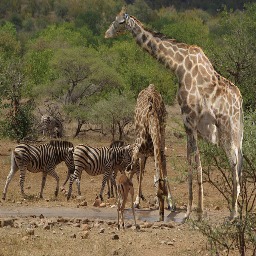} &
    \includegraphics[width=\mysizewf]{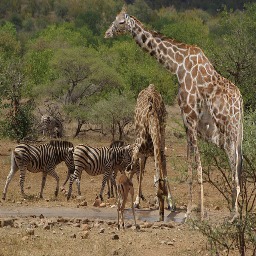} &
    \includegraphics[width=\mysizewf]{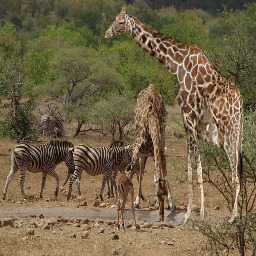} &
        \includegraphics[width=\mysizewf]{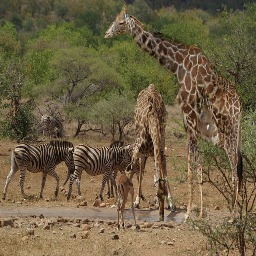} \\
      
 \includegraphics[width=\mysizewf, height=\mysizewf]{figures_cvpr/fig_giraffe_retexturing_supp/000000459757.jpg} &
     \includegraphics[width=\mysizewf]{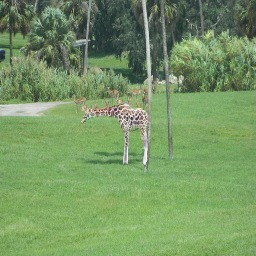} &
        \includegraphics[width=\mysizewf]{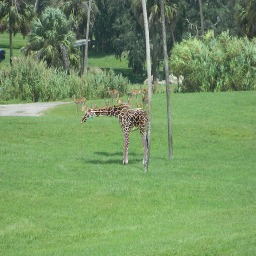} &
    \includegraphics[width=\mysizewf]{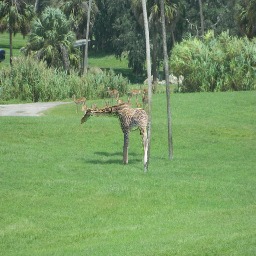} &
            \includegraphics[width=\mysizewf]{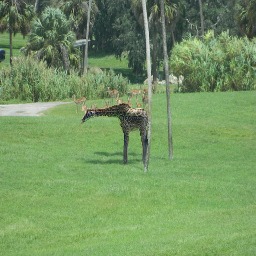} \\

 \includegraphics[width=\mysizewf, height=\mysizewf]{figures_cvpr/fig_giraffe_retexturing_supp/000000445675.jpg} &
    \includegraphics[width=\mysizewf]{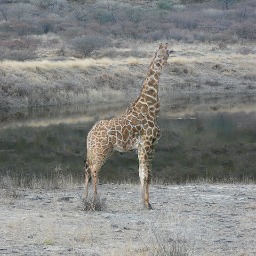} &
    \includegraphics[width=\mysizewf]{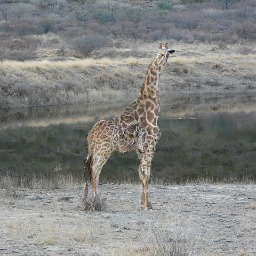} &
            \includegraphics[width=\mysizewf]{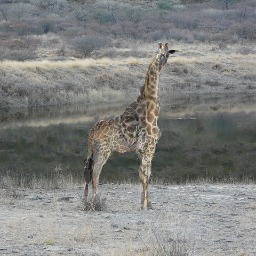} &
                    \includegraphics[width=\mysizewf]{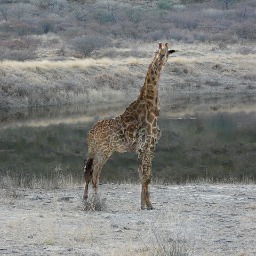}\\
            
 \includegraphics[width=\mysizewf, height=\mysizewf]{figures_cvpr/fig_giraffe_retexturing_supp/000000165336.jpg} &
    \includegraphics[width=\mysizewf]{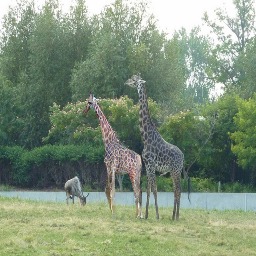} &
    \includegraphics[width=\mysizewf]{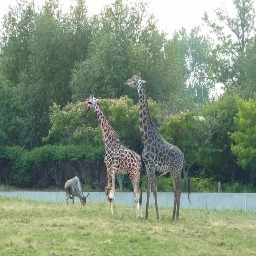} &
        \includegraphics[width=\mysizewf]{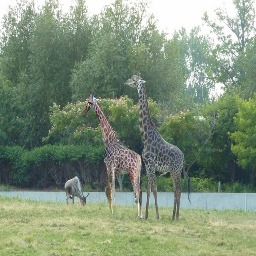} &
            \includegraphics[width=\mysizewf]{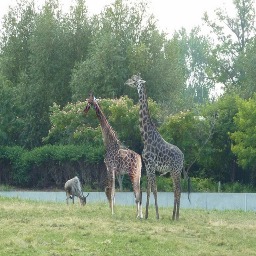} \\
            
 \includegraphics[width=\mysizewf, height=\mysizewf]{figures_cvpr/fig_giraffe_retexturing_supp/000000507042.jpg} &
     \includegraphics[width=\mysizewf]{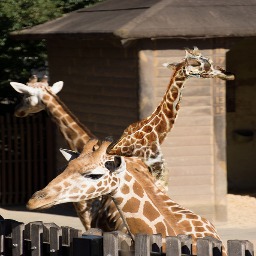} &
          \includegraphics[width=\mysizewf]{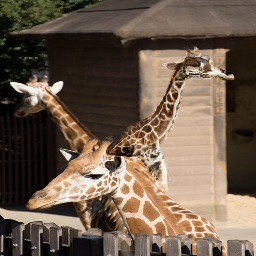} &
    \includegraphics[width=\mysizewf]{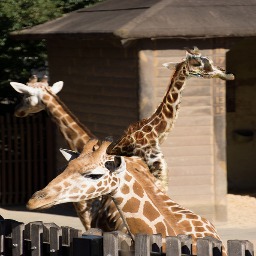} &
     \includegraphics[width=\mysizewf]{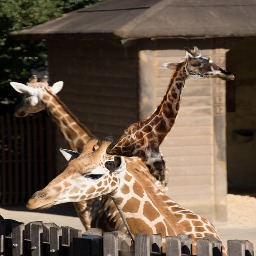}\\
            
 \includegraphics[width=\mysizewf, height=\mysizewf]{figures_cvpr/fig_giraffe_retexturing_supp/000000248810.jpg} &
  \includegraphics[width=\mysizewf]{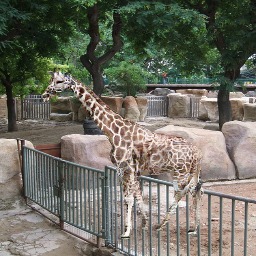} &
  \includegraphics[width=\mysizewf]{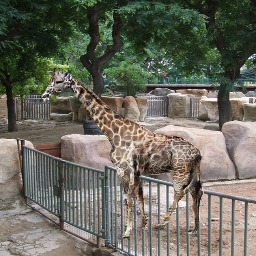} &
    \includegraphics[width=\mysizewf]{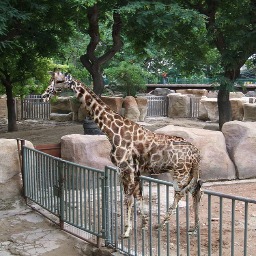} &
        \includegraphics[width=\mysizewf]{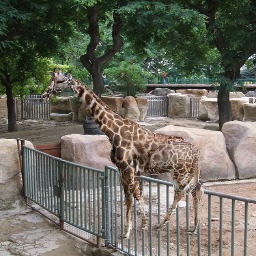} \\

\end{tabular}

 \captionof{figure}{Retexturing results for the COCO class `giraffe'. First column contains real images, subsequent columns are new generated textures. Notice how in the last row, the giraffe is supposed to be behind the fence. However because the corresponding mask does not hide the giraffe parts behind the fence, then our algorithm infills the giraffe on top of the fence, making the final result uncanny.}
  \label{fig:fig_giraffe_retexturing_supp}
 \end{table*}
\setlength{\tabcolsep}{1pt}
\begin{table*}[t]
  \centering
\begin{tabular}{ccccc}
 \includegraphics[width=\mysizewf, height=\mysizewf]{figures_cvpr/fig_pizza_retexturing_supp/000000369757.jpg} &
    \includegraphics[width=\mysizewf]{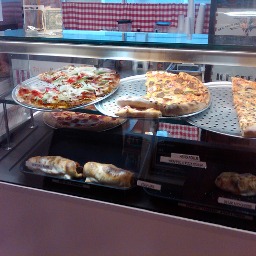} &
    \includegraphics[width=\mysizewf]{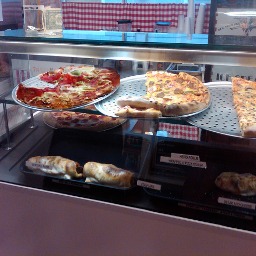} &
        \includegraphics[width=\mysizewf]{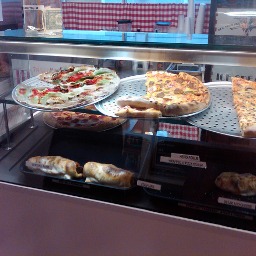} &
            \includegraphics[width=\mysizewf]{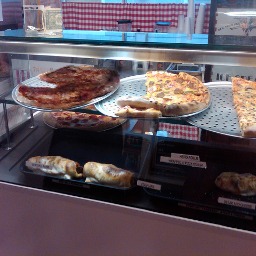} \\
      
 \includegraphics[width=\mysizewf, height=\mysizewf]{figures_cvpr/fig_pizza_retexturing_supp/000000324927.jpg} &
    \includegraphics[width=\mysizewf]{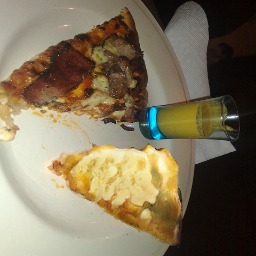} &
    \includegraphics[width=\mysizewf]{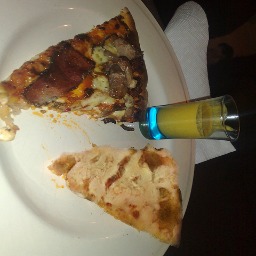} &
        \includegraphics[width=\mysizewf]{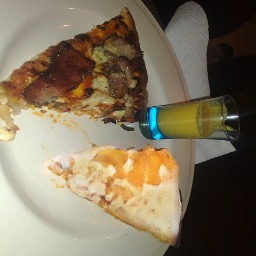} &
            \includegraphics[width=\mysizewf]{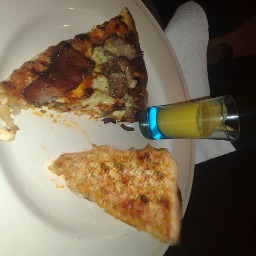} \\
        
 \includegraphics[width=\mysizewf, height=\mysizewf]{figures_cvpr/fig_pizza_retexturing_supp/000000276285.jpg} &
    \includegraphics[width=\mysizewf]{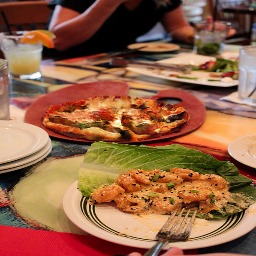} &
    \includegraphics[width=\mysizewf]{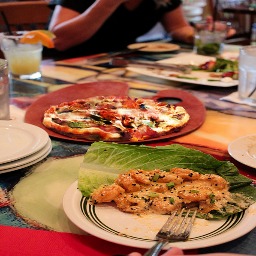} &
        \includegraphics[width=\mysizewf]{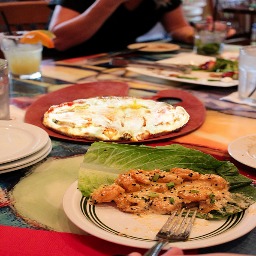} &
            \includegraphics[width=\mysizewf]{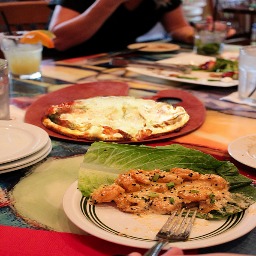} \\
            
 \includegraphics[width=\mysizewf, height=\mysizewf]{figures_cvpr/fig_pizza_retexturing_supp/000000192191.jpg} &
    \includegraphics[width=\mysizewf]{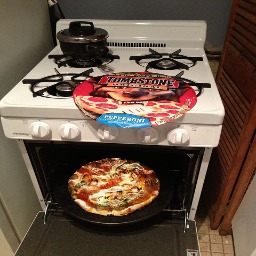} &
    \includegraphics[width=\mysizewf]{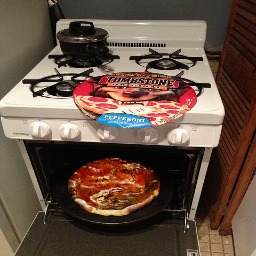} &
        \includegraphics[width=\mysizewf]{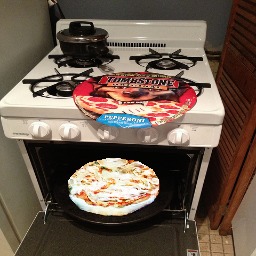} &
            \includegraphics[width=\mysizewf]{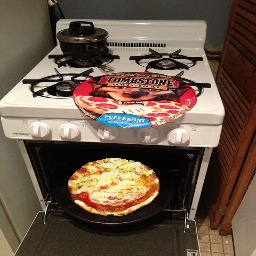} \\
            
 \includegraphics[width=\mysizewf, height=\mysizewf]{figures_cvpr/fig_pizza_retexturing_supp/000000513041.jpg} &
    \includegraphics[width=\mysizewf]{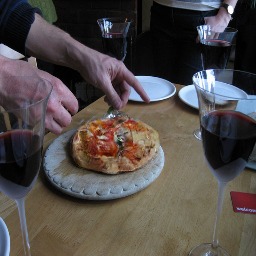} &
    \includegraphics[width=\mysizewf]{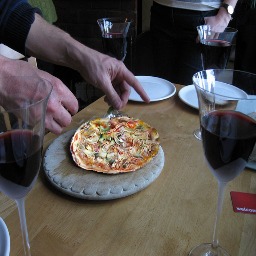} &
        \includegraphics[width=\mysizewf]{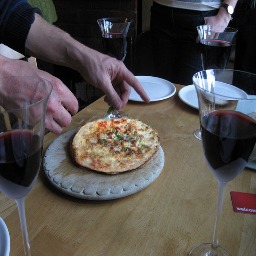} &
            \includegraphics[width=\mysizewf]{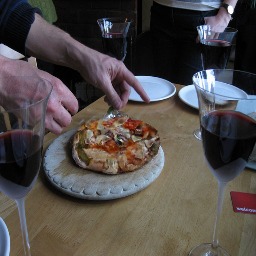} \\
            
 \includegraphics[width=\mysizewf, height=\mysizewf]{figures_cvpr/fig_pizza_retexturing_supp/000000307172.jpg} &
    \includegraphics[width=\mysizewf]{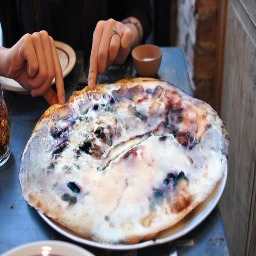} &
    \includegraphics[width=\mysizewf]{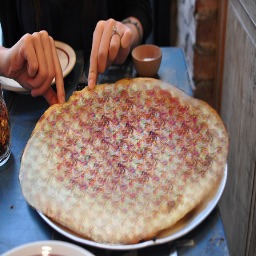} &
        \includegraphics[width=\mysizewf]{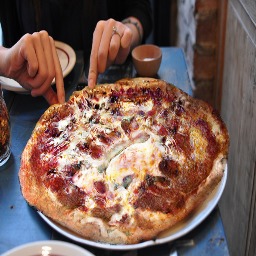} &
            \includegraphics[width=\mysizewf]{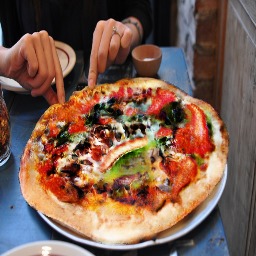} \\

\end{tabular}
 \captionof{figure}{Retexturing results for the COCO class `pizza'. First column contains real images, subsequent columns are new textures.}
  \label{fig:fig_pizza_retexturing_supp}
 \end{table*}
\setlength{\tabcolsep}{1pt}
\begin{table*}[t]
  \centering
\begin{tabular}{ccccc}
 \includegraphics[width=\mysizewf, height=\mysizewf]{figures_cvpr/fig_donut_retexturing_supp/000000378605.jpg} &
    \includegraphics[width=\mysizewf]{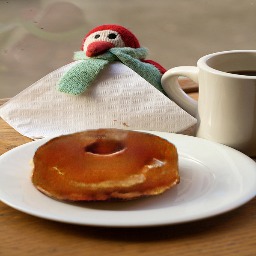} &
    \includegraphics[width=\mysizewf]{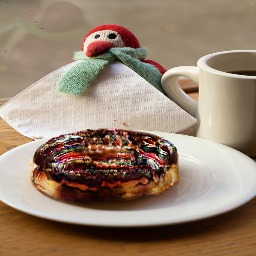} &
        \includegraphics[width=\mysizewf]{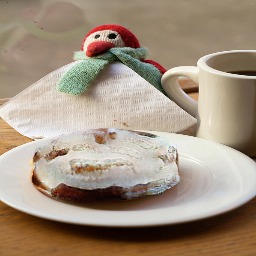} &
            \includegraphics[width=\mysizewf]{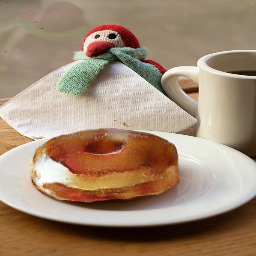} \\
      
 \includegraphics[width=\mysizewf, height=\mysizewf]{figures_cvpr/fig_donut_retexturing_supp/000000377814.jpg} &
    \includegraphics[width=\mysizewf]{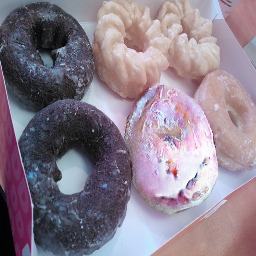} &
    \includegraphics[width=\mysizewf]{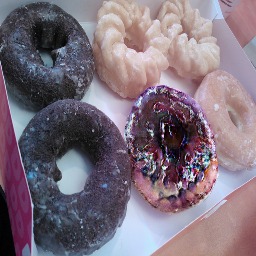} &
        \includegraphics[width=\mysizewf]{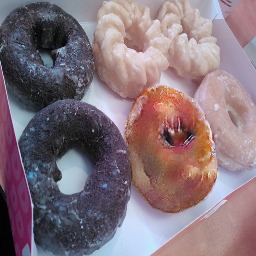} &
            \includegraphics[width=\mysizewf]{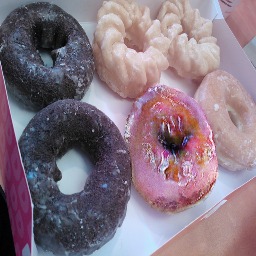} \\
        
 \includegraphics[width=\mysizewf, height=\mysizewf]{figures_cvpr/fig_donut_retexturing_supp/000000561465.jpg} &
    \includegraphics[width=\mysizewf]{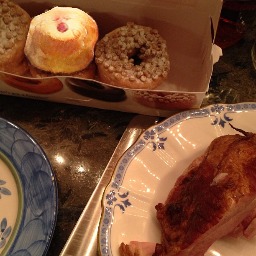} &
    \includegraphics[width=\mysizewf]{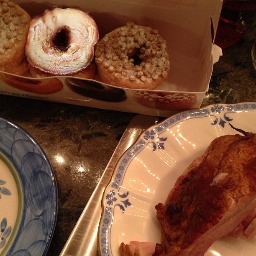} &
        \includegraphics[width=\mysizewf]{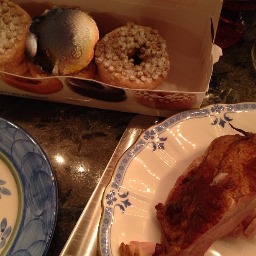} &
            \includegraphics[width=\mysizewf]{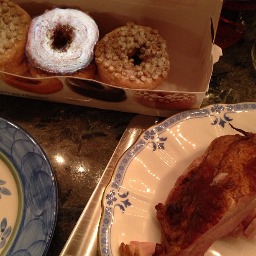} \\
            
 \includegraphics[width=\mysizewf, height=\mysizewf]{figures_cvpr/fig_donut_retexturing_supp/000000002587.jpg} &
    \includegraphics[width=\mysizewf]{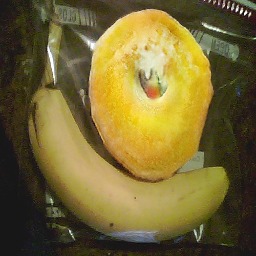} &
    \includegraphics[width=\mysizewf]{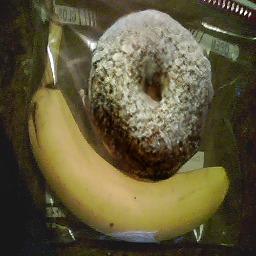} &
        \includegraphics[width=\mysizewf]{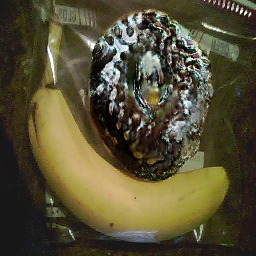} &
            \includegraphics[width=\mysizewf]{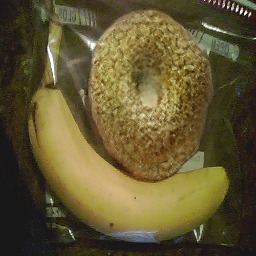} \\
            
 \includegraphics[width=\mysizewf, height=\mysizewf]{figures_cvpr/fig_donut_retexturing_supp/000000419379.jpg} &
    \includegraphics[width=\mysizewf]{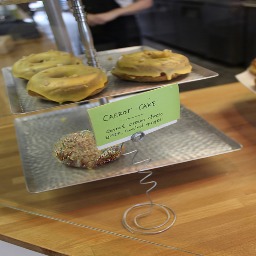} &
    \includegraphics[width=\mysizewf]{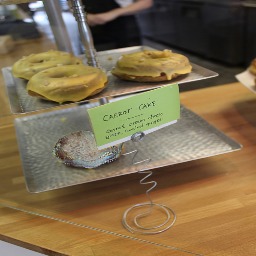} &
        \includegraphics[width=\mysizewf]{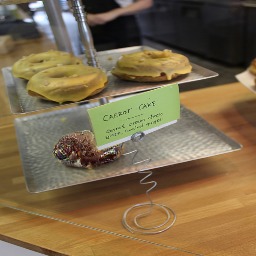} &
            \includegraphics[width=\mysizewf]{figures_cvpr/fig_donut_retexturing_supp/im_226.jpg} \\
            
 \includegraphics[width=\mysizewf, height=\mysizewf]{figures_cvpr/fig_donut_retexturing_supp/000000451150.jpg} &
    \includegraphics[width=\mysizewf]{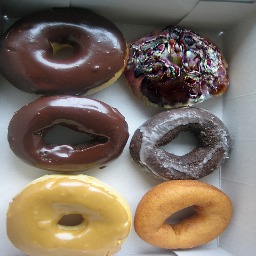} &
    \includegraphics[width=\mysizewf]{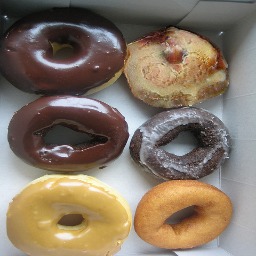} &
        \includegraphics[width=\mysizewf]{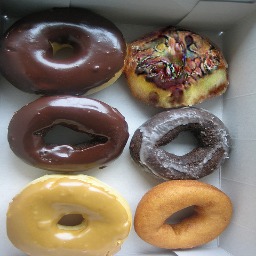} &
            \includegraphics[width=\mysizewf]{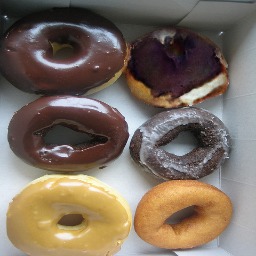} \\

\end{tabular}
 \captionof{figure}{Retexturing results for the COCO class `donut'. First column contains real images, subsequent columns are new textures.}
  \label{fig:fig_donut_retexturing_supp}
 \end{table*}

\section{Network Architectures}
Tables~\ref{tab:Garchitecture}--\ref{tab:encT} show the detailed architecture of all our network components described in the main paper.

\newcommand{\maskrow}[1]{\textcolor{Blue}{#1}}
\newcommand{\rgbrow}[1]{\textcolor{Orange}{#1}}

\begin{table}[t]
    \caption{Architecture of the shape generator $G_M$. `Conv.' is convolutional layer; `Res.' is residual block; `InstNorm' is instance normalization; `Act.' is activation function. `LReLU' denotes Leaky ReLU with a factor of 0.2.}
    \centering
    \begin{tabular}{lrrrcl}
        \toprule
         Layer & \#Filters & Size & Stride & InstNorm & Act.   \\
         \midrule
         Conv.   & 64  & $7\times7$ & 1 & \checkmark & LReLU \\
         Conv.   & 256 & $3\times3$ & 2 & \checkmark & LReLU \\
         Conv.   & 512 & $3\times3$ & 2 & \checkmark & LReLU \\
         Res.    & 512 & $3\times3$ & 1 & \checkmark & LReLU \\
         Res.    & 512 & $3\times3$ & 1 & \checkmark & LReLU \\
         Res.    & 512 & $3\times3$ & 1 & \checkmark & LReLU \\
         Res.    & 512 & $3\times3$ & 1 & \checkmark & LReLU \\
         Res.    & 512 & $3\times3$ & 1 & \checkmark & LReLU \\
         Res.    & 512 & $3\times3$ & 1 & \checkmark & LReLU \\
         Res.    & 512 & $3\times3$ & 1 & \checkmark & LReLU \\
         Res.    & 512 & $3\times3$ & 1 & \checkmark & LReLU \\
         Res.    & 512 & $3\times3$ & 1 & \checkmark & LReLU \\
         Deconv.   & 512  & $3\times3$ & 2 & \checkmark & ReLU \\
         Deconv.   & 256 & $3\times3$ & 2 & \checkmark & ReLU \\
         Conv.   & 1 & $7\times7$ & 1 & - & Tanh\\
         \bottomrule
    \end{tabular}
    \label{tab:Garchitecture}
\end{table}

\begin{table}[t]
    \caption{Residual block architecture used in the shape generation networks.}
    \centering
    \begin{tabular}{lrrrcl}
        \toprule
         Layer & \#Filters & Size & Stride & InstNorm & Act.   \\
         \midrule
         Conv.    & 512 & $3\times3$ & 1 & \checkmark & LReLU \\
         Conv.    & 512 & $3\times3$ & 1 & \checkmark & LReLU \\
         \bottomrule
    \end{tabular}
    \label{tab:GMresblockarchitecture}
\end{table}

\begin{table}[t]
    \caption{Architecture of the encoder $\text{Dec}^M$. FC refers to a fully connected layer, with 128 being the dimensionality of $\mathbf{z}_m$.}
    \centering
    \begin{tabular}{lrl}
        \toprule
         Layer & \#neurons & Act.   \\
         \midrule
         FC.    & 128  & LReLU \\
         FC.    & 128  & LReLU \\
         \bottomrule
    \end{tabular}
    \label{tab:DECm}
\end{table}

\begin{table}[t]
    \caption{Architecture of $\text{ENC}^M$. The output is of size 1024, 512 for each of the affine parameters predicted by $\text{ENC}^M$.}
    \centering
    \begin{tabular}{lrl}
        \toprule
         Layer & \#neurons & Act.   \\
         \midrule
         FC.    & 1024  & LReLU \\
         FC.    & 1024  & LReLU \\
         FC.    & 1024  & LReLU \\
         \bottomrule
    \end{tabular}
    \label{tab:ENCm}
\end{table}

\begin{table}[t]
    \caption{Architecture of both the shape and texture discriminators. `LReLU' denotes Leaky ReLU with a factor of 0.2.}
    \centering
    \begin{tabular}{lrrrcl}
        \toprule
         Layer & \#Filters & Size & Stride & InstNorm & Act.   \\
         \midrule
         Conv.   & 64  & $4\times4$ & 2 & - & LReLU \\
         Conv.   & 128 & $4\times4$ & 2 & \checkmark & LReLU \\
         Conv.   & 256 & $4\times4$ & 2 & \checkmark & LReLU \\
         Conv.   & 512  & $4\times4$ & 1 & \checkmark & LReLU \\
         Conv.   & 1 & $4\times4$ & 1 & - & Ident \\
         \bottomrule
    \end{tabular}
    \label{tab:Dencoderarchitecture}
\end{table}

\begin{table}[t]
    \caption{Architecture of the texture generator $G_T$. `Conv.' is convolutional layer; `Up+conv.' is nearest neighbor up-sampling + convolution. `Res.' is residual block; `InstNorm' is instance normalization; `Act.' is activation function. `LReLU' denotes Leaky ReLU with a factor of 0.2, `GNLRELU' denotes a Gaussian Noise Layer followed by Leaky ReLU. $\mathbf{z}_t'$ and $\mathbf{z}_t$ are tiled and concatenated layer-wise until the first residual block.}
    \centering
    \begin{tabular}{lrrrcl}
        \toprule
         Layer & \#Filters & Size & Stride & InstNorm & Act.   \\
         \midrule
         Conv.   & 64  & $7\times7$ & 1 & - & LReLU \\
         Conv.   & 128 & $3\times3$ & 2 & \checkmark & LReLU \\
         Conv.   & 256 & $3\times3$ & 2 & \checkmark & LReLU \\
         Conv.   & 512 & $3\times3$ & 2 & \checkmark & LReLU \\
         Res.    & 520 & $3\times3$ & 1 & \checkmark & GNLReLU \\
         Res.    & 520 & $3\times3$ & 1 & \checkmark & GNLReLU \\
         Res.    & 520 & $3\times3$ & 1 & \checkmark & GNLReLU \\
         Res.    & 520 & $3\times3$ & 1 & \checkmark & GNLReLU \\
         Res.    & 520 & $3\times3$ & 1 & \checkmark & GNLReLU \\
         Res.    & 520 & $3\times3$ & 1 & \checkmark & GNLReLU \\
         Res.    & 520 & $3\times3$ & 1 & \checkmark & GNLReLU \\
         Res.    & 520 & $3\times3$ & 1 & \checkmark & GNLReLU \\
         Res.    & 520 & $3\times3$ & 1 & \checkmark & GNLReLU \\
         Up+conv.   & 256  & $3\times3$ & 2 & \checkmark & GNLReLU \\
         Up+conv.   & 128 & $3\times3$ & 2 & \checkmark & GNLReLU \\
         Up+conv.   & 64 & $3\times3$ & 2 & \checkmark & GNLReLU  \\
         Conv.   & 3 & $7\times7$ & 1 & - & Tanh\\
         \bottomrule
    \end{tabular}
    \label{tab:GTarchitecture}
\end{table}

\begin{table}[t]
    \caption{Residual architecture used for the shape generation network. `GNLRELU' denotes a Gaussian Noise Layer followed by Leaky ReLU with a factor of 0.2.}
    \centering
    \begin{tabular}{lrrrcl}
        \toprule
         Layer & \#Filters & Size & Stride & InstNorm & Act.   \\
         \midrule
         Conv.    & 520 & $3\times3$ & 1 & \checkmark & GNLReLU \\
         Conv.    & 520 & $3\times3$ & 1 & \checkmark & GNLReLU \\
         \bottomrule
    \end{tabular}
    \label{tab:GTresblockarchitecture}
\end{table}

\begin{table}[t]
    \caption{Architecture of the encoder $\text{Enc}^T$. The blue row predicts the mean vector, and the orange row predicts the standard deviation vector that are used to sample $\mathbf{z}'_{t}$. When reconstructing $\mathbf{z}_t$, we only use the mean (blue row).}
    \centering
    \begin{tabular}{lrrrcl}
        \toprule
         Layer & \#Filters & Size & Stride & InstNorm & Act.   \\
         \midrule
         Conv.   & 64  & $3\times3$ & 2 & \checkmark  & LReLU \\
         Conv.   & 128 & $3\times3$ & 2 & \checkmark & LReLU \\
         Conv.   & 256 & $3\times3$ & 2 & \checkmark & LReLU \\
         Conv.   & 512  & $3\times3$ & 2 & \checkmark & LReLU \\
         Conv.   & 512  & $3\times3$ & 2 & \checkmark & LReLU \\
         Conv.   & 512  & $3\times3$ & 2 & \checkmark & LReLU \\
         \rowcolor{Blue}FC.   & 8 & - & - & - & Ident\\
         \rowcolor{Orange}FC.   & 8 & - & - & - & Ident \\
         \bottomrule
    \end{tabular}
    \label{tab:encT}
\end{table}

\end{document}